\documentclass{article}

\usepackage{hyperref}

\usepackage[accepted]{icml2021}

\usepackage{bibentry}
\usepackage{titlesec}
\usepackage[utf8]{inputenc} 
\usepackage[T1]{fontenc}    
\usepackage{bibentry}
\usepackage{booktabs}       
\usepackage{amsfonts}       
\usepackage{nicefrac}       
\usepackage{microtype}      
\usepackage{xcolor}
\usepackage[normalem]{ulem}
\usepackage{comment}
\definecolor{mydarkblue}{rgb}{0,0.08,0.45}
\usepackage{amsmath,amssymb,amsthm,amsfonts,latexsym}        
\usepackage{mathtools}
\usepackage{graphicx}
\usepackage{bm}
\usepackage{pifont}
\usepackage{subfigure}
\usepackage{color}
\usepackage{appendix}
\usepackage{enumitem}
\usepackage{environ}         
\usepackage{etoolbox}        
\usepackage{setspace}
\usepackage{enumitem}
\usepackage{mdframed}
\usepackage{adjustbox}
\usepackage{algorithm}
\usepackage{algpseudocode}
\usepackage{dsfont}
\usepackage[normalem]{ulem}
\usepackage{thmtools} 
\usepackage{thm-restate}
\usepackage{lipsum}
\usepackage{float}

\usepackage{silence}
\usepackage{todonotes}
\usepackage[capitalise,nameinlink]{cleveref}

\newcommand{\nleft}{\mathclose\bgroup\left}
\newcommand{\nright}{\aftergroup\egroup\right}

\newcommand{\x}{w}
\newcommand{\xk}{w_{k}}

\newcommand{\etakk}{\eta_{k+1}}
\newcommand{\etakm}{\eta_{k-1}}

\newcommand{\inner}[2]{\langle #1,\, #2 \rangle}
\newcommand{\mk}{m_{k}}
\newcommand{\mkm}{m_{k-1}}
\newcommand{\mkk}{m_{k+1}}

\newcommand{\xkk}{w_{k+1}}
\newcommand{\xkm}{w_{k-1}}
\newcommand{\xopt}{w^{*}}

\newcommand{\grad}[1]{\nabla f(#1)}

\newcommand{\sgradf}[2]{\nabla f_{#1}(#2)}

\newcommand{\norm}[1]{\left\|#1\right\|}
\newcommand{\normsq}[1]{\left\|#1\right\|^{2}}
\newcommand{\Tr}[1]{\mathsf{Tr}\nleft(#1\nright)}
\newcommand{\E}{\mathbb{E}}

\newcommand{\fk}{f_{i_k}}
\newcommand{\fkopt}{f_{i_k}\kern-.65em{}^{\raisebox{1pt}{\tiny $*$}}\kern.05em}
\newcommand{\fj}{f_{i}}
\newcommand{\fjopt}{f_i^*}
\newcommand{\etak}{\eta_{k}}

\newcommand{\Ak}{A_{k}}
\newcommand{\invAk}{A^{-1}_{k}}

\newcommand{\indnormsq}[2]{\left\|#1\right\|_{#2}^{2}}

\newcommand{\gradk}[1]{\nabla f_{i_k}(#1)}
\newcommand{\gradi}[1]{\nabla f_{i}(#1)}

\DeclareMathOperator*{\argmin}{arg\,min}
\newtheorem{theorem}{Theorem}
\newtheorem{lemma}{Lemma}
\newtheorem{proposition}{Proposition}

\newcommand{\defeq}{\coloneqq}

\newcommand{\aligns}[1]{\begin{align*} #1 \end{align*}}
\newcommand{\alignn}[1]{\begin{align} #1 \end{align}}

\newcommand{\sfkk}{f_{i_k}(w_k)}

\newcommand{\sfik}{f_{i}(w)}

\newcommand{\sfistar}{f_{i}(w^*)}
\newcommand{\sfimin}{f_{i}^*}

\usepackage{amsmath}
\usepackage{amssymb}
\usepackage{amsthm}
\usepackage{mathrsfs}
\usepackage{amsfonts}
\usepackage{bm}
\usepackage{xparse}

\def\1{\bm{1}}

\DeclareMathAlphabet{\mathsfit}{\encodingdefault}{\sfdefault}{m}{sl}
\SetMathAlphabet{\mathsfit}{bold}{\encodingdefault}{\sfdefault}{bx}{n}

\newcommand{\R}{\mathbb{R}}

\newcommand{\lin}[1]{\langle#1\rangle}

\DeclareMathOperator{\diag}{diag}

\newcommand{\Expect}[2][]{\mathop{\mathbb{E}}_{#1}\nleft[#2\nright]}

\usepackage{thmtools, thm-restate}

\newtheoremstyle{scplain}
  {\topsep}
  {\topsep}
  {\itshape}
  {10em}
  {\sc}
  {.}
  { }
  {}

\newtheoremstyle{indented}
  {\topsep}
  {\topsep}
  {\addtolength{\@totalleftmargin}{2em}
   \addtolength{\linewidth}{-2em}
   \parshape 1 2em \linewidth \it}
  {}
  {\sc}
  {.}
  {.25em}
  {}

\newtheorem{corollary}{Corollary}[theorem]

\newcommand{\tightsub}[1]{{\kern -.1em \raise-.1em\hbox{\tiny$#1$}}{}}
\DeclareDocumentCommand{\Normal}{g}{\mathop{\mathcal{N}}\IfNoValueF{#1}{\nleft(#1\nright)}}
\DeclareDocumentCommand{\bigO}{g}{\mathop{\mathcal{O}}\IfNoValueF{#1}{\nleft(#1\nright)}}

\newcommand{\maxiter}{T}
\newcommand{\startiter}{1}
\newcommand{\fullsum}{\sum_{k=\startiter}^\maxiter}

\NewDocumentCommand{\paren}{sm}{%
  \IfBooleanTF{#1}{(#2)}{\nleft(#2\nright)}
}
\NewDocumentCommand{\braces}{sm}{%
  \IfBooleanTF{#1}{(#2)}{\nleft\{#2\nright\}}
}
\NewDocumentCommand{\brackets}{sm}{%
  \IfBooleanTF{#1}{(#2)}{\nleft[#2\nright]}
}

\newlength{\dittowidth}

\newcommand{\cmark}{\ding{51}}%
\newcommand{\xmark}{\ding{55}}%

\newcommand{\Bk}{P_{k}}
\newcommand{\Bstart}{P_{0}}
\newcommand{\BT}{P_{T}}
\newcommand{\AT}{A_T}
\newcommand{\invBk}{P^{-1}_{k}}
\newcommand{\Bkinv}{\invBk}
\newcommand{\Bkm}{P_{\scriptscriptstyle k-1}}
\newcommand{\Bkminv}{\Bkm^{-1}}

\newenvironment{tsmath}{
    \let\olddisplaystyle\displaystyle%
    \let\displaystyle\textstyle%
}{%
    \let\displaystyle\olddisplaystyle%
    \ignorespacesafterend%
}

\newcommand{\hk}{h_k}

\def\adagrad/{{AdaGrad}}
\def\amsgrad/{{AMSGrad}}
\def\adam/{{Adam}}
\def\rmsprop/{{RMSProp}}
\def\adadelta/{{AdaDelta}}
\def\pytorch/{{PyTorch}}
\def\radam/{{RAdam}}
\def\adabound/{{AdaBound}}

\def\m!{\mkern-2mu}

\newcommand{\smallSqrtT}{\m!\sqrt{\raisebox{-1pt}{{\small $T$}}}}
\newcommand{\smallAkInverse}{A_{\scriptscriptstyle k}^{\scriptscriptstyle -1}}
\newcommand{\smallAkmInverse}{A_{\scriptscriptstyle k-1}^{\scriptscriptstyle -1}}
\newcommand{\smallAkm}{A_{\scriptscriptstyle k-1}}
\newcommand{\smallGkSqrt}{G_{k}^{\nicefrac{1}{2}}}

\newcommand{\myquote}[1]{\null~\\{\null\hspace{.05\textwidth}\begin{minipage}[t]{.90\textwidth} #1 \end{minipage}}}
\newcommand{\myquoten}[1]{{\null\hspace{.05\textwidth}\begin{minipage}[t]{.90\textwidth} #1 \end{minipage}}}

\newcommand{\hazank}{\nabla_{\!\scriptscriptstyle k}}
\newcommand{\hazanktop}{\hazank{}^{\scriptscriptstyle \!\! \top}}
\newcommand{\ggt}{\hazank\hazanktop}

\newcommand{\Akinv}{A_{\scriptscriptstyle k}^{\scriptscriptstyle -1}}

\newcommand{\Akm}{A_{\scriptscriptstyle k-1}}

\mdfdefinestyle{mdframedthmbox}{%
	leftmargin=.0\textwidth,
	rightmargin=.0\textwidth,%
	innertopmargin=0.75em,
	innerleftmargin=.5em,
	innerrightmargin=.5em,
}

\newenvironment{thmbox}
	{%
		\begin{mdframed}[style=mdframedthmbox]%
	}{%
		\end{mdframed}%
	}

\graphicspath{{Paper/figs/}}

\icmltitlerunning{Adaptive Gradient Methods Converge Faster with Over-Parameterization}

\begin{document}
\twocolumn[
\icmltitle{Adaptive Gradient Methods Converge Faster with Over-Parameterization \\(but you should do a line-search)}



\icmlsetsymbol{equal}{*}

\begin{icmlauthorlist}
\icmlauthor{Sharan Vaswani}{alberta}
\icmlauthor{Issam Laradji}{mcgill}
\icmlauthor{Frederik Kunstner}{ubc}
\icmlauthor{Si Yi Meng}{cornell}
\icmlauthor{Mark Schmidt}{ubc}
\icmlauthor{Simon Lacoste-Julien}{udem}
\end{icmlauthorlist}

\icmlaffiliation{alberta}{University of Alberta}
\icmlaffiliation{mcgill}{Mila, McGill University}
\icmlaffiliation{ubc}{University of British Columbia}
\icmlaffiliation{cornell}{Cornell University}
\icmlaffiliation{udem}{ Mila, Universit\'e de Montr\'eal}

\icmlcorrespondingauthor{Sharan Vaswani}{vaswani.sharan@gmail.com}

\icmlkeywords{Adaptive gradient methods, Over-parameterized models, Stochastic line-search}
\vskip 0.3in
]
\setlength{\abovedisplayskip}{4pt}
\setlength{\belowdisplayskip}{4pt}
\makeatletter
\def\thm@space@setup{\thm@preskip=0pt
\thm@postskip=0pt}
\makeatother


\printAffiliationsAndNotice{}  

\begin{abstract}
Adaptive gradient methods are typically used for training over-parameterized models. 
To better understand their behaviour, we study a simplistic setting -- smooth, convex losses with models over-parameterized enough to \emph{interpolate} the data. 
In this setting, we prove that \amsgrad/ with constant step-size and momentum converges to the minimizer at a faster $\bigO(1/T)$ rate. 
When interpolation is only approximately satisfied, constant step-size \amsgrad/ converges to a neighbourhood of the solution at the same rate, while \adagrad/ is robust to the violation of interpolation.
However, even for simple convex problems satisfying interpolation, the empirical performance of both methods heavily depends on the step-size and requires tuning, questioning their adaptivity. 
We alleviate this problem by automatically determining the step-size using stochastic line-search or Polyak step-sizes.
With these techniques, we prove that both \adagrad/ and \amsgrad/ retain their convergence guarantees, without needing to know problem-dependent constants. 
Empirically, we demonstrate that these techniques improve the convergence and generalization of adaptive gradient methods across tasks, from binary classification with kernel mappings to multi-class classification with deep networks.
\end{abstract}
\vspace{-4ex}
\section{Introduction}
\label{sec:introduction}
Adaptive gradient methods such as \adagrad/~\citep{duchi2011adaptive}, \rmsprop/~\citep{tieleman2012lecture}, \adadelta/~\citep{zeiler2012adadelta}, \adam/~\citep{kingma2014adam}, and \amsgrad/~\citep{reddi2019convergence} are popular optimizers for training deep neural networks~\citep{goodfellow2016deep}. 
These methods scale well and exhibit good performance across problems, making them the default choice for many machine learning applications. Theoretically, these methods are usually studied in the non-smooth, online convex optimization setting~\citep{duchi2011adaptive, reddi2019convergence} with recent extensions to the strongly-convex~\citep{mukkamala2017variants, wang2019sadam,xie2019linear} and non-convex settings~\citep{li2018convergence, ward2018adagrad,zhou2018convergence,chen2018convergence,wu2019global, defossez2020convergence,staib2019escaping}. 
Further, an online-to-batch reduction gives guarantees similar to stochastic gradient descent (SGD) in the offline setting~\citep{cesa2004generalization, hazan2014beyond, levy2018online}. 

While constant step-size \adagrad/ has been shown to be ``universal'' as it converges with any step-size in the stochastic smooth and non-smooth settings~\citep{levy2018online},
its empirical performance is often disappointing when training deep models~\citep{kingma2014adam}. 
Improving the empirical performance was indeed the main motivation behind \adam/ and other methods~\citep{tieleman2012lecture, zeiler2012adadelta} that followed \adagrad/. However, there are several discrepancies between the theory and application of these methods. Although the theory advocates for using decreasing step-sizes for \adam/, \amsgrad/ and its variants~\citep{kingma2014adam, reddi2019convergence}, a constant step-size is typically used in practice. Similarly, the standard analyses of these methods requires a decreasing momentum parameter~\citep{reddi2019convergence}, which is also fixed in practice. Consequently, there are no theoretical results corroborating the adaptivity to the step-size for the \adam/ variants used in practice.

Furthermore, adaptive gradient methods are typically used to train highly expressive, over-parameterized models~\citep{zhang2016understanding, liang2018just} capable of \emph{interpolating} the data. 
However, the standard theoretical analyses do not take advantage of these additional structural properties. 
A recent line of work~\citep{schmidt2013fast, jain2017accelerating, ma2018power, cevher2018linear, vaswani2019fast, vaswani2019painless,wu2019global, liu2020accelerating, loizou2020stochastic} focuses on the convergence of SGD in this interpolation setting. 
For a finite-sum of loss functions, interpolation implies that all the functions in the sum are minimized at the same solution. 
Under this additional assumption, these works show that constant step-size SGD converges at a faster rate for both convex and non-convex smooth functions. 

\subsection{Background and Contributions}
\label{sec:contributions}
As a first step to reconcile the theory and practice of adaptive gradient methods, we focus on their convergence in a simplistic setting - minimizing smooth, convex loss functions using models capable of interpolating the data. We study two practical methods - constant step-size \adagrad/, and \amsgrad/ with a constant step-size and constant momentum. In particular, we make the contributions below.  

\textbf{Constant step-size \adagrad/.} For smooth, convex functions,~\citet{levy2018online} prove that constant step-size \adagrad/ adapts to the smoothness and gradient noise, resulting in an $\bigO(\nicefrac{1}{T} + \nicefrac{\zeta}{\m!\sqrt{T}})$ convergence rate, where $T$ is the number of iterations and $\zeta^2$ is a global bound on the variance in the stochastic gradients. This convergence rate matches that of SGD under the same setting~\citep{moulines2011non}. 

\textit{Contribution.} In~\cref{sec:adagrad-constant}, we show that constant step-size \adagrad/ also adapts to the degree of interpolation and prove an $\bigO(\nicefrac{1}{T} + \nicefrac{\sigma}{\m!\sqrt{T}})$ rate, where $\sigma$ is the extent to which interpolation is violated. Similarly to the result of ~\citet{levy2018online}, this holds for any bounded constant step-size. 

\textbf{\amsgrad/ with constant step-size and momentum.} 
Unlike \adagrad/, the preconditioner for \amsgrad/ does not have nice structural properties~\cite{defossez2020convergence}, 
making it difficult to prove strong guarantees. 
To analyze the convergence of \amsgrad/, 
we make the simplifying assumption that its corresponding preconditioner remains bounded. 

\textit{Contribution.} With this additional assumption, 
we show that \amsgrad/ with a constant step-size and momentum converges to the minimizer 
at a $\bigO(\nicefrac{1}{T})$ rate under interpolation (\cref{sec:amsgrad-constant}). 
Unlike \adagrad/, this result requires a specific range of step-sizes that depend on the smoothness of the problem. 
In general, constant step-size \amsgrad/ converges to a neighbourhood of the solution, 
attaining an $\bigO(\nicefrac{1}{T} + \sigma^2)$ rate, 
matching the rate of constant step-size SGD~\citep{schmidt2013fast, vaswani2019fast}. 
For over-parameterized models, 
$\sigma^2 \approx 0$ and this result provides some justification for the faster 
($\bigO(\nicefrac{1}{T})$ vs. $\bigO(\nicefrac{1}{\m!\sqrt{T}})$) 
convergence of the \amsgrad/ variants used in practice. 

\textbf{Sensitivity of \adagrad/ and \amsgrad/.} Although \adagrad/ converges at the same asymptotic rate for any step-size, it is unclear how the choice of step-size affects its practical performance. On the other hand, our theoretical results for convex minimization indicate that \amsgrad/ is sensitive to the step-size, converging only for a specific range that depends on typically unknown problem-dependent constants. 

\textit{Contribution.} In~\cref{sec:sensitivity}, we empirically demonstrate this sensitivity to the step-size for both methods. In particular, we show that even for a convex problem satisfying interpolation, such as logistic regression on linearly separable data, the choice of step-size has a big impact on the performance of both \adagrad/ and \amsgrad/, questioning their adaptivity. 

\textbf{\adagrad/ and \amsgrad/ with an adaptive step-size.} To improve the robustness to the step-size, we use recent techniques~\citep{vaswani2019fast, loizou2020stochastic} that automatically determine the step-size for each iteration of SGD. These works use stochastic variants of the classical Armijo line-search~\citep{armijo1966minimization} and Polyak step-size~\citep{polyak1963gradient} and prove their convergence under interpolation.

\textit{Contribution.} In~\cref{sec:adaptive}, we modify these techniques for their use with adaptive gradient methods. In particular, we show that a variant of stochastic line-search (SLS) can be used to set the step-size in each iteration of \adagrad/. SLS enables \adagrad/ to adapt to the smoothness of the underlying function, while retaining its favourable convergence properties (\cref{sec:adagrad-adaptive}). Under the same bounded preconditioner assumption, we prove that \amsgrad/ used with a variant of SLS or stochastic Polyak step-size (SPS) can match the convergence rate of its constant step-size counterpart, but without requiring the knowledge of problem-dependent constants (\cref{sec:amsgrad-adaptive}). For the logistic regression example considered in~\cref{sec:sensitivity}, we observe that using SLS/SPS improves the convergence of both \adagrad/ and \amsgrad/, matching or out-performing the best constant step-size. 

\textbf{Large-scale experiments.} To demonstrate the empirical advantage of SLS/SPS beyond convex minimization, we consider both convex and non-convex tasks, ranging from binary classification with a kernel mapping to multi-class classification with standard deep network architectures (\cref{sec:experiments}). We benchmark the performance of \adagrad/ and \amsgrad/ equipped with SLS, and compare against tuned \adam/ and recently proposed variants~\citep{luo2019adaptive, liu2019variance}. To disentangle the effects of the step-size and the adaptive preconditioner, we compare against SGD using SLS~\citep{vaswani2019painless} and SPS~\citep{loizou2020stochastic}. We find that both the step-size and the adaptive preconditioner contribute to good performance, with the SLS variants of \adagrad/ and \amsgrad/ consistently outperforming other methods. Furthermore, for the experiments with deep neural networks, these variants generalize better than SGD, demonstrating the effect of the step-size in the generalization performance~\citep{nar2018step}. 
\section{Problem formulation}
\label{sec:notation}
\begin{table*}[t]
    \centering
    \caption[]{
    Adaptive preconditioners (analyzed methods are \textbf{bolded}), with $G_0\m!=\m!0$ and $\beta_{1}, \beta_{2}\m!\in\m![0,1)$.
    In practice, a small $\epsilon I$ is added 
    to ensure $A_k\m!\succ\m!0$.
	*: We use the \pytorch/ implementation in experiments which includes bias correction.
    }
    \vspace{.25em}
    \label{tab:adaptive-updates}
    \begin{tabular}{@{}llll@{}}
        \toprule
        Optimizer & $G_k$ \hfill $(\hazank\defeq\gradk{\xk})$ \hfill\null & $A_k$ & $\beta$ \\ \midrule 
        \textbf{\adagrad/}
	        & $G_{k-1}+\diag(\ggt)$             
	        & $\smallGkSqrt$      
	        & 0
        \\
        \rmsprop/
	        &  $\beta_{2} G_{k-1} + (1-\beta_{2})\diag(\ggt)$            
	        & $\smallGkSqrt$      
	        & 0
        \\
        \adam/
	        & $(\beta_{2} G_{k-1}+(1-\beta_{2})\diag(\ggt))/(1-\beta_{2}^k)$   
	        & $\smallGkSqrt$      
	        & $\beta_1$
        \\
        \textbf{\amsgrad/}*
	        & $(\beta_{2} G_{k-1}+(1-\beta_{2})\diag(\ggt))/(1-\beta_{2}^k)$
	        & $\max\{A_{k-1}, \smallGkSqrt\}$ 
            & $\beta_1$
        \\ \bottomrule \\
    \end{tabular}
    \vspace{-1em}
\end{table*}
Adaptive methods are still poorly understood, and state-of-the-art analyses~\citep{levy2018online, reddi2019convergence, alacaoglu2020new} do not show an improvement over stochastic gradient descent in the worst-case. The objective of our theoretical analysis is to better understand the interplay 
between over-parameterization, step-sizes and momentum. 
To this end, we make the simplifying assumptions described in this section.

We consider the unconstrained minimization of an objective $f:\R^d\rightarrow\R$ with a finite-sum structure, $f(w)=\frac{1}{n}\sum_{i=1}^n f_i(w)$. In supervised learning, $n$ represents the number of training examples, and $f_i$ is the loss of training example $i$. Although we focus on the finite-sum setting, our results can be generalized to the online optimization setting. We assume $f$ and each $f_i$ are differentiable, convex, and lower-bounded by $f^*$ and $\fjopt$, respectively. Furthermore, we assume that each function $\fj$ in the finite-sum is $L_i$-smooth, implying that $f$ is $L_{\max}$-smooth with $L_{\max} \m!=\m! \max_{i} L_i$. We include formal definitions of these properties in~\cref{sec:setup}.

We also assume that the iterates remain bounded in a ball of radius $D$ around a global minimizer, $\norm{\xk - \xopt} \leq D$ for all $\xk$~\citep{ahn2020sgd}. We remark that the bounded iterates assumption simplifies the analysis but is not essential. Indeed, similarly to~\cite{reddi2019convergence, duchi2011adaptive, levy2018online} we can consider constrained minimization over a compact, feasible set $F$ with a bounded diameter $D$, and our theoretical results can be easily extended to include an explicit projection step. Without an explicit projection step, we believe it is possible to prove that the iterates do remain bounded with high-probability~\citep{mertikopoulos2020almost}. This would complicate the analysis without changing the conclusions, and we thus leave this for future work.

The interpolation assumption means that the gradient of \emph{each} $f_i$ in the finite-sum converges to zero at an optimum. If the overall objective $f$ is minimized at $\xopt$, $\grad{\xopt} = 0$, then for all $\fj$ we have $\gradi{\xopt} = 0$. The interpolation condition can be exactly satisfied for numerous machine learning models such as linear classification on a separable dataset, non-parametric kernel regression without regularization~\citep{belkin2019does,liang2018just} 
and over-parameterized deep neural networks~\citep{zhang2016understanding}. Since exact interpolation is a relatively strong assumption, we consider a weaker version~\citep{loizou2020stochastic} of it. Specifically, we measure the extent to which interpolation is violated by the disagreement between the minimum overall function value $f^*$ and the minimum value of each individual functions $\fjopt$, $\sigma^2 \coloneqq \E_i [f^* - \fjopt ] \in [0,\infty)$. Interpolation is said to be exactly satisfied if $\sigma^2 = 0$. Since $\sigma^2$ only depends on the minimum function values, the minimizer $\xopt$ of $f$ need not be unique for $\sigma^2$ to be uniquely defined.  

We first consider the update for a generic adaptive gradient method at iteration $k$. For a preconditioner matrix $\Ak$ and a constant momentum parameter $\beta \in [0,1)$, the update is
\begin{align}
\xkk & = \xk - \etak \, \smallAkInverse \mk \nonumber \\
\mk & = \beta  \mkm + (1 - \beta) \gradk{\xk}.
\label{eq:gen-update}    
\end{align}
Here, $\gradk{\xk}$ is the stochastic gradient of a randomly chosen function $\fk$, and $\etak$ is the step-size. Adaptive gradient methods typically differ in how their preconditioners are constructed and whether or not they include the momentum term $\beta \mkm$ (see~\cref{tab:adaptive-updates} for a list of common methods). Both \rmsprop/ and \adam/ maintain an exponential moving average of past stochastic gradients, but as \citet{reddi2019convergence} pointed out, unlike \adagrad/, the corresponding preconditioners do not guarantee that $A_{k+1}\m!\succeq\m!\Ak$ and the resulting per-dimension step-sizes do not go to zero. This can lead to large fluctuations in the effective step-size and prevent these methods from converging. To mitigate this problem, they proposed \amsgrad/, which ensures $A_{k+1}\m!\succeq\m!\Ak$ and the convergence of iterates to a stationary point. Consequently, our theoretical results focus on \adagrad/ and \amsgrad/, though we considered \adam/ in our experiments.

Although our theory holds for both the full matrix and diagonal variants 
of these methods, we use only the latter in experiments for scalability. The diagonal variants perform a per-dimension scaling of the gradient and avoid computing full matrix inverses, so their per-iteration cost is the same as SGD, although with an additional $\bigO(d)$ memory. Unlike \adagrad/, the \amsgrad/ preconditioner does not possess nice structural properties. To prove convergence results for \amsgrad/, we assume that the preconditioners are well-behaved in the sense that their eigenvalues are bounded in an interval $[a_{\min}, a_{\max}]$. 
This is a common assumption in the analysis of adaptive methods, and a small diagonal matrix ($\epsilon I_d$) is typically added to the preconditioners to ensure they remain positive definite. For diagonal preconditioners, this boundedness property is easy to verify, and it is also inexpensive to maintain the desired range by an explicit projection. Furthermore, observe that even though the stochastic gradients can become zero due to over-parameterization, both the $\max$ operation and diagonal matrix used in constructing the \amsgrad/ preconditioner ensures its positive definiteness.

\vspace{-2ex}
\section{Convergence with a constant step-size}
\newcommand{\fB}{f_{\!\scriptscriptstyle B}}
\newcommand{\xoptB}{\x_{\m!\scriptscriptstyle B}^*}
\newcommand{\batchvar}{\sigma_b^{\scriptscriptstyle 2}}
In this section, we analyze the convergence of constant step-size \adagrad/ (\cref{sec:adagrad-constant}) and \amsgrad/ with a constant step-size and momentum parameter (\cref{sec:amsgrad-constant}). 

\subsection{Constant step-size \adagrad/}
\label{sec:adagrad-constant}
For smooth, convex objectives,~\citet{levy2018online} showed that \adagrad/ converges at a rate $\bigO(\nicefrac{1}{T} + \nicefrac{\zeta}{\m!\sqrt{T}})$, 
where $\zeta{}^2 = \sup_\x \E_i[\norm{\grad{\x}-\gradi{\x}}\!{}^2]$ is a uniform bound on the variance of the stochastic gradients. We show that constant step-size \adagrad/ achieves the $\bigO(\nicefrac{1}{T})$ rate when interpolation is exactly satisfied ($\sigma^2 = 0$) and a slower convergence to the minimizer if interpolation is violated. The proofs for this section are in~\cref{app:adagrad-proofs}.

\begin{restatable}[Constant step-size \adagrad/]{theorem}{restateThmConstantStepSizeAdagrad}
\label{thm:adagrad-constant}
Assuming (i) convexity, (ii) $L_{\max}$-smoothness of each $f_i$, and (iii) bounded iterates, \adagrad/ with a constant step-size $\eta$ and uniform averaging $\bar{\x}_T =\frac{1}{T}\sum_{k=1}^{\raisebox{-.5pt}{\tiny $T$}} \xk$, converges at a rate
\begin{align*}
	\Expect{f(\bar{\x}_T) - f^*} 
	\leq \frac{\alpha }{T} + \frac{ \sqrt{\alpha} \sigma}{\sqrt{T}},
	\quad 
\end{align*}
where $\alpha = \frac{1}{2}\big(\frac{D^2}{\eta} + 2 \eta\big)^{\m!2} d L_{\max}$.
\end{restatable}

Theorem~\ref{thm:adagrad-constant}
shows that \adagrad/ is robust to the violation of interpolation and converges to the minimizer at the desired rate for \emph{any} reasonable step-size. In the over-parameterized setting, $\sigma^2$ can be much smaller than $\zeta^2$~\citep{zhang2019stochastic}, implying a faster convergence compared to the result of~\citet{levy2018online}. In particular, when interpolation is satisfied, $\sigma^2 = 0$ while $\zeta^2$ can still be large. 

In the online convex optimization framework, for smooth functions, a similar proof technique can be used to show that \adagrad/ incurs only $O(1)$ regret when interpolation is exactly satisfied and retains its $\bigO(\smallSqrtT)$-regret guarantee in the general setting (\cref{thm:adagrad-regret} in~\cref{app:adagrad-regret}). 
A similar first-order regret bound was proven independently by~\citet{orabona2019modern},
for a scalar version of \adagrad/. Our theorem generalizes their result to use a matrix preconditioner. 

\subsection{Constant step-size \amsgrad/}
\label{sec:amsgrad-constant}
\begin{figure*}[t]
    \centering
    \subfigure[\adagrad/]{
    \includegraphics[width=.49\textwidth]{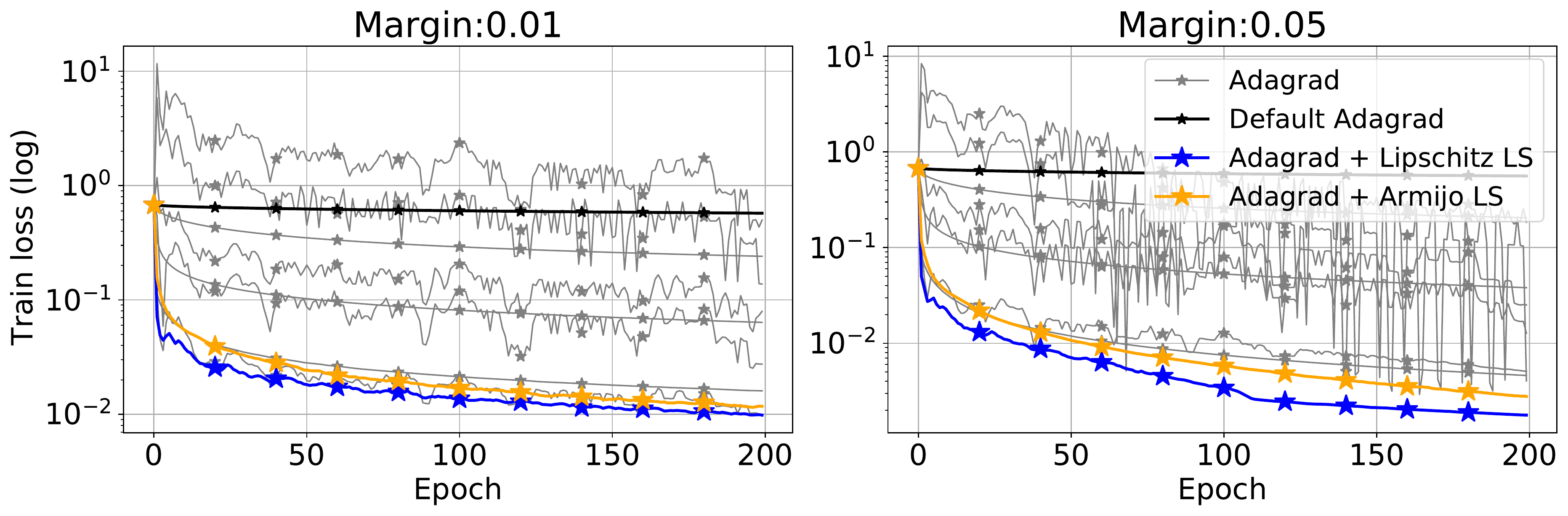}
    }%
    \subfigure[\amsgrad/]{
    \includegraphics[width=.49\textwidth]{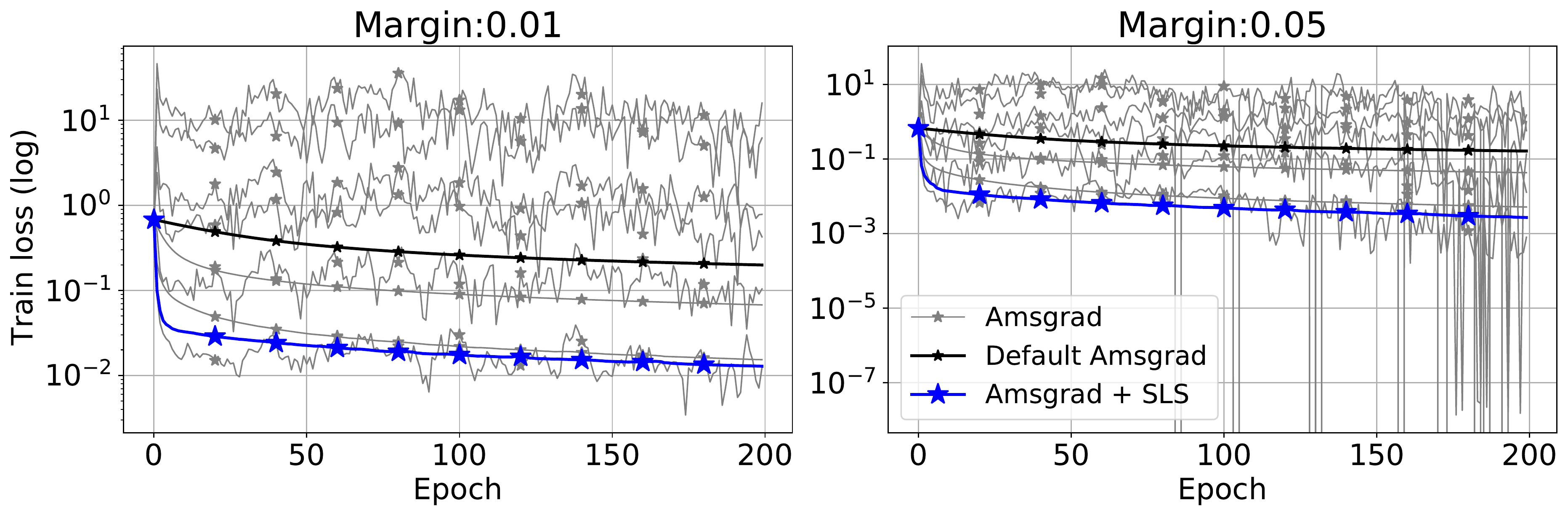}
    }
    \caption{Logistic regression on a linearly-separable synthetic dataset to show the impact of the step-size on performance. 
    We compare \adagrad/ and \amsgrad/ (fixed momentum) with varying step-sizes, including the default in \pytorch/, against SLS. 
    \adagrad/ and \amsgrad/ are sensitive to the choice of step-size, 
    while the SLS variants of both methods match or out-perform the best constant step-size.
    }
    \label{fig:syn}
\end{figure*}
As explained earlier, in order to analyze the convergence of \amsgrad/, we assume that the effect of the preconditioning is bounded, meaning that the eigenvalues of $\Ak$ lie in the $\left[ a_{\min}, a_{\max}\right]$ range. This is a common assumption in the analysis of preconditioned gradient and second-order methods~\citep{yu2010quasi, berahas2016multi, moritz2016linearly, bollapragada2018progressive, meng2019fast}. We consider \amsgrad/ without bias correction, as its effect is minimal after the first few iterations. The proofs for this section are in~\cref{app:amsgrad-proofs} and~\cref{app:amsgrad-momentum-proofs}. 

The original analysis of \amsgrad/~\citep{reddi2019convergence} uses a \emph{decreasing step-size and a decreasing momentum parameter}. Assuming a bounded diameter of the feasible set, bounded maximum eigenvalue of the positive-definite preconditioner and bounded gradients, it shows an $\bigO(\nicefrac{D^2 a_{\max}}{\!\sqrt{T}})$ convergence for \amsgrad/ in the non-smooth, convex setting. Under the same assumptions,~\citet{alacaoglu2020new} showed that this analysis is loose and that \amsgrad/ does not require a decreasing momentum parameter (but still requires a decreasing step-size) to obtain the $\bigO(\nicefrac{1}{\m!\sqrt{T}})$ rate. However, in practice, \amsgrad/ is typically used with \emph{both a constant step-size and momentum parameter}. Next, we analyze the convergence for this commonly-used variant of \amsgrad/. 
\begin{restatable}{theorem}{restateThmConstantAmsgradmom}
\label{thm:amsgrad-constant-mom}
Under the same assumptions as~\cref{thm:adagrad-constant}, 
and assuming (iv) non-decreasing preconditioners 
(v) bounded eigenvalues in the $[a_{\min}, a_{\max}]$ interval, 
where $\kappa = \nicefrac{a_{\max}}{a_{\min}}$, 
\amsgrad/ with $\beta \in [0,1)$, 
constant step-size $\eta = \frac{1-\beta}{1+\beta}\frac{a_{\min}}{2L_{\max}}$ 
and uniform averaging converges at a rate,
\aligns{
	\Expect{f(\bar{\x}_T) - f^*}
	\leq
	\paren{\frac{1+\beta}{1-\beta}}^{\!\!2}
	\frac{2 L_{\max} D^2 d \kappa}{  T}
	+ \sigma^2.
}
\end{restatable}
In contrast to~\citep{reddi2019convergence, alacaoglu2020new}, the above theorem considers the smooth setting and does not require the bounded gradient assumption. However, our rate has an additional dependence on $a_{\min}$, the minimum eigenvalue of the preconditioner. Unless we take advantage of some structure in the \amsgrad/ preconditioner, we believe that such a dependence on $a_{\min}$ is unavoidable in the constant step-size setting. For the common diagonal variant of \amsgrad/, we can project the preconditioner entries onto a reasonable $[a_{\min}, a_{\max}]$ interval, and ensure $\kappa$ is small. 

When $\sigma = 0$, we obtain a faster $\bigO(\nicefrac{1}{T})$ convergence to the minimizer. When interpolation is only approximately satisfied, unlike \adagrad/ which converges to the minimizer at a slower rate, \amsgrad/ converges to a neighbourhood whose size depends on $\sigma^2$. A similar distinction between the convergence of constant step-size \adam/ (or \amsgrad/) vs. \adagrad/ has also been discussed in the non-convex setting~\citep{defossez2020convergence}. We observe that the ``noise'' $\sigma^2$ is not amplified because of the non-decreasing momentum (or step-size), in contrast to the stochastic accelerated gradient method~\citep{devolder2014first, vaswani2019fast}. 
 
Since \amsgrad/ is typically used for optimizing large, over-parameterized models, the violation of interpolation is small and $\sigma^2 \approx 0$. Another reason that explains the practical effectiveness of \amsgrad/ is the use of large batch-sizes that result in a smaller $\sigma^2$. To understand the effect of the batch-size, note that if we use a batch-size of $b$, $\batchvar \coloneqq \E_{B; |B| = b} \left[\fB(\xopt) - \fB(\xoptB) \right]$ where $\x^*_B$ is the minimizer of a batch $B$ of training examples. By convexity, $\batchvar \leq \E [\norm{\xopt - \xoptB} \norm{\nabla \fB(\xopt)}]$. If we assume that the distance $\norm{\xopt - \xoptB}$ is bounded, $\batchvar \propto \E \Vert\nabla \fB(\xopt)\Vert$. Since the examples in each batch are sampled with replacement, using the bounds in~\citep{lohr2009sampling}, $\batchvar \propto \frac{n  - b}{n b} \norm{\nabla f_{i}(\xopt)}$, showing that $\batchvar$ shrinks as the batch-size becomes larger, becoming zero for the full-batch variant. With over-parameterization and large batch-sizes, $\batchvar$ is small enough for constant step-size \amsgrad/ to be useful for machine learning tasks that do not require exact convergence to the solution. 

Finally, observe that the constant step-size required for the above result depends on $L_{\max}$, which is typically unknown and difficult to estimate. In the next section, we empirically demonstrate the sensitivity of \amsgrad/ to its step-size.




\section{Sensitivity of \adagrad/ and \amsgrad/}
\label{sec:sensitivity}
In this section, we empirically demonstrate that the commonly-used constant step-size variants of \adagrad/ and \amsgrad/ are sensitive to their choice of step-size, even for convex problems satisfying interpolation.
We use their \pytorch/ implementations~\citep{paszke2017automatic} on a binary classification task with (unregularized) logistic regression. Following the protocol of~\citet{meng2019fast}, we generate a linearly-separable dataset with $n\m!=\m!10^3$ examples, ensuring interpolation is satisfied, and $d\m!=\m!20$ features with varying margins. For \adagrad/ and \amsgrad/ with a batch-size of $100$, we show the training loss for a grid of step-sizes in the $[10^{3}, 10^{-3}]$ range along with their default step-size in \pytorch/. 
In~\cref{fig:syn}, we observe a large variance across step-sizes and poor performance of the default step-size. The best performing variant of \adagrad/ and \amsgrad/ has a step-size of order $10^2$, which is very different from the default step-size used in practice. The above experiment demonstrates the inability of these methods to adapt to the properties of the function, questioning their robustness. In the next section, we design techniques to enable both \adagrad/ and \amsgrad/ to adapt to the 
objective's smoothness. 


\section{Convergence with adaptive step-sizes} 
\label{sec:adaptive}
In this section, we modify the stochastic line-search (SLS)~\citep{vaswani2019painless} and stochastic Polyak's step-size (SPS)~\citep{loizou2020stochastic} techniques to automatically determine the step-size in each iteration of \adagrad/ (\cref{sec:adagrad-adaptive}) and \amsgrad/ (\cref{sec:amsgrad-adaptive}). 

\subsection{Adaptive step-size \adagrad/}
\label{sec:adagrad-adaptive}
From~\cref{fig:syn}, we have seen that the performance of \adagrad/ depends heavily on choosing the correct step-size. To alleviate this problem, we use a \emph{conservative Lipschitz line-search} that sets the step-size on the fly, improving the empirical performance of \adagrad/ (\cref{sec:experiments}), while retaining its favourable convergence guarantees. 

\textbf{Lipschitz line-search}: At each iteration $k$, this line-search selects a step-size $\etak \leq \eta_{\max}$ that satisfies the property
\begin{align}
\begin{aligned}
&\fk(\xk - \etak \nabla \fk(\xk)) 
\\
&\quad
\leq \fk(\xk) -  c \, \etak \norm{\nabla \fk(\xk)}^2.
\end{aligned}
\label{eq:conservative-lip-linesearch}
\end{align}
Here, $\eta_{\max}$ is the upper-bound on the step-size. The resulting step-size is used in the standard \adagrad/ update (Eq.~\ref{eq:gen-update}). To find an acceptable step, our results use a backtracking line-search, described in~\cref{app:exp-details}. For simplicity, the theoretical results assume access to the largest step-size that satisfies the above condition.\footnotemark{} Here, $c$ is a hyper-parameter determined theoretically and set to $\nicefrac{1}{2}$ in our experiments. 
\footnotetext{The difference between the exact and backtracking line-search is minimal, and the bounds are only changed by a constant depending on the backtracking parameter.}

We refer to it as the Lipschitz line-search as it is only used to estimate the local Lipschitz constant. Unlike the classical Armijo line-search for preconditioned gradient descent~\citep{armijo1966minimization}, the line-search in~\cref{eq:conservative-lip-linesearch} is in the gradient direction, even though the update is in the preconditioned direction. Intuitively, the Lipschitz line-search enables \adagrad/ to take larger steps at iterates where the underlying function is smoother. 

By choosing $\eta_{\max} = \eta_{k-1}$ for iteration $k$, we obtain a conservative variant of the Lipschitz line-search. The \emph{conservative Lipschitz line-search} imposes a non-increasing constraint on the step-sizes, which is essential for convergence to the minimizer when interpolation is violated.\footnotemark{} The resulting step-size is guaranteed to be in the range $\left[\nicefrac{2(1-c)}{L_{\max}},\; \eta_{k-1} \right]$~\citep{vaswani2019painless} and allows us to prove the following theorem. 
\footnotetext{If interpolation is exactly satisfied, we can obtain an $\bigO(\nicefrac{1}{T})$ convergence without the conservative step-sizes (\cref{app:adagrad-interpolation}).}

\begin{restatable}{theorem}{restateThmLinesearchAdagrad}
\label{thm:adagrad-linesearch-conservative}
Under the same assumptions as~\cref{thm:adagrad-constant}, \adagrad/ with a conservative Lipschitz line-search with $c = 1/2$, and uniform averaging converges at a rate
\aligns{
\Expect{f(\bar{\x}_T) - f^*} \leq \frac{\alpha}{T} + \frac{\sqrt{\alpha}\sigma}{\sqrt{T}},
}
where $\alpha = \frac{1}{2} \big(D^2 \max\big\{\frac{1}{\eta_{0}}, L_{\max}\big\} + 2 \, \eta_{0}\big)^{\m!2} d L_{\max}$. 
\end{restatable}
Here, $\eta_0$ is the step-size used to initialize the line-search in the first iteration of \adagrad/ and is set to a large value in practice. The Lipschitz line-search attains the same rate as constant step-size \adagrad/ but automatically chooses a step-size in each iteration, and allows \adagrad/ to adapt to the function's local smoothness properties. This enables it perform better on some problems, even though its worst-case performance is the same as the constant step-size variant.  

\subsection{Adaptive step-size \amsgrad/}
\label{sec:amsgrad-adaptive}
For \amsgrad/, we consider variants of both the stochastic line-search~\citep{vaswani2019painless} and the stochastic Polyak step-sizes~\citep{loizou2020stochastic, berrada2019training}. 

\textbf{Armijo stochastic line-search}: We design a stochastic variant of the \emph{Armijo line-search}~\citep{armijo1966minimization} to determine the step-size. Unlike the Lipschitz line-search whose sole purpose is to estimate the smoothness constant, the \emph{stochastic Armijo line-search} (Armijo SLS) selects a suitable step-size in the preconditioned stochastic gradient direction. In particular, it returns the largest step-size $\etak$ satisfying the following conditions at iteration $k$, $\etak \leq \eta_{\max}$ and 
\begin{align}
\begin{aligned}
&\fk(\xk - \etak \Ak^{-1} \gradk{\xk}) \\
&\quad \leq \fk(\xk) - c \etak \indnormsq{\gradk{\xk}}{\Ak^{-1}}. 
\end{aligned}
\label{eq:armijo-ls}    
\end{align}
Similarly to~\cref{eq:conservative-lip-linesearch}, the step-size is upper-bounded by $\eta_{\max}$ 
(typically chosen to be a large value). Unlike the Lipschitz line-search, the stochastic Armijo line-search guarantees descent on the current function $\fk$ and that $\etak$ lies in the $\left[\nicefrac{2 a_{\min} \, (1-c)}{L_{\max}}, \eta_{\max}\right]$ range. 

\textbf{Armijo stochastic Polyak step-size}: We first define the stochastic Polyak step-size (SPS) and Armijo SPS, its modification for the adaptive case. 
\begin{align*}
	\text{SPS:} \,\,\,
	\etak & = \min \m!\bigg\{
		\frac{\fk(\xk) - \fkopt}{c \normsq{\sgradf{i_k}{\xk}}}, \eta_{\max} 
	\bigg\}\m!,
	\\
	\text{Armijo SPS:} \,\,\,
	\etak & = \min \m!\bigg\{
		\frac{\fk(\xk) - \fkopt}{c \normsq{\sgradf{i_k}{\xk}}_{\Ak^{-1}}}, \eta_{\max}
	\bigg\}\m!.
\end{align*}
Here, $\fkopt$ is the minimum value for the function $\fk$. The SPS variants require knowledge of $\fjopt$ for each function in the finite-sum. This value is difficult to obtain for general functions but is readily available in the interpolation setting for many machine learning applications. For example, common loss functions are lower-bounded by zero, and the interpolation setting ensures that these lower-bounds are tight. Consequently, using SPS with $\fjopt \m!=\m! 0$ has been shown to yield good performance for over-parameterized problems~\citep{loizou2020stochastic, berrada2019training}. The advantage of SPS over a line-search is that it does not require a backtracking procedure to set the step-size. 

For \amsgrad/, we propose to use a \emph{conservative} variant of Armijo SPS that sets $\eta_{\max} = \eta_{k-1}$ at iteration $k$ ensuring that $\etak \leq \eta_{k-1}$. This is because using a potentially increasing step-size sequence along with momentum can make the optimization unstable and result in divergence. Using this step-size, we prove the following result.
\begin{restatable}{theorem}{restateThmSPSAmsgradmom}
\label{thm:amsgrad-sps-mom}
Under the same assumptions of~\cref{thm:adagrad-constant} and assuming (iv) non-decreasing preconditioners (v) bounded eigenvalues in the $[a_{\min}, a_{\max}]$ interval with $\kappa = \nicefrac{a_{\max}}{a_{\min}}$, \amsgrad/ with $\beta \in [0,1)$, conservative Armijo SPS with $c = \nicefrac{1 + \beta}{1 - \beta}$ and uniform averaging converges at a rate,
\aligns{
\Expect{f(\bar{\x}_T) - f^*}
	\leq \,
	\bigg(\frac{1+\beta}{1-\beta}\bigg)^{\!2}
	\frac{2 L_{\max} D^2 d \kappa}{  T} 
	+ \sigma^2.
}
\end{restatable}
The above result matches the convergence rate in~\cref{thm:amsgrad-constant-mom} but does not require knowledge of the smoothness constant $L_{\max}$ or the minimum eigenvalue of the preconditioner $a_{\min}$;
the step-size adapts to both these quantities. 
A similar convergence rate can be obtained with a conservative variant of Armijo SLS (\cref{app:amsgrad-mom-proofs}), although our proof technique only allow for a restricted range of $\beta$. 

\subsection{SGD with stochastic heavy-ball momentum}
When $\Ak = I_d$, the \amsgrad/ update is equivalent to the update for SGD with heavy-ball momentum~\citep{sebbouh2020convergence}. By setting $\Ak = I_d$ in~\cref{thm:amsgrad-constant-mom}, we recover an $O(1/T + \sigma^2)$ rate for constant step-size SGD with heavy-ball momentum, matching the result for smooth, convex functions in~\citet{sebbouh2020convergence}. Furthermore, setting $\Ak = I_d$ in~\cref{thm:amsgrad-sps-mom} and using SPS to set the step-size helps match this rate without requiring the knowledge of the Lipschitz constant. To the best of our knowledge, this is the first result for SGD with an adaptive step-size and stochastic heavy-ball momentum. This result also provides theoretical justification for the heuristic used for incorporating heavy-ball momentum for SLS of~\citet{vaswani2019painless}. 

\begin{figure*}[t]
    \centering
    \includegraphics[scale =0.27]{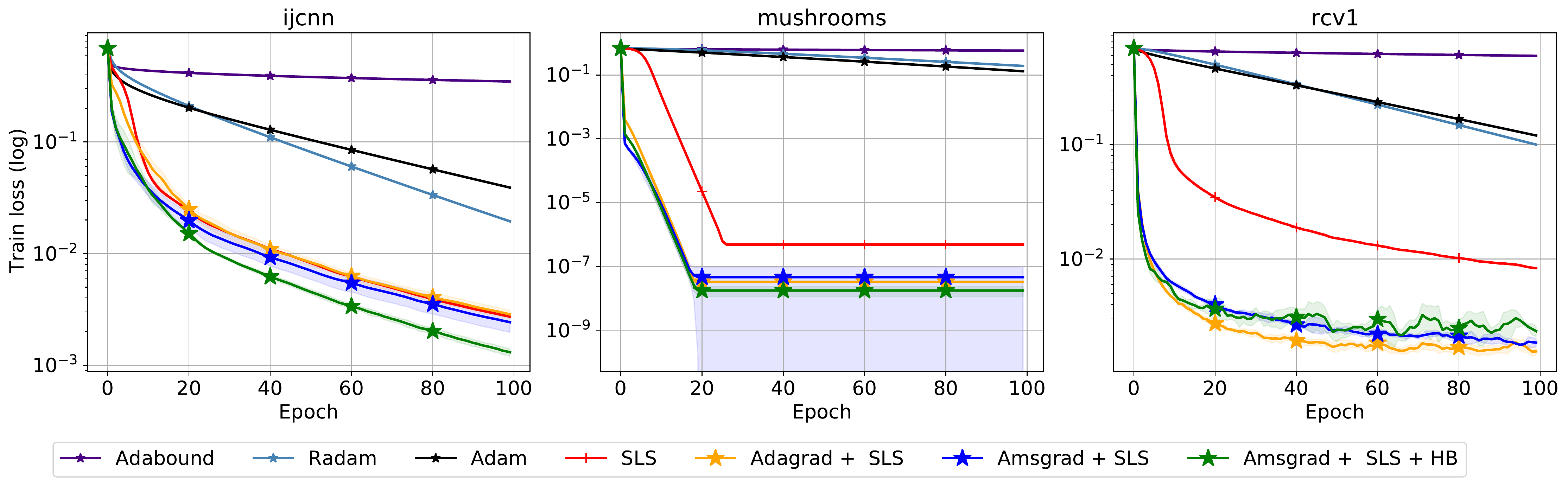}
    \caption{%
    Comparison on convex objectives: 
    binary classification on LIBSVM datasets using RBF kernel mappings,
    selected to (approximately) satisfy interpolation. 
    All optimizers use default step-sizes.
	Adam and its variants have poor performance due to the default step-size,
	while the stochastic line-search (SLS) adresses this issue and improves performance. 
    }
    \label{fig:kernels}
\end{figure*}

\subsection{Alternative stochastic heavy-ball momentum}
For a general preconditioner, the \amsgrad/ update in~\cref{eq:gen-update} is not equivalent to heavy-ball momentum. The standard stochastic heavy-ball update~\citep{loizou2017linearly} is:
\begin{align}
\xkk = \xk - \alpha_{k} \, \smallAkInverse \gradk{\xk} + \gamma \left(\xk - \xkm \right).
\label{eq:hb}
\end{align}
where $\alpha_k$ is the step-size and $\gamma \in [0,1)$ is the constant momentum parameter. Unlike this update, \amsgrad/ also preconditions the momentum direction $\paren{\xk - \xkm}$ (refer to~\cref{app:amsgrad-momentum-equivalence} for a relation between the two updates). If we consider the zero-momentum variant of  adaptive gradient methods as preconditioned gradient descent,~\cref{eq:hb} is a more natural way to incorporate momentum. We explore this alternate method and prove the same $O(1/T + \sigma^2)$ convergence rate for constant step-size, conservative Armijo SPS and Armijo SLS techniques in~\cref{app:amsgrad-hb-proofs}. We compare the two forms of heavy-ball momentum in~\cref{sec:experiments}. 

\section{Experimental evaluation}
\label{sec:experiments}

\begin{figure*}[t]
    \centering
    \subfigure{
    \hskip-.3em\includegraphics[scale=0.32]{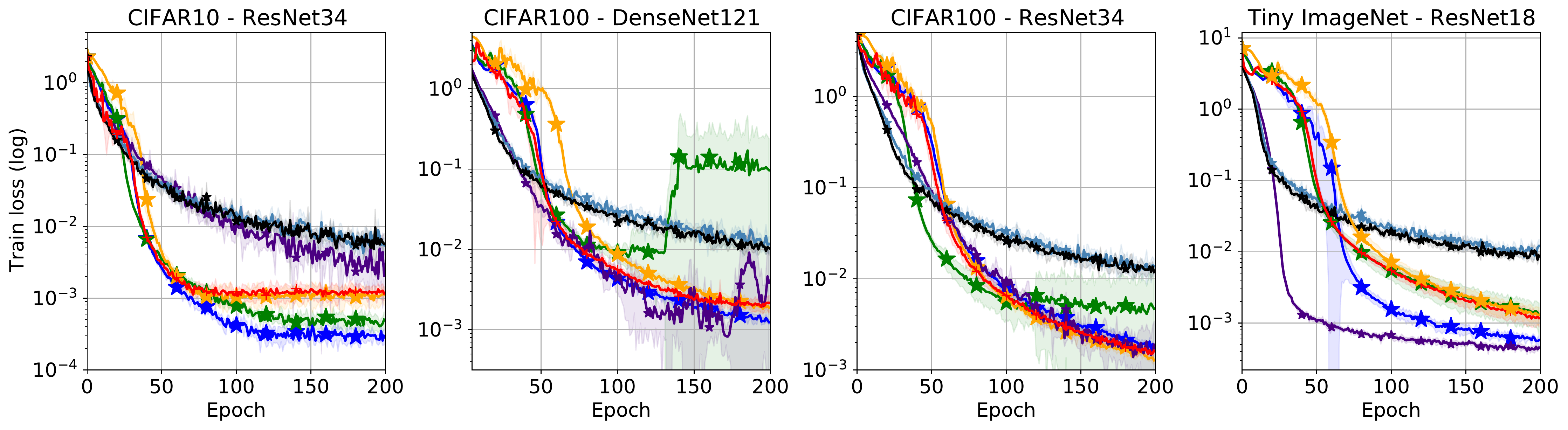}
    }\\[-.5em]%
    \subfigure{
    \includegraphics[scale=0.32]{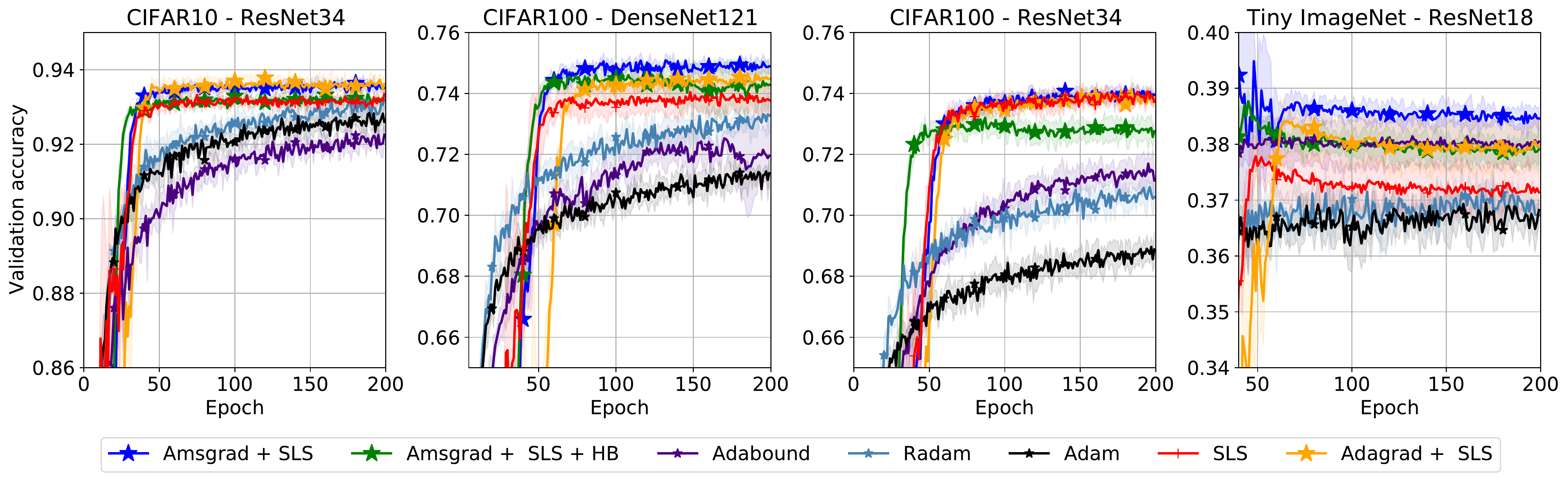}
    }
    \caption{Comparing optimizers for multi-class classification with deep networks. Training loss (top) and validation accuracy (bottom) for CIFAR-10, CIFAR-100 and Tiny ImageNet.
    \adagrad/ and \amsgrad/ equipped with the Armijo SLS not only converge faster than Adam and Radam but also have considerably better test performance. 
    }
\label{fig:deep}    
\end{figure*}

\paragraph*{Synthetic experiments}
We verify the effectiveness of the proposed line-search variants on the logistic regression example considered in~\cref{fig:syn}. For \adagrad/, we compare against the proposed Lipschitz line-search and Armijo SLS variants. As suggested by the theory, for each of these variants, we set the value of $c\m!=\m!\nicefrac{1}{2}$. For \amsgrad/, we compare against the variant employing the Armijo SLS with $c\m!=\m!\nicefrac{1}{2}$.\footnote{This corresponds to the largest allowable step-size in~\cref{thm:amsgrad-linesearch} \emph{without momentum}. Unfortunately, the values of $c$ suggested by the momentum analysis (\cref{thm:amsgrad-sps-mom}) are too conservative.} and use the default (in \pytorch/) momentum parameter of $\beta = 0.9$. In~\cref{fig:syn}, we observe that the line-search variants have good performance across margins, often better than the best-performing constant step-size. In~\cref{app:additional-exps}, we use  synthetic deep matrix factorization to systematically study the effect of over-parameterization for ~\citep{rolinek2018l4, vaswani2019painless}. Our results indicate that over-parameterization improves the convergence of all methods, but a line-search is essential to fully exploit it. 

\vspace{-2ex}
\paragraph*{Experiments with real data:} For experiments with real data, we use a batch-size of $128$ and compare against \adam/ and its improved variants; \radam/~\citep{liu2019variance} and \adabound/~\citep{luo2019adaptive}. To see the effect of preconditioning, we compare against SGD with SLS~\citep{vaswani2019fast} and SPS~\citep{loizou2020stochastic}. We find that SGD with SLS is more stable and has consistently better test performance than SPS. Hence, we only show results for SLS. Similar to~\citet{vaswani2019fast}, we observed that tuned constant step-size SGD is consistently outperformed by SGD with SLS, and do not show the corresponding plots. 

For the proposed methods, we consider the combinations with theoretical guarantees in the convex setting, specifically \adagrad/ and \amsgrad/ with the Armijo SLS. For \adagrad/, we only show Armijo SLS since it consistently outperforms the Lipschitz line-search. For all variants with Armijo SLS, we use $c\m!=\m!\nicefrac{1}{2}$ for all convex experiments (suggested by~\cref{thm:amsgrad-linesearch} and~\citet{vaswani2019painless}). Since we do not have a theoretical analysis for non-convex problems, we follow the protocol in~\cite{vaswani2019painless} and set $c\m!=\m!0.1$ for all the non-convex experiments. Throughout, we set $\beta\m!=\m!0.9$ for \amsgrad/. We also compare to the \amsgrad/ variant with heavy-ball momentum (with $\gamma\m!=\m!0.25$ found by grid-search). We refer to~\cref{app:exp-details} for a detailed discussion about the practical considerations and pseudocode for SLS and SPS. 

\textbf{Binary classification using RBF kernels:} We first consider convex minimization for a binary classification task using RBF kernels without regularization. The kernel bandwidths are chosen by cross-validation following the protocol in~\citep{vaswani2019painless}. This setup ensures interpolation is (approximately) satisfied in a convex setting. Following the protocol in~\citep{vaswani2019painless,loizou2020stochastic}, we experiment with standard datasets from LIBSVM~\cite{libsvm}: \textit{mushrooms}, \textit{rcv1} and \textit{ijcnn} and use the default parameters for all the optimizers. Figure \ref{fig:kernels} shows the training performance for the different methods using the logistic loss. We observe the (i) superior convergence of the optimizers using SLS including \adagrad/ and \amsgrad/ with both types of momentum. (ii) Adam and its variants have poor performance, completely stalling for the \textbf{mushrooms} dataset. (iii) The \adagrad/ and \amsgrad/ variants have better convergence than SGD with SLS demonstrating the positive effects of their preconditioning. 

\textbf{Multi-class classification using deep neural networks:} Following the protocol in~\citep{luo2019adaptive, vaswani2019painless, loizou2020stochastic}, we consider training standard neural network architectures for multi-class classification on CIFAR-10, CIFAR-100 and variants of the ImageNet datasets. We show a subset of results for CIFAR-10, CIFAR-100 and Tiny ImageNet and defer the rest to~\cref{app:additional-exps}. We compare against tuned \adam/ with its step-size found by a grid-search for each experiment.

From~\cref{fig:deep} we observe that, (i) in terms of generalization, \adagrad/ and \amsgrad/ with Armijo SLS have consistently the best performance, while SGD with SLS is often competitive. (ii) the \adagrad/ and \amsgrad/ variants not only converge faster than Adam and Radam but also have considerably better test performance. AdaBound has comparable convergence in training loss, but does not generalize as well. (iii) \amsgrad/ momentum is consistently better than the heavy-ball (HB) variant. Moreover, we observed that HB momentum was quite sensitive to the setting of $\gamma$, whereas \amsgrad/ is robust to $\beta$. In~\cref{app:additional-exps}, we include ablation results for \amsgrad/ with Armijo SLS \emph{without} momentum, and conclude that momentum does indeed improve the performance. We also plot the wall-clock time for the SLS variants and verify that the performance gains justify the increase in wall-clock time. We show the variation of the step-size across epochs and  observe a warm-up phase where the step-size increases followed by a constant or decreasing step-size~\citep{goyal2017accurate}. In~\cref{app:additional-exps}, we show that similar trends hold for different datasets and models. 

Our results indicate that simply setting the correct step-size on the fly can lead to substantial empirical gains, often more than those obtained by designing a different preconditioner. Furthermore, we see that with an appropriate step-size adaptation, adaptive gradient methods can generalize better than SGD. By disentangling the effect of the step-size from the preconditioner, we observe that \adagrad/ has good empirical performance, contradicting common knowledge~\citep{kingma2014adam}. Moreover, our techniques are orthogonal to designing better preconditioners and can be used with other adaptive gradient or even second-order methods.

\vspace{-2.5ex}
\section{Discussion}
\label{sec:conclusion}
\vspace{-1ex}
When training over-parameterized models in the interpolation setting, we showed that for smooth, convex functions, constant step-size variants of both \adagrad/ and \amsgrad/ are guaranteed to converge to the minimizer at $\bigO(1/T)$ rates. We proposed to use stochastic line-search techniques to help these methods adapt to the function's local smoothness, alleviating the need to tune their step-size and resulting in consistent empirical improvements across tasks. Although adaptive gradient methods outperform SGD in practice, their convergence rates are worse than constant step-size SGD and we hope to address this discrepancy in the future. 
\section{Acknowledgments}
We would like to thank Reza Babanezhad, Aaron Mishkin, Nicolas Loizou and Nicolas Le Roux for helpful discussions. This research was partially supported by the Canada CIFAR AI Chair Program, a Google Focused Research award, an IVADO postdoctoral scholarship, and by the NSERC Discovery Grants RGPIN-2017-06936 and 2015-06068. Simon Lacoste-Julien is a CIFAR Associate Fellow in the Learning in Machines \& Brains program.

\bibliographystyle{icml2021}
\bibliography{ref}

\newpage
\onecolumn
\appendix
\newcommand{\appendixTitle}{%
\vbox{
    \centering
    \vskip 0.05in
	{\LARGE \bf Supplementary material}
	\vskip 0.2in
	\hrule height 1pt 
}}

\newcommand{\makeappendixtable}[1]{%
~\\[.75em]\hskip 2em \begin{tabular}{@{}p{0.44\textwidth}p{0.2\textwidth}p{0.14\textwidth}@{}}
\toprule
#1\\
\bottomrule
\end{tabular}\\[.0em]
}
\newcommand{\makeappendixheader}{%
\textbf{Step-size} & \textbf{Rate} & \textbf{Reference} \\
\midrule
}
\appendixTitle

\section*{Organization of the Appendix}
\begin{itemize}
    \item[\ref{sec:setup}] \nameref{sec:setup}
    \item[\ref{app:line-search}] \nameref{app:line-search}
    \item[\ref{app:adagrad-proofs}] \nameref{app:adagrad-proofs}
    \makeappendixtable{
        \makeappendixheader
        Constant 
        & $\bigO(\nicefrac{1}{T} + \nicefrac{\sigma}{\m!\sqrt{T}})$     
        & \cref{thm:adagrad-constant}      
        \\
        Conservative Lipschitz LS    
        & $\bigO(\nicefrac{1}{T} + \nicefrac{\sigma}{\m!\sqrt{T}})$       
        & \cref{thm:adagrad-linesearch-conservative}      
        \\
        Non-conservative LS (with interpolation)
        &$\bigO(\nicefrac{1}{T})$
        & \cref{cor:adagrad-armijo-interpolation}        
    }
    \item[\ref{app:amsgrad-proofs}] \nameref{app:amsgrad-proofs}
    \makeappendixtable{
        Constant 
        & $\bigO(\nicefrac{1}{T} + \sigma^2)$ 
        & \cref{thm:amsgrad-constant}       
        \\
        Armijo LS 
        & $\bigO(\nicefrac{1}{T} + \sigma^2)$
        & \cref{thm:amsgrad-linesearch}      
    }
    \item[\ref{app:amsgrad-momentum-proofs}] \nameref{app:amsgrad-momentum-proofs}
    \makeappendixtable{
        Constant 
        & $\bigO(\nicefrac{1}{T} + \sigma^2)$ 
        & \cref{thm:amsgrad-constant-mom} 
        \\
        Conservative Armijo LS 
        & $\bigO(\nicefrac{1}{T} + \sigma^2)$
        & \cref{thm:amsgrad-sls-mom}
        \\
        Conservative Armijo SPS
        & $\bigO(\nicefrac{1}{T} + \sigma^2)$
        & \cref{thm:amsgrad-sps-mom}
    }
    ~\\[.5em]
    \nameref{app:amsgrad-hb-proofs}
        \makeappendixtable{
        Constant 
        & $\bigO(\nicefrac{1}{T} + \sigma^2)$
        & \cref{thm:amsgrad-constant-HB}
        \\
        Conservative Armijo LS
        & $\bigO(\nicefrac{1}{T} + \sigma^2)$
        & \cref{thm:amsgrad-sls-HB}
        \\
        Conservative Armijo SPS
        & $\bigO(\nicefrac{1}{T} + \sigma^2)$
        & \cref{thm:amsgrad-sps-HB}
    }
    \item[\ref{app:exp-details}] \nameref{app:exp-details}
    \item[\ref{app:additional-exps}] \nameref{app:additional-exps}
\end{itemize}



\clearpage

\clearpage

\section{Setup and assumptions}
\label{sec:setup}

\begin{table}[t]
    \centering
    \caption{Summary of notation}
    \label{table:notation-recap}
    \vskip -2em
    \null\hfill\begin{minipage}[t]{.45\textwidth}
    \centering~\\
    \begin{tabular}{ll}
        \toprule
        {\bf Concept}                            & {\bf Symbol} 
        \\ \midrule
        Iteration counter, maximum         & $k$, $T$
        \\ Iterates, minimum                     & $\x_k, \x^*$
        \\ Step-size                             & $\eta_k$
        \\ Function value, minimum               & $f(\x), f^*$
        \\ Stoch. function value, minimum    & $f_i(\x), f_i^*$
        ~\vphantom{Variance}\\
        \\ \bottomrule
    \end{tabular}%
    \end{minipage}\hspace{2.5em}%
    \begin{minipage}[t]{.45\textwidth}
    \centering~\\
    \begin{tabular}{ll}
        \toprule
        {\bf Concept}              & {\bf Symbol} 
        \\ \midrule
        General preconditioner  & $A_k$
        \\ Preconditioner bounds   & $[a_{\min}, a_{\max}]$
        \\ Maximum smoothness      & $L_{\max}$
        \\ Dimensionality          & $d$
        \\ Diameter bound          & $D$
        \\ Variance                & $\sigma^{\raisebox{-1pt}{\tiny $2$}} = \E_i[f_i(\xopt) - f^*_i]$
        \\ \bottomrule
    \end{tabular}
    \end{minipage}\hfill\null\\[1em]
\end{table}

We restate the main notation in \cref{table:notation-recap}. We now restate the main assumptions required for our theoretical results

We assume our  objective $f:\R^d\rightarrow\R$ has a finite-sum structure,
\begin{equation}
    f(\x) = \frac{1}{n}\sum_{i=1}^n f_i(\x),
\end{equation}
and analyze the following update, with $i_k$ selected uniformly at random,
\begin{equation}
    \label{eq:appendix-update-rule}
    \xkk = \xk - \etak \,  \smallAkInverse \mk  \quad \text{;} \quad \mk = \beta  \mkm + (1 - \beta) \gradk{\xk}
    \tag{Update rule}
\end{equation}
where $\etak$ is either a pre-specified constant or selected on the fly. We consider \adagrad/ and \amsgrad/ and use the fact that the preconditioners are non-decreasing i.e. $A_k \succeq A_{k-1}$. For \adagrad/, $\beta = 0$. For \amsgrad/, we further assume that the preconditioners remain bounded 
with eigenvalues in the range $[a_{\min}, a_{\max}]$,
\aligns{
    a_{\min} I \preceq A_k \preceq a_{\max} I.
    \tag{Bounded preconditioner}
    \label{eq:bounded-preconditioner}
}
For all algorithms, 
we assume that the iterates do not diverge and remain in  a ball of radius $D$, 
as is standard in the literature on online learning~\citep{duchi2011adaptive,levy2018online} and adaptive gradient methods~\citep{reddi2019convergence},
\aligns{
    \norm{\xk - \xopt} \leq D.
    \tag{Bounded iterates}
    \label{eq:bounded-iterates}
}
Our main assumptions are that each individual function $f_i$ 
is convex, differentiable, has a finite minimum $f_i^*$, and is $L_i$-smooth, meaning that for all $v$ and $w$, 
\aligns{
    f_i(v) &\geq f_i(w) - \lin{\nabla f_i(w), w-v},
    \tag{Individual Convexity}
    \label{eq:individual-convexity}
    \\
    f_i(v) & \leq f_i(w) + \inner{\nabla f_i(w)}{v - w} + \frac{L_i}{2} \normsq{v - w},
    \tag{Individual Smoothness}
    \label{eq:individual-smoothness}
}
which also implies that $f$ is convex and $L_{\max}$-smooth, where $L_{\max}$ 
is the maximum smoothness constant of the individual functions.
A consequence of smoothness is the following bound on the norm of the gradient stochastic gradients,
\aligns{
    \norm{\nabla f_i(\x)}^2 
    \leq
    2 L_{\max} (f_i(\x) - f_i^*).
}
To characterize interpolation, we define the 
expected difference between the minimum of $f$, $f(\xopt)$, 
and the minimum of the individual functions $f_i^*$,
\aligns{
    \sigma^2 &= \Expect[i]{f_i(\xopt) - f_i^*} < \infty.
    \tag{Noise}
}
When interpolation is exactly satisfied, every data point can be fit exactly, such that $f_i^* = 0$ and $f(\xopt) = 0$, we have $\sigma^2 = 0$.

\clearpage

\section{Line-search and Polyak step-sizes}
\label{app:line-search}
We now give the main guarantees on the step-sizes returned by the line-search. In practice, we use a backtracking line-search to find a step-size that satisfies the constraints, described in \cref{alg:conservative-sls} (\cref{app:exp-details}).
For simplicity of presentation, here we assume the line-search returns the largest step-size that satisfies the constraints.

When interpolation is not exactly satisfied, the procedures 
need to be equipped with an additional safety mechanism;
either by capping the maximum step-size by some $\eta_{\max}$
or by ensuring non-increasing step-sizes, $\etak \leq \etakm$. In this case, $\eta_{\max}$ ensures that a bad iteration of the line-search procedure does not result in divergence. When interpolation is satisfied, those conditions can be dropped (e.g., setting $\eta_{\max} \to \infty$) and the rate does not depend on $\eta_{\max}$. The line-searches depend on a parameter $c \in (0,1)$ that controls 
how much decrease is necessary to accept a step (larger $c$ means more decrease is demanded).

Assuming the Lipschitz and Armijo line-searches select the largest $\eta$ such that
\aligns{
    \fj(\x - \eta \nabla \fj(\x)) 
    &\leq \fj(\x) - c \eta \norm{\nabla \fj(\x)}^2,
    &&
    \eta \leq \eta_{\max},
    \tag{Lipschitz line-search}
    \label{eq:app-lipschitz-ls}
    \\
    \fj(\x - \eta A^{-1} \nabla \fj(\x)) 
    &\leq \fj(\x) - c \eta \norm{\nabla \fj(\x)}^2_{A^{-1}},
    &&
    \eta \leq \eta_{\max},
    \tag{Armijo line-search}
    \label{eq:app-armijo-ls}
}
the following lemma holds.
\begin{thmbox}
\begin{lemma}[Line-search]
\label{lem:lipschitz-line-search}
\label{lem:armijo-line-search}
If $f_i$ is $L_i$-smooth, 
the Lipschitz and Armijo lines-searches ensure
\aligns{
    \eta \norm{\nabla f_i(w)}^2 \leq \frac{1}{c}(f_i(w) - f_i^*),
    &&\text{and}&&
    \min\left\{
        \eta_{\max},
        \frac{2 \, (1-c)}{L_{i}}
    \right\}
    \leq \eta \leq \eta_{\max},
    \\
    \eta \norm{\nabla f_i(w)}^2_{A^{-1}} \leq \frac{1}{c}(f_i(w) - f_i^*),
    &&\text{and}&&
    \min\left\{\eta_{\max}, \frac{2 \, \lambda_{\min}(A) \,  (1-c)}{L_{i}}\right\}
    \leq
    \eta \leq \eta_{\max}.
}
\end{lemma}
\end{thmbox}
We do not include the backtracking line-search parameters in the analysis for simplicity, 
as the same bounds hold, up to some constant. 
With a backtracking line-search, 
we start with a large enough candidate step-size and multiply it 
by some constant $\gamma < 1$ until the Lipschitz or Armijo line-search condition is satisfied.
If $\eta'$ was a proposal step-size that did not satisfy the constraint, but $\gamma \eta'$ does, the maximum step-size $\eta$ that satisfies the constraint must be in the range $\gamma \eta' \leq \eta < \eta'$.

\begin{proof}[Proof of \cref{lem:lipschitz-line-search,lem:armijo-line-search}]
Recall that if $f_i$ is $L_i$-smooth, 
then for an arbitrary direction $d$,
\aligns{
    f_i(w - d) \leq f_i(w) - \lin{\nabla f_i(w), d} + \frac{L_i}{2} \norm{d}^2.
}
For the Lipschitz line-search, $d=\eta\nabla f_i(w)$.
The smoothness and the line-search condition are then
\aligns{
    \text{Smoothness:}
    &&
    f_i(w - \eta \nabla f_i(w)) - f_i(w)
    &\leq \, \textstyle \paren{\frac{L_i}{2}\eta^2-\eta}\norm{\nabla f_i(w)}^2,
    \\
    \text{Line-search:}
    &&
    f_i(w - \eta \nabla f_i(w)) - f_i(w) &\leq -c \eta \norm{\nabla f_i(w)}^2.
}
\vskip .5em
\begin{minipage}[t]{.48\textwidth}
As illustrated in \cref{fig:line-search-illustration},
the line-search condition is looser than smoothness if 
\aligns{
    \textstyle
    \paren{\frac{L_i}{2}\eta^2-\eta}\norm{\nabla f_i(w)}^2 
    &\leq 
    -c \eta \norm{\nabla f_i(w)}^2.
}
The inequality is satisfied for any $\eta \in [a, b]$, 
where $a, b$ are values of $\eta$ that satisfy
the equation with equality,
$a = 0, b = \nicefrac{2(1-c)}{L_i}$,
and the line-search condition holds for $\eta \leq \nicefrac{2(1-c)}{L_i}$.
\end{minipage}\hfill%
\begin{minipage}[t]{.48\textwidth}
\vskip-1.8em
\begin{figure}[H]
\centering
\makebox[\textwidth][c]{%
    \adjustbox{trim={.0\width} {.0\height} {.0\width} {.1\height},clip}{%
        \scalebox{0.8}{{\definecolor{mblue}{RGB}{0,68,136}
\definecolor{mred}{RGB}{187,85,102}
\definecolor{myellow}{RGB}{221,170,51}

\begin{tikzpicture}

	\node[anchor=west,color=mblue] at (1.75,2) {Smoothness: };
	\node[anchor=west,color=mblue] at (1.75,1.5) {$f_i(w) + (\frac{L_i}{2}\eta^2 - \eta)\Vert\nabla f_i(w)\Vert^2$};
	\node[anchor=west,color=mred] at (1.75,0.5) {Line search:};
	\node[anchor=west,color=mred] at (1.75,0) {$f_i(w) - c \eta \Vert \nabla f_i(w)\Vert^2$};

	\coordinate (v3) at (-1.29,-0.2) {} {} {};
	\coordinate  (v4) at (0.79,-0.2) {} {} {};
	\coordinate  (v6) at (-1.29,-0.4) {} {} {};
	\coordinate  (v5) at (-1.29,1.65) {} {} {};
	\coordinate  (v8) at (0.79,-0.4) {} {} {};
	\coordinate  (v7) at (0.79,0.6) {} {};

	\draw  (v3) edge (v4);
	\draw  (v5) edge (v6);
	\node at (-1.29,-0.7) {$\eta = 0$};
	\node at (0.79,-0.7) {$\eta = \frac{2(1-c)}{L_i}$};
	\draw  (v7) edge (v8);

	\draw[color=mblue,scale=1,domain=-1.6:1.6,smooth,variable=\x,line width=2] plot ({\x},{\x*\x});
	\node (v1) at (-1.8,1.9) {};
	\node (v2) at (1.7,0.15) {};
	\draw[color=mred,line width=2] (v1) edge (v2);

	\node at (-0.8,1.9) {$f_i(w)$};
	\node at (-1.29,1.65) {$\bullet$};
\end{tikzpicture}}}
    }
}
\vskip-1.00em
\caption{Sketch of the line-search inequalities.}
\label{fig:line-search-illustration}
\end{figure}
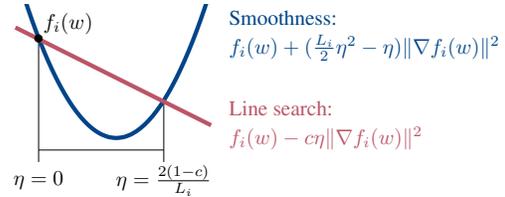
\end{minipage}
\vskip .5em

As the line-search selects the largest feasible step-size,
$\eta \geq \nicefrac{2(1-c)}{L_i}$.
If the step-size is capped at $\eta_{\max}$, we have
$\eta \geq \min\{\eta_{\max}, \nicefrac{2(1-c)}{L_i}\}$, 
and the proof for the Lipschitz line-search is complete. 
The proof for the Armijo line-search is identical except for the smoothness property, 
which is modified to use the $\norm{\cdot}_{A}$-norm for the direction
 $d=\eta A^{-1}\nabla f_i(w)$;
\aligns{
    f_i(w - \eta A^{-1} \nabla f_i(w))
    &\leq
    f_i(w) 
    - \eta \lin{\nabla f_i(w), A^{-1} \nabla f_i(w)}
    + \frac{L_i}{2}\eta^2 \norm{A^{-1} \nabla f_i(w)}^2,
    \\
    &\leq
    f_i(w)
    - \eta \norm{\nabla f_i(w)}_{A^{-1}}^2
    + \frac{L_i}{2 \lambda_{\min}(A)} \eta^2 \norm{\nabla f_i(w)}^2_{A^{-1}},
    \\
    &=
    f_i(w)
    + \paren{\frac{L_i}{2 \lambda_{\min}(A)}\eta^2 - \eta}\norm{\nabla f_i(w)}_{A^{-1}}^2,
}
where the second inequality comes from $\|A^{-1} \nabla f_i(w)\|{}^2 \leq \frac{1}{\lambda_{\min}(A)} \norm{\nabla f_i(w)}^2_{A^{-1}}$.
\end{proof}

Similarly, the stochastic Polyak step-sizes (SPS) for $f_i$ at $\x$ are defined as
\aligns{
    \text{SPS:}
    &&
    \eta
    =
    \min\left\{
        \frac{f_i(\x) - f_i^*}{c \normsq{\nabla f_i(\x)}},
        \eta_{\max}
    \right\},
    &&
    \text{Armijo SPS:}
    &&
    \eta
    =
    \min\left\{
        \frac{f_i(\x) - f_i^*}{c \normsq{\nabla f_i(\x)}_{A^{-1}}},
        \eta_{\max}
    \right\},
}
where the parameter $c > 0$ controls the scaling of the step 
(larger $c$ means smaller steps).
\begin{thmbox}
\begin{lemma}[SPS guarantees]
\label{lem:sps-bounds}
If $f_i$ is $L_i$-smooth, SPS and Armijo SPS ensure that 
\aligns{
    \text{SPS:}
    &&
    \textstyle
    \eta \normsq{\gradi{w}} \leq \frac{1}{c} (f_i(w) - f_i^*),
    &&
    \textstyle
    \min\left\{\eta_{\max}, \frac{1}{2 c L_{i}}\right\} \leq \eta &\leq \eta_{\max},
    \\
    \text{Armijo SPS:}
    &&
    \textstyle
    \eta \normsq{\gradi{w}}_{A^{-1}} \leq \frac{1}{c} (f_i(w) - f_i^*),
    &&
    \textstyle \min\left\{\eta_{\max}, \frac{\lambda_{\min}(A)}{2 c L_{i}}\right\} \leq \eta &\leq \eta_{\max}
}
\end{lemma}
\end{thmbox}
\begin{proof}[Proof of \cref{lem:sps-bounds}]
The first guarantee follows directly from the definition of the step-size.
For SPS,
\aligns{
    \eta \norm{\nabla f_i(\x)}^2
    &=
    \min\left\{
        \frac{f_i(\x) - f_i^*}{c \normsq{\nabla f_i(\x)}},
        \eta_{\max}
    \right\}
    \norm{\nabla f_i(\x)}^2,
    \\
    &=
    \min\left\{
        \frac{f_i(\x) - f_i^*}{c},
        \eta_{\max}  \norm{\nabla f_i(\x)}^2 
    \right\}
    \leq
    \frac{1}{c}(f_i(\x) - f_i^\star).
}
The same inequalities hold for Armijo SPS with $\norm{\nabla f_i(\x)}^2_{A^{-1}}$.
To lower-bound the step-size, we use the $L_i$-smoothness of $f_i$, which implies $f_i(w) - f_i^* \geq \frac{1}{2 L_i} \norm{\gradi{w}}\!{}^2$. 
For SPS,
\aligns{
\frac{f_i(w) - f_i^*}{c \normsq{\gradi{w}}} 
\geq \frac{\frac{1}{2 L_i} \normsq{\gradi{w}}}{c \normsq{\gradi{w}}} 
= \frac{1}{2 c L_i}.
}
For Armijo SPS, we additionally use $\norm{\nabla f_i(\x)}^2_{A^{-1}} \leq \frac{1}{\lambda_{\min}(A)}\norm{\nabla f_i(\x)}^2$,
\aligns{
\frac{f_i(w) - f_i^*}{c \norm{\gradi{w}}^2_{A^{-1}}} 
\geq \frac{\frac{1}{2 L_i} \normsq{\gradi{w}}}{c \frac{1}{\lambda_{\min}(A)} \norm{\gradi{w}}^2} 
= \frac{\lambda_{\min}(A)}{2 c L_i}.
\tag*{\qedhere}
}
\end{proof}

\clearpage

\section{Proofs for \adagrad/}
\label{app:adagrad-proofs}
We now move to the proof of the convergence of \adagrad/
in the smooth setting with a constant step-size (\cref{thm:adagrad-constant}) 
and the conservative Lipschitz line-search (\cref{thm:adagrad-linesearch-conservative}).
We first give a rate for an arbitrary step-size $\etak$ in the range $[\eta_{\min}, \eta_{\max}]$, and derive the rates of \cref{thm:adagrad-constant,thm:adagrad-linesearch-conservative} 
by specializing the range to a constant step-size or line-search.

\begin{thmbox}
\begin{restatable}[\adagrad/ with non-increasing step-sizes]{proposition}{restatePropositionAdagrad}
\label{thm:adagrad-linesearch-decreasing}
Assuming (i) convexity and (ii) $L_{\max}$-smoothness of each $f_i$, and (iii) bounded iterates,
\adagrad/ with non-increasing ($\etak\m!\leq\m!\eta_{k-1}$), bounded step-sizes ($\etak \in [\eta_{\min}, \eta_{\max}]$),
and uniform averaging 
$\bar{\x}_T\m!=\m!\frac{1}{T}\sum_{k=1}^{\raisebox{-.5pt}{\tiny $T$}} \!\xk$,
converges at a rate
\aligns{
    \Expect{f(\bar{\x}_T) - f^*}
    \leq
    \frac{\alpha}{T} + \frac{\sqrt{\alpha}\sigma}{\sqrt{T}},
    \qquad 
    \text{where }
    \alpha = \frac{1}{2} \paren{\frac{D^2}{\eta_{\min}} + 2 \eta_{\max}}^{\!2} d L_{\max}.
}
\end{restatable}
\end{thmbox}
We first use the above result to prove 
\cref{thm:adagrad-constant,thm:adagrad-linesearch-conservative}.
The proof of \cref{thm:adagrad-constant} is immediate by 
plugging $\eta = \eta_{\min} = \eta_{\max}$ in \cref{thm:adagrad-linesearch-decreasing}.
We recall its statement;%
\myquote{\begin{thmbox}\restateThmConstantStepSizeAdagrad*\end{thmbox}}

For \cref{thm:adagrad-linesearch-conservative}, we use the properties of the conservative Lipschitz line-search.
We recall its statement;%
\myquote{\begin{thmbox}\restateThmLinesearchAdagrad*\end{thmbox}}
\begin{proof}[Proof of \cref{thm:adagrad-linesearch-conservative}]
Using Lemma~\ref{lem:lipschitz-line-search}, there is a step-size $\etak$ 
that satisfies the Lipschitz line-search with 
$\etak \geq \nicefrac{2 \, (1-c)}{L_{\max}}$. 
Setting $c = \nicefrac{1}{2}$ and using a maximum step-size $\eta_{\max}$, 
we have 
\aligns{
    \min\left\{\eta_{\max}, \frac{1}{L_{\max}}\right\}
    \leq \etak \leq \eta_{\max},
    &&\implies&&
    \frac{1}{\eta_{\min}} = \max\left\{\frac{1}{\eta_{\max}}, L_{\max}\right\}.
    \tag*{\qedhere}
}
\end{proof}

Before going into the proof of \cref{thm:adagrad-linesearch-decreasing},
we recall some standard lemmas from the adaptive gradient literature
(Theorem 7 \& Lemma 10 in \citep{duchi2011adaptive}, Lemma 5.15 \& 5.16 in \citep{hazan2016oco}), 
and a useful quadratic inequality \citep[Part of Theorem 4.2]{levy2018online}).
We include proofs in \cref{app:proof-adagrad-lemmas} for completeness.

\begin{thmbox}
\begin{restatable}{lemma}{restateLemmaTelescoping}
\label{lem:telescoping-distance}
If the preconditioners are non-decreasing ($A_{k}\m!\succeq\m! A_{k-1}$), 
the step-sizes are non-increasing ($\etak\m!\leq\m!\eta_{k-1}$),
and the iterates stay within a ball of radius $D$ of the minima,
\aligns{
    \textstyle
    \sum_{k=1}^T \norm{\xk - \xopt}^2_{\frac{1}{\eta_k}A_k - \frac{1}{\eta_{k-1}} A_{k-1}}\leq \frac{D^2}{\eta_T} \Tr{A_T}.
}
\end{restatable}
\end{thmbox}

\begin{thmbox}
\begin{restatable}{lemma}{restateLemmaAdagrad}
\label{lem:adagrad}
For \adagrad/, $A_k = \left[\sum_{i=1}^k \nabla \fk(\xk) \nabla \fk(\xk)^\top \right]^{1/2}$ and satisfies,
\aligns{
    \textstyle
    \sum_{k=1}^T \norm{\nabla \fk(\xk)}^2_{A_k^{-1}}\leq 2 \Tr{A_T}, \qquad \Tr{A_T} 
    \leq \sqrt{d \sum_{k=1}^T \norm{\nabla \fk(\xk)}^2}.
}
\end{restatable}
\end{thmbox}

\begin{thmbox}
\begin{restatable}{lemma}{restateLemmaQuadratic}
\label{lem:quadratic-inequality}
If $x^2 \leq a(x+b)$ for $a \geq 0$ and $b \geq 0$,
\aligns{
    x \leq \frac{1}{2}\paren{\sqrt{a^2 + 4ab} + a} \leq a + \sqrt{ab}.
}
\end{restatable}
\end{thmbox}

We now prove \cref{thm:adagrad-linesearch-decreasing}.

\begin{proof}[{Proof of \cref{thm:adagrad-linesearch-decreasing}}]
We first give an overview of the main steps.
Using the definition of the update rule, 
along with \cref{lem:telescoping-distance,lem:adagrad}, we will show that
\alignn{
    \label{eq:adagrad-linesearch-decreasing-1}
    \textstyle
    2 \sum_{k=1}^T \lin{\nabla \fk(\xk), \xk - \xopt}
    \leq
    \paren{\frac{D^2}{\eta_{\min}} + 2 \eta_{\max}} \Tr{A_T}.
}
Using the definition of $A_T$, individual smoothness and convexity, we then show that for a constant $a$, 
\alignn{
    \label{eq:adagrad-linesearch-decreasing-2}
    \textstyle
    \sum_{k=1}^T \Expect{f(\xk) - f^*}
    \leq
    a \Big(\Expect{\sqrt{\sum_{k=1}^T \fk(\xk) - \fk(\xopt)}} + T\sigma^2\Big),
}
Using the quadratic inequality (\cref{lem:quadratic-inequality}), averaging and using Jensen's inequality finishes the proof.

To derive \cref{eq:adagrad-linesearch-decreasing-1}, 
we start with the \ref{eq:appendix-update-rule},
measuring distances to $\xopt$ in the $\norm{\cdot}_{A_k}$ norm,
\aligns{
    \norm{\xkk-\xopt}^2_{A_k} 
    =
    \norm{\xk-\xopt}^2_{A_k} 
    -2 \etak \lin{\nabla \fk(\xk), \xk - \xopt}
    + \etak^2 \norm{\nabla \fk(\xk)}^2_{A_k^{-1}}.
}
Dividing by $\etak$, reorganizing the equation and summing across iterations yields
\aligns{
    2 \sum_{k=1}^T \lin{\nabla \fk(\xk), \xk - \xopt}
    &\leq
    \sum_{k=1}^T 
    \norm{\xk - \xopt}^2_{\paren{
        \frac{A_k}{\eta_k} - \frac{A_{k-1}}{\eta_{k-1}}
    }}
    +  \sum_{k=1}^T \eta_k \norm{\nabla \fk(\xk)}^2_{A_k^{-1}},
    \\
    &\leq
    \sum_{k=1}^T 
    \norm{\xk - \xopt}^2_{\paren{
        \frac{A_k}{\eta_k} - \frac{A_{k-1}}{\eta_{k-1}}
    }}
    + \eta_{\max} \sum_{k=1}^T \norm{\nabla \fk(\xk)}^2_{A_k^{-1}}.
\intertext{%
We use the Lemmas~\ref{lem:telescoping-distance},~\ref{lem:adagrad}
to bound the RHS by the trace of the last preconditioner,
}
    &\leq
    \frac{D^2}{\eta_T} \Tr{A_T} + 2 \eta_{\max} \Tr{A_T},
    \tag{\cref{lem:telescoping-distance,lem:adagrad}}
    \\
    &\leq
    \paren{\frac{D^2}{\eta_{\min}} + 2 \eta_{\max}} \Tr{A_T}.
    \tag{$\etak \geq \eta_{\min}$}
}
To derive \cref{eq:adagrad-linesearch-decreasing-2}, we bound the trace of $A_T$ using 
\cref{lem:adagrad} and \ref{eq:individual-smoothness},
\aligns{
    \Tr{A_T}
    &\leq
    \textstyle
    \sqrt{d} \sqrt{\sum_{k=1}^T \norm{\nabla \fk(\xk)}^2},
    \tag{\cref{lem:adagrad}, Trace bound}
    \\
    &\leq
    \textstyle
    \sqrt{2 d L_{\max}} \sqrt{\sum_{k=1}^T \fk(\xk) -  \fk^*}.
    \tag{\ref{eq:individual-smoothness}}
    \\
    &\leq
    \textstyle
    \sqrt{2 d L_{\max}} \sqrt{
        \sum_{k=1}^T \fk(\xk) - \fk(\xopt) + 
        \fk(\xopt) - \fk^*
    }
    \tag{$\pm \fk(\xopt)$}
}
Combining the above inequalities 
with 
$\delta_{i_k} = \fk(\xopt) - \fk^*$
and 
$a = \frac{1}{2}\paren*{\frac{D^2}{\eta_{\min}} + 2 \eta_{\max}} \sqrt{2 d L_{\max}}$,
\aligns{
    \textstyle
    \sum_{k=1}^T \lin{\nabla \fk(\xk), \xk - \xopt}
    \leq
    a \sqrt{\sum_{k=1}^T \fk(\xk) - \fk(\xopt) + \delta_{i_k}}.
}
Using \ref{eq:individual-convexity} and taking expectations, 
\aligns{
    \textstyle
    \sum_{k=1}^T \Expect{f(\xk) - f^*}
    &\leq
    \textstyle
    a \Expect{\sqrt{\sum_{k=1}^T \fk(\xk) - \fk(\xopt) + \delta_{i_k}}\,},
    \\
    &\leq 
    \textstyle
    a \sqrt{\Expect{\sum_{k=1}^T \fk(\xk) - \fk(\xopt) + \delta_{i_k}}}.
    \tag{Jensen's inequality}
}
Letting $\sigma^2 := \Expect[i]{\delta_i} = \Expect[i]{f_i(\xopt) - f_{i}^*}$ and taking the square on both sides yields
\aligns{
    \paren{\sum_{k=1}^T \Expect{f(\xk) - f^*}}^{\!\!2}
    \leq
    a^2 \paren{\Expect{\sum_{k=1}^T \fk(\xk) - \fk(\xopt)} + T \sigma^2}.
}
The quadratic bound (\cref{lem:quadratic-inequality}) 
$x^2 \leq \alpha(x+\beta)$ implies $x \leq \alpha + \sqrt{\alpha \beta}$, with 
\aligns{
    x = \sum_{k=1}^T \Expect{f(\xk) - f^*},
    &&
    \alpha = \frac{1}{2} \paren{D^2 \frac{1}{\eta_{\min}} + 2 \eta_{\max}}^{\!\!2} d L_{\max},
    &&
    \beta = T \sigma^2,
}
gives the first bound below. 
Averaging $\bar{\x}_T\m!=\m!\frac{1}{T}\!\sum_{k=1}^T \!\xk$ and using Jensen's inequality 
give the result;
\aligns{
    \sum_{k=1}^T \Expect{f(\xk) - f^*}
    \leq
    \alpha + \sqrt{\alpha \beta}
    && \implies &&
    \Expect{f(\bar{\x}_T ) - f^*}
    \leq
    \frac{\alpha}{T} + \frac{\sqrt{\alpha} \sigma}{\sqrt{T}}.
    \tag*{\qedhere}
}
\end{proof}

\newpage
\subsection{Proofs of adaptive gradient lemmas}
\label{app:proof-adagrad-lemmas}
For completeness, we give proofs for the lemmas used in the previous section.
We restate them here;
\myquote{\begin{thmbox}
\restateLemmaTelescoping*
\end{thmbox}
}
\begin{proof}[Proof of \cref{lem:telescoping-distance}]
Under the assumptions that $\Ak$ is non-decreasing and $\etak$ is non-increasing, 
$\frac{1}{\eta_k}A_k - \frac{1}{\eta_{k-1}} A_{k-1} \succeq 0$,
so we can use the \ref{eq:bounded-iterates} assumption to bound
\vskip-1.5em
\aligns{
    \textstyle
    \sum_{k=1}^T 
    \norm{\xk - \xopt}^2_{\frac{A_k}{\eta_k} - \frac{A_{k-1}}{\eta_{k-1}} } 
    &\leq 
    \textstyle
    \sum_{k=1}^T \lambda_{\max}\paren{\frac{A_k}{\eta_k} - \frac{A_{k-1}}{\eta_{k-1}}}\m!\norm{\xk - \xopt}^2
    \\
    &\leq 
    \textstyle
    D^2\!\sum_{k=1}^T\lambda_{\max}\paren{\frac{A_k}{\eta_k} - \frac{A_{k-1}}{\eta_{k-1}}}.
\intertext{We then upper-bound $\lambda_{\max}$ by the trace and use the linearity of the trace to telescope the sum,}
    \textstyle
    &\leq 
    \textstyle
    D^2 \sum_{k=1}^T\Tr{\frac{A_k}{\eta_k} - \frac{A_{k-1}}{\eta_{k-1}} }
    = D^2 \sum_{k=1}^T\Tr{\frac{A_k}{\eta_k}} - \Tr{ \frac{A_{k-1}}{\eta_{k-1}} },
    \\ 
    &= 
    \textstyle
    D^2 \paren{\Tr{\frac{A_T}{\eta_T}} - \Tr{\frac{A_0}{\eta_0}}}
    \leq D^2\frac{1}{\eta_T}\Tr{A_T} \tag*{\qedhere}.
}
\end{proof}
\vskip-1em
\myquoten{\begin{thmbox}
\restateLemmaAdagrad*
\end{thmbox}
}
\begin{proof}[Proof of \cref{lem:adagrad}]
For ease of notation, let $\hazank\m!\defeq\m!\sgradf{i_k}{\x_k}$.
By induction, starting with $T\m!=\m!1$,
\aligns{
    \normsq{\sgradf{i_1}{\x_1}}_{A_1^{-1}}  
    & = \nabla_1^\top A_1^{-1}\nabla_1
    = \Tr{\nabla_1^\top A_1^{-1}\nabla_1}  
    = \Tr{A_1^{-1}\nabla_1\nabla_1^\top},
    \tag{Cyclic property of trace} 
    \\ 
    &= \Tr{A_1^{-1}A_1^2} 
    = \Tr{A_1}.
    \tag{$A_1=(\nabla_1\nabla_1^\top)^{\nicefrac{1}{2}}$}
}
Suppose that it holds for $T\m!-\m!1$, $\sum_{k=1}^{T-1} \norm{\hazank}^2_{A_k^{-1}}\leq 2 \Tr{A_{T-1}}$.
We will show that it also holds for $T$. 
Using the definition of the preconditioner and the cyclic property of the trace,
\aligns{
    \hspace{-1em} \textstyle 
    \sum_{k=1}^T \norm{\nabla \fk(\xk)}^2_{A_k^{-1}} 
    &\leq 2\Tr{A_{T-1}} + \normsq{\nabla_T}_{A_T^{-1}} 
    \tag{Induction hypothesis}
    \\ 
    &= 2\Tr{(A_T^2-\nabla_T\nabla_T^\top)^{\nicefrac{1}{2}}} + \Tr{A_T^{-1}\nabla_T\nabla_T^\top} 
    \tag{AdaGrad update}
}
We then use the fact that for
any $X \succeq Y \succeq 0$, we have \citep[Lemma 8]{duchi2011adaptive}
\aligns{
    2\Tr{(X-Y)^{\nicefrac{1}{2}}}+\Tr{X^{-\nicefrac{1}{2}}Y}\leq 2\Tr{X^{\nicefrac{1}{2}}}.
}
As $X = A_T^2\succeq Y = \nabla_T\nabla_T^\top \succeq 0$, we can use the above inequality and the induction holds for $T$.

For the trace bound, recall that $A_T = G_T^{1/2}$ where $G_T = \sum_{i=1}^T \nabla \fk(\xk) \nabla \fk(\xk)^\top$. We use Jensen's inequality, 
\begin{tsmath}\aligns{
    \Tr{A_T} = \Tr{G_T^{\nicefrac{1}{2}}}
    &= \sum_{j=1}^d \sqrt{\lambda_j(G_T)}
    = d\bigg(\frac{1}{d}\sum_{j=1}^d \sqrt{\lambda_j(G_T)} \bigg),
    \\
    &\leq d\sqrt{\frac{1}{d}\sum_{j=1}^d\lambda_j(G_T)}
    = \sqrt{d}\sqrt{\Tr{G_T}}.
}\end{tsmath}
To finish the proof, we use the definition of $G_T$ and the linearity of the trace to get
\aligns{
    \textstyle 
    \sqrt{\Tr{G_T}}
    = \sqrt{\Tr{\sum_{k=1}^T\ggt}} 
    = \sqrt{\sum_{k=1}^T\Tr{\ggt}}
    = \sqrt{\sum_{k=1}^T\normsq{\hazank}} 
    \tag*{\qedhere}.
}
\end{proof}
\vskip-1em
\myquoten{\begin{thmbox}
\restateLemmaQuadratic*
\end{thmbox}
}
\begin{proof}[Proof of \cref{lem:quadratic-inequality}]
The starting point is the quadratic inequality
$x^2 - ax - ab \leq 0$.
Letting $r_1 \leq r_2$ be the roots of the quadratic, 
the inequality holds if $x \in [r_1, r_2]$. 
The upper bound is 
then given by using $\sqrt{a+b} \leq \sqrt{a} +\sqrt{b}$
\aligns{
    r_2 
    = \frac{a + \sqrt{a^2 + 4ab}}{2}
    \leq \frac{a + \sqrt{a^2} + \sqrt{4ab}}{2}
    = a + \sqrt{ab}. 
    \tag*{\qedhere}
}
\end{proof}

\subsection{Regret bound for \adagrad/ under interpolation}
\label{app:adagrad-regret}
In the online convex optimization framework, we consider a sequence of functions $f_k\vert_{k = 1}^{T}$, chosen potentially adversarially by the environment. The aim of the learner is to output a series of strategies $\xk\vert_{k = 1}^{T}$ \emph{before} seeing the function $f_k$. After choosing $\xk$, the learner suffers the loss $f_k(\xk)$ and observes the corresponding gradient vector $\nabla f_k(\xk)$. They suffer an instantaneous regret $r_k = f_k(\xk) - f_k(\x)$ compared to a fixed strategy $\x$. The aim is to bound the cumulative regret, 
\aligns{
R(T) = \fullsum [f_k(\xk) - f_k(\xopt)]
}
where $\xopt = \argmin \fullsum f_k(\x)$ is the best strategy if we had access to the entire sequence of functions in hindsight. Assuming the functions are convex but non-smooth, \adagrad/ obtains an $\bigO(1/\sqrt{T})$ regret bound~\citep{duchi2011adaptive}. For online convex optimization, the interpolation assumption implies that the learner model is powerful enough to fit the entire sequence of functions. For large over-parameterized models like neural networks, where the number of parameters is of the order of millions, this is a reasonable assumption for large $T$. 

We first recall the update of \adagrad/, at iteration $k$, the learner decides to play the strategy $\xk$, suffers loss $f_k(\xk)$ and uses the gradient feedback $\nabla f_k(\xk)$ to update their strategy as 
\aligns{\textstyle
\xkk = \xk - \eta A_k^{-1} \nabla f_k(\xk),
\quad \text{ where }
A_k = \left[\sum_{i=1}^k \nabla f_k(\xk) \nabla f_k(\xk)^\top \right]^{1/2}. 
}
Now we show that for smooth, convex functions under the interpolation assumption, \adagrad/ with a constant step-size can result in \emph{constant} regret. 
\begin{thmbox}
\begin{theorem}\label{thm:adagrad-regret}
For a sequence of $L_{\max}$-smooth, convex functions $f_k$, assuming the iterates remain bounded s.t. for all $k$, $\norm{\x_k - \xopt} \leq D$, \adagrad/ with a constant step-size $\eta$ achieves the following regret bound, 
\aligns{
R(T) & \leq \frac{1}{2} \paren{D^2 \frac{1}{\eta} + 2 \eta}^{\!\!2} d L_{\max} + \sqrt{ \frac{1}{2} \paren{D^2 \frac{1}{\eta} + 2 \eta}^{\!\!2} d L_{\max} \sigma^2} \; \sqrt{T} 
}
where $\sigma^2$ is an upper-bound on $f_k(\xopt) - f_k^*$.
\label{thm:adagrad-regret}
\end{theorem}
\end{thmbox}
Observe that $\sigma^2$ is the degree to which interpolation is violated, and if $\sigma^2 \neq 0$, $R(T) = \bigO(\sqrt{T})$ matching the regret of~\citep{duchi2011adaptive}. However, when interpolation is exactly satisfied, $\sigma^2 = 0$, and $R(T) = \bigO(1)$. 

\begin{proof}[Proof of~\cref{thm:adagrad-regret}]
The proof follows that of~\cref{thm:adagrad-linesearch-decreasing} which is inspired from~\citep{levy2018online}. For convenience, we repeat the basic steps. Measuring distances to $\xopt$ in the $\norm{\cdot}_{A_k}$ norm,
\aligns{
    \norm{\xkk-\xopt}^2_{A_k} 
    =
    \norm{\xk-\xopt}^2_{A_k} 
    -2 \eta \lin{\nabla f_k(\xk), \xk - \xopt}
    + \eta^2 \norm{\nabla f_k(\xk)}^2_{A_k^{-1}}.
}
Dividing by $2 \eta$, reorganizing the equation and summing across iterations yields
\aligns{
    \sum_{k=1}^T \lin{\nabla f_k(\xk), \xk - \xopt}
    &\leq
    \sum_{k=1}^T 
    \norm{\xk - \xopt}^2_{\paren{
        \frac{A_k}{2 \eta} - \frac{A_{k-1}}{2 \eta}
    }}
    +  \frac{\eta}{2} \sum_{k=1}^T  \norm{\nabla f_k(\xk)}^2_{A_k^{-1}}.
\intertext{By convexity of $f_k$, $\lin{\nabla f_k(\xk), \xk - \xopt} \geq f_k(\xk) - f_k(\xopt)$. Using the definition of regret,}
R(T) & \leq \sum_{k=1}^T 
    \norm{\xk - \xopt}^2_{\paren{
        \frac{A_k}{2 \eta} - \frac{A_{k-1}}{2 \eta}
    }}
    +  \frac{\eta}{2} \sum_{k=1}^T  \norm{\nabla f_k(\xk)}^2_{A_k^{-1}}.
}
We use the Lemmas~\ref{lem:telescoping-distance},~\ref{lem:adagrad} to bound the RHS by the trace of the last preconditioner,
\aligns{
    R(T) & \leq \left(\frac{D^2}{2 \eta} +  \eta \right) \Tr{A_T}.
}
We now bound the trace of $A_T$ using \cref{lem:adagrad} and \ref{eq:individual-smoothness},
\aligns{
    \Tr{A_T}
    &\leq
    \textstyle
    \sqrt{d} \sqrt{\sum_{k=1}^T \norm{\nabla f_k(\xk)}^2},
    \tag{\cref{lem:adagrad}, Trace bound}
    \\
    &\leq
    \textstyle
    \sqrt{2 d L_{\max}} \sqrt{\sum_{k=1}^T f_k(\xk) -  f_k^*},
    \tag{\ref{eq:individual-smoothness}}
    \\
    &\leq
    \textstyle
    \sqrt{2 d L_{\max}} \sqrt{
        \sum_{k=1}^T f_k(\xk) - f_k(\xopt) + 
        f_k(\xopt) - f_k^*
    },
    \tag{$\pm f_k(\xopt)$}
    \\
    &\leq \sqrt{2 d L_{\max}} \sqrt{R(T) + \sigma^2 T}.
    \tag{Since $f_k(\xopt) - f_k^* \leq \sigma^2$}
}
Plugging this back into the regret bound, 
\aligns{
    R(T) 
    &\leq 
    \paren{\frac{D^2}{2\eta} + \eta} \sqrt{2 d L_{\max}} [\sqrt{R(T)  + \sigma^2 T}].
\intertext{Squaring both sides and denoting $a = \left(\frac{D^2}{2 \eta} + \eta \right) \sqrt{2 d L_{\max}}$,}
    [R(T)]^2 
    &\leq a^2 [R(T) + \sigma^2 T].
}
Using the quadratic bound (\cref{lem:quadratic-inequality}) 
$x^2 \leq \alpha(x+\beta)$ implies $x \leq \alpha + \sqrt{\alpha \beta}$, with 
\aligns{
    x = R(T),
    &&
    \alpha = \frac{1}{2} \paren{D^2 \frac{1}{\eta} + 2 \eta}^{\!\!2} d L_{\max},
    &&
    \beta = \sigma^2 T,
}
yields the bound, 
\aligns{
R(T) & \leq \alpha + \sqrt{\alpha \beta} = \frac{1}{2} \paren{D^2 \frac{1}{\eta} + 2 \eta}^{\!\!2} d L_{\max} + \sqrt{ 
\frac{1}{2} \paren{D^2 \frac{1}{\eta} + 2 \eta}^{\!\!2} d L_{\max} \sigma^2 T}.
\tag*{\qedhere}
} 
\end{proof}

\clearpage

\subsection{With interpolation, without conservative line-searches}
\label{app:adagrad-interpolation}
In this section, we show that the conservative constraint $\etakk \leq \etak$ is not necessary if interpolation is satisfied.
We give the proof for the Armijo line-search, that has better empirical performance, but a worse theoretical dependence on the problem's constants. For the theorem below, $a_{\min}$ is lower-bounded by $\epsilon$ in practice. A similar proof also works for the Lipschitz line-search. 
\begin{thmbox}
\begin{theorem}[\adagrad/ with Armijo line-search under interpolation]
\label{cor:adagrad-armijo-interpolation}
Under the same assumptions of \cref{thm:adagrad-linesearch-decreasing},
but without non-increasing step-sizes, if interpolation is satisfied,
\adagrad/ with the Armijo line-search and uniform averaging converges at the rate,
\aligns{
    \Expect{f(\bar{\x}_T) - f^*}
    \leq \frac{\paren{D^2 + 2\eta_{\max}^2}^2 d L_{\max}}{2T}
    \left(\max\left\{\frac{1}{\eta_{\max}}, \frac{L_{\max}}{a_{\min}} \right\}\right)^2.
}
where $a_{\min} = \min_{k} \{ \lambda_{\min}(A_k) \}$.
\end{theorem}
\end{thmbox}
\begin{proof}[Proof of Theorem~\ref{cor:adagrad-armijo-interpolation}]
Following the proof of \cref{thm:adagrad-linesearch-decreasing},
\aligns{
    2 \sum_{k=1}^T \etak
    \lin{\nabla \sfkk, \xk - \xopt}
    =
    \sum_{k=1}^T 
    \norm{\xk - \xopt}_{A_k}^2
    - \norm{\xkk-\xopt}_{A_k}^2
    + \etak^2 \norm{\nabla \sfkk}^2_{A_k^{-1}}.
}
On the left-hand side, 
we use individual convexity and interpolation, which implies $f_{i_k}(\xopt) = \min_\x f_{i_k}(\x)$ and we can bound $\etak$ by $\eta_{\min}$, giving
\aligns{
    \etak \lin{\nabla \sfkk, \xk - \xopt}
    \geq
    \etak \underbrace{\paren{\sfkk - f_{i_k}(\xopt)}}_{\geq 0}
    \geq
    \eta_{\min} \paren{\sfkk - f_{i_k}(\xopt)}.
}
On the right-hand side, 
we can apply the \adagrad/ lemmas (\cref{lem:adagrad})
\aligns{
    & \sum_{k=1}^T 
    \norm{\xk - \xopt}_{A_k}^2
    - \norm{\xkk-\xopt}_{A_k}^2
    + \eta_{\max}^2 \norm{\nabla \sfkk}^2_{A_k^{-1}},
    \\
    &\leq
    D^2 \Tr{A_T}
    + 2 \eta_{\max}^2 \Tr{A_T},
    \tag{By \cref{lem:telescoping-distance,lem:adagrad}}
    \\
    &\leq
    \textstyle
    \paren{D^2 + 2\eta_{\max}^2}
    \sqrt{d}
    \sqrt{\sum_{k=1}^T \norm{\nabla f_{i_k}(\xk)}^2},
    \tag{By the trace bound of \cref{lem:adagrad}}
    \\
    &\leq
    \textstyle
    \paren{D^2 + 2\eta_{\max}^2}
    \sqrt{2 d L_{\max}}
    \sqrt{\sum_{k=1}^T f_{i_k}(\xk) - f_{i_k}(\xopt)}.
    \tag{By~\ref{eq:individual-smoothness} and interpolation}
}

Defining $a = \frac{1}{2\eta_{\min}}\paren{D^2 + 2\eta_{\max}^2}\sqrt{2dL_{\max}}$
and combining the previous inequalities yields
\aligns{
    \sum_{k=1}^T \paren{f_{i_k}(\xk) - f_{i_k}(\xopt)}
    \leq
    \textstyle
    a \sqrt{\sum_{k=1}^T f_{i_k}(\xk) - f_{i_k}(\xopt)}.
}
Taking expectations and applying Jensen's inequality yields
\aligns{
    \textstyle
    \textstyle
    \sum_{k=1}^T \Expect{f(\xk) - f(\xopt)}
    \leq
    a \sqrt{\sum_{k=1}^T \Expect{f(\xk) - f(\xopt)}}.
}
Squaring both sides, dividing by $\sum_{k=1}^T \Expect{f(\xk) - f(\xopt)}$, followed by dividing by $T$ and applying Jensen's inequality,
\aligns{
    \Expect{f(\bar{\x}_T) - f(\xopt)}
    \leq 
    \frac{a^2}{T}
    =
    \frac{\paren{D^2 + 2\eta_{\max}^2}^2 d L_{\max}}{2\eta_{\min}^2 T}.
}
Using the Armijo line-search guarantee (\cref{lem:armijo-line-search})
with $c = \nicefrac{1}{2}$ and a maximum step-size $\eta_{\max}$, 
\aligns{
    \eta_{\min} = \min\left\{ \eta_{\max}, \frac{a_{\min}}{L_{\max}} \right\},
}
where $a_{\min} = \min_{k} \{ \lambda_{\min}(A_k) \}$, giving the rate
\aligns{
    \Expect{f(\bar{\x}_T) - f(\xopt)}
    \leq 
    \frac{\paren{D^2 + 2\eta_{\max}^2}^2 d L_{\max}}{2T}
    \left(\max\left\{\frac{1}{\eta_{\max}}, \frac{L_{\max}}{a_{\min}} \right\}\right)^2.
    \tag*{\qedhere}
}
\end{proof}

\clearpage

\section{Proofs for \amsgrad/ and non-decreasing preconditioners without momentum}
\label{app:amsgrad-proofs}

We now give the proofs for \amsgrad/ and general bounded, non-decreasing preconditioners in the smooth setting, 
using a constant step-size (\cref{thm:amsgrad-constant}) and the Armijo line-search (\cref{thm:amsgrad-linesearch}).
As in \cref{app:adagrad-proofs}, we prove a general proposition and specialize it for each of the theorems;
\begin{thmbox}
\begin{proposition}
\label{prop:amsgrad}
In addition to assumptions of \cref{thm:adagrad-constant}, assume
that (iv) the preconditioners are non-decreasing and have (v) bounded eigenvalues in the $[a_{\min}, a_{\max}]$ range. If the step-sizes are constrained to lie in the range $[\eta_{\min}, \eta_{\max}]$ and satisfy
\alignn{
    \etak \norm{\nabla \fk(\xk)}^2_{A_k^{-1}} \leq
    M \paren{\fk(\xk) - \fkopt},
    \quad \text{ for some $M < 2$},
    \label{eq:amsgrad-assumption-step-size}
}
using uniform averaging $\bar{\x}_T = \frac{1}{T}\sum_{k=1}^T \xk$ leads to the rate
\aligns{
    \Expect{f(\bar{\x}_T) - f^*}
    \leq 
    \frac{1}{T} \frac{D^2 d a_{\max}}{(2-M) \eta_{\min}} 
    + \paren{\frac{2}{2-M} \frac{\eta_{\max}}{\eta_{\min}} - 1} \sigma^2.
} 
\end{proposition}
\end{thmbox}

\begin{thmbox}
\begin{restatable}{theorem}{restateThmConstantAmsgrad}
\label{thm:amsgrad-constant}
Under the assumptions of~\cref{thm:adagrad-constant} and assuming (iv) non-decreasing preconditioners (v) bounded eigenvalues in the $[a_{\min}, a_{\max}]$ interval, \amsgrad/ with no momentum, constant step-size $\eta = \frac{a_{\min}}{2 L_{\max}}$ and uniform averaging converges at a rate,
\aligns{
	\Expect{f(\bar{w}_T) - f^*} & \leq \frac{2 D^2 d \, a_{\max} \, L_{\max}}{a_{\min} \, T} + \sigma^2.
}
\end{restatable}
\end{thmbox}
\begin{proof}[Proof of \cref{thm:amsgrad-constant}]
Using \ref{eq:bounded-preconditioner} and \ref{eq:individual-smoothness}, we have that 
\aligns{
    \norm{\nabla \fk(\xk)}^2_{A_k^{-1}}
    \leq
    \frac{1}{a_{\min}} \norm{\nabla \fk(\xk)}^2 
    \leq 
    \frac{2L_{\max}}{a_{\min}} (\fk(\xk) - \fkopt).
}
A constant step-size $\eta_{\max}\m!=\m!\eta_{\min}\m!=\m!\frac{a_{\min}}{2L_{\max}}$ satisfies the step-size assumption 
(Eq.~\ref{eq:amsgrad-assumption-step-size}) with $M\m!=\m!1$ and 
\aligns{
    \frac{1}{T} \frac{D^2 d a_{\max}}{(2-M) \eta_{\min}} 
    \,+\, \paren{\frac{2}{2-M}\frac{\eta_{\max}}{\eta_{\min}} - 1} \sigma^2
    =
    \frac{1}{T} \frac{2 L_{\max} D^2 d a_{\max}}{a_{\min}} 
    + \sigma^2.
    \tag*{\qedhere}
}
\end{proof}

\myquote{\begin{thmbox}
\begin{restatable}{theorem}{restateThmLinesearchAmsgrad}
\label{thm:amsgrad-linesearch}
Under the same assumptions as~\cref{thm:adagrad-constant}, \amsgrad/ with zero momentum, Armijo line-search with $c = \nicefrac{3}{4}$, a step-size upper bound $\eta_{\max}$ and uniform averaging converges at a rate,
\aligns{
	\Expect{f(\bar{\x}_T) - f^*} 
	& \leq 
	\paren{\frac{3D^2 d\cdot a_{\max}}{2T} + 3\eta_{\max} \sigma^2} \\ 
	& \times \max\left\{\frac{1}{\eta_{\max}}, \frac{2L_{\max}}{a_{\min}}\right\}.
}
\end{restatable}
\end{thmbox}
}
\begin{proof}[Proof of \cref{thm:amsgrad-linesearch}]
For the \ref{eq:app-armijo-ls}, \cref{lem:armijo-line-search} guarantees that
\aligns{
    \eta \norm{\nabla \fk(\xk)}^2_{\Ak^{-1}} \leq \frac{1}{c}(\fk(\xk) - \fk^*),
    &&\text{and}&&
    \min\left\{\eta_{\max}, \frac{2 \, \lambda_{\min}(\Ak) \,  (1-c)}{L_{\max}}\right\}
    \leq
    \eta \leq \eta_{\max}.
}
Selecting $c = \nicefrac{3}{4}$ gives $M = \nicefrac{4}{3}$ and 
$\eta_{\min} = \min\left\{\eta_{\max}, \frac{a_{\min}}{2L_{\max}}\right\}$, so 
\aligns{
    &\frac{1}{T} \frac{D^2 d a_{\max}}{(2-M) \eta_{\min}} 
    + \paren{\frac{2}{2-M} \frac{\eta_{\max}}{\eta_{\min}} - 1} \sigma^2
    \\
    &\quad= 
    \frac{1}{T} \frac{D^2 d a_{\max}}{(2-\nicefrac{4}{3}) \eta_{\min}} 
    + \paren{\frac{2}{2-\nicefrac{4}{3}}\frac{\eta_{\max}}{\eta_{\min}} - 1} \sigma^2,
    \\
    &\quad=
    \frac{1}{T} \frac{3 D^2 d a_{\max}}{2 \eta_{\min}} 
    + \paren{\frac{3\eta_{\max}}{\eta_{\min}} - 1} \sigma^2,
    \\
    &\quad\leq
    \frac{3 D^2 d a_{\max}}{2 T}
    \max\left\{\frac{1}{\eta_{\max}}, \frac{2L_{\max}}{a_{\min}}\right\}
    + 3 \eta_{\max} \sigma^2
    \max\left\{\frac{1}{\eta_{\max}}, \frac{2L_{\max}}{a_{\min}}\right\}\!.
    \tag*{\qedhere}
}
\end{proof}

\begin{thmbox}
\begin{restatable}{theorem}{restateThmSPSAmsgrad}
\label{thm:amsgrad-sps}
Under the assumptions of~\cref{thm:adagrad-constant} and assuming (iv) non-decreasing preconditioners (v) bounded eigenvalues in the $[a_{\min}, a_{\max}]$ interval, \amsgrad/ with no momentum, Armijo SPS with $c = \nicefrac{3}{4}$ and uniform averaging converges at a rate,
\aligns{
	\Expect{f(\bar{\x}_T) - f^*} 
	\leq 
	\paren{\frac{3D^2 d\cdot a_{\max}}{2T} + 3\eta_{\max} \sigma^2}
	\max\left\{\frac{1}{\eta_{\max}}, \frac{3L_{\max}}{2a_{\min}}\right\}.
}
\end{restatable}
\end{thmbox}
\begin{proof}[Proof of \cref{thm:amsgrad-sps-mom}]
For Armijo SPS, \cref{lem:sps-bounds} guarantees that
\aligns{
    \etak \norm{\nabla \fk(\xk)}^2_{\Ak^{-1}} \leq \frac{1}{c}(\fk(\xk) - \fk^*),
    &&\text{and}&&
    \min\left\{\eta_{\max}, \frac{a_{\min}}{2 c \, L_{\max}}\right\}
    \leq
    \eta \leq \eta_{\max}.
}
Selecting $c = \nicefrac{3}{4}$ gives $M = \nicefrac{4}{3}$ and 
$\eta_{\min} = \min\left\{\eta_{\max}, \frac{2 a_{\min}}{3 L_{\max}}\right\}$, so 
\aligns{
    &\frac{1}{T} \frac{D^2 d a_{\max}}{(2-M) \eta_{\min}} 
    + \paren{\frac{2}{2-M} \frac{\eta_{\max}}{\eta_{\min}} - 1} \sigma^2
    \\
    &\quad= 
    \frac{1}{T} \frac{D^2 d a_{\max}}{(2-\nicefrac{4}{3}) \eta_{\min}} 
    + \paren{\frac{2}{2-\nicefrac{4}{3}}\frac{\eta_{\max}}{\eta_{\min}} - 1} \sigma^2,
    \\
    &\quad=
    \frac{1}{T} \frac{3 D^2 d a_{\max}}{2 \eta_{\min}} 
    + \paren{\frac{3\eta_{\max}}{\eta_{\min}} - 1} \sigma^2,
    \\
    &\quad\leq
    \frac{3 D^2 d a_{\max}}{2 T}
    \max\left\{\frac{1}{\eta_{\max}}, \frac{3L_{\max}}{2a_{\min}}\right\}
    + 3 \eta_{\max} \sigma^2
    \max\left\{\frac{1}{\eta_{\max}}, \frac{3L_{\max}}{2 a_{\min}}\right\}\!.
    \tag*{\qedhere}
}
\end{proof}

Before diving into the proof of \cref{prop:amsgrad},
we prove the following lemma to handle terms of the form $\etak \paren{\fk(\xk) - \fk(\xopt)}$.
If $\etak$ depends on the function sampled at the current iteration, $\fk$, 
as in the case of line-search, we cannot take expectations as the terms are not independent.
\cref{lem:sps} bounds $\etak \paren{\fk(\xk) - \fk(\xopt)}$ in terms of the range $[\eta_{\min}, \eta_{\max}]$;

\begin{thmbox}
\begin{lemma}
\label{lem:sps}
If $0 \leq \eta_{\min} \leq \eta \leq \eta_{\max}$
and the minimum value of $f_i$ is $f_i^*$, then 
\aligns{
    \eta \paren{\sfik - \sfistar} 
    \geq
    \eta_{\min} \paren{\sfik - \sfistar} - (\eta_{\max} - \eta_{\min}) \paren{\sfistar - \sfimin}.
}
\end{lemma}
\end{thmbox}

\begin{proof}[Proof of \cref{lem:sps}]
    By adding and subtracting $f_i^*$, the minimum value of $f_i$, 
    we get a non-negative and a non-positive term multiplied by $\eta$. 
    We can use the bounds $\eta \geq \eta_{\min}$ and $\eta \leq \eta_{\max}$ separately;
\aligns{
    \eta [f_i(\x) - f_i(\xopt)]
    &=
    \eta [
        \underbrace{f_i(\x) - f_i^*}_{\geq 0} 
        + \underbrace{f_i^* - f_i(\xopt)}_{\leq 0}
    ],
    \\
    &\geq
    \eta_{\min} [f_i(\x) - f_i^*]
    + \eta_{\max} [f_i^* - f_i(\xopt)].
\intertext{Adding and subtracting $\eta_{\min} f_i(\xopt)$ finishes the proof,}
    &=
    \eta_{\min} [f_i(\x) - f_i(\xopt) + f_i(\xopt)- f_i^*]
    + \eta_{\max} [f_i^* - f_i(\xopt)],
    \\
    &=
    \eta_{\min} [f_i(\x) - f_i(\xopt)]
    + (\eta_{\max} - \eta_{\min}) [f_i^* - f_i(\xopt)].
    \qedhere
}
\end{proof}

\vskip 2em

\begin{proof}[Proof of \cref{prop:amsgrad}]
We start with the \ref{eq:appendix-update-rule},
measuring distances to $\xopt$ in the $\norm{\cdot}_{A_k}$ norm,
\alignn{
    \label{eq:amsgrad-prop-start}
    \norm{\xkk - \xopt}_{\Ak}^2 
    &=
    \norm{\xk - \xopt}_{\Ak}^2 - 2 \etak \lin{\nabla \sfkk, \xk - \xopt} + \etak^2 \norm{\nabla \sfkk}^2_{\Ak^{-1}}
}
To bound the RHS, 
we use the assumption on the step-sizes (\cref{eq:amsgrad-assumption-step-size})
and \ref{eq:individual-convexity},
\aligns{
    &- 2 \etak \lin{\nabla \sfkk, \xk - \xopt} + \etak^2 \norm{\nabla \sfkk}^2_{\Ak^{-1}},
    \\
    &\leq 
    - 2 \etak \lin{\nabla \sfkk, \xk - \xopt} 
    + M \etak (\fk(\xk) - \fkopt),
    \tag{Step-size assumption, \cref{eq:amsgrad-assumption-step-size}}
    \\
    &\leq 
    - 2 \etak [\fk(\xk) - \fk(\xopt)]
    + M \etak (\fk(\xk) - \fkopt),
    \tag{\ref{eq:individual-convexity}} 
    \\
    &\leq 
    - 2 \etak [\fk(\xk) - \fk(\xopt)]
    + M \etak (\fk(\xk) - \fk(\xopt) + \fk(\xopt) - \fkopt),
    \tag{$\pm \fk(\xopt)$} 
    \\
    &\leq
    - (2 - M) \etak [\fk(\xk) - \fk(\xopt)] + 
    M \eta_{\max} \paren{\fk(\xopt) - \fkopt}. \tag{$\etak\leq\eta_{\max}$}
}
Plugging the inequality back into \cref{eq:amsgrad-prop-start} and reorganizing the terms yields
\alignn{
    \label{eq:amsgrad-proof-interim-1}
    \begin{aligned}
    (2 - M) \etak [\fk(\xk) - \fk(\xopt)]
    \leq 
    &\,\paren{\norm{\xk - \xopt}_{\Ak}^2 - \norm{\xkk - \xopt}_{\Ak}^2}
    \\
    &+ M\eta_{\max} \paren{\fk(\xopt) - \fkopt}
    \end{aligned}
}
Using \cref{lem:sps}, we have that 
\aligns{
    \begin{aligned}
    (2 - M) \etak [\fk(\xk) - \fk(\xopt)]
    \geq
    &\, (2-M) \eta_{\min} \paren{\fk(\xk) - \fk(\xopt)}
    \\
    &- (2-M) (\eta_{\max} - \eta_{\min}) \paren{\fk(\xopt) - \fkopt}.
    \end{aligned}
}
Using this inequality in \cref{eq:amsgrad-proof-interim-1}, we have that 
\aligns{
    &(2-M) \eta_{\min} \paren{\fk(\xk) - \fk(\xopt)}
    - (2-M) (\eta_{\max} - \eta_{\min}) \paren{\fk(\xopt) - \fkopt}
    \\
    &\quad\quad\quad \leq 
    \,\paren{\norm{\xk - \xopt}_{\Ak}^2 - \norm{\xkk - \xopt}_{\Ak}^2}
    + M \eta_{\max} \paren{\fk(\xopt) - \fkopt},
}
Moving the terms depending on $\fk(\xopt) - \fkopt$ to the RHS, 
\aligns{
    (2-M) \eta_{\min} \paren{\fk(\xk) - \fk(\xopt)}
    \leq 
    &\,\paren{\norm{\xk - \xopt}_{\Ak}^2 - \norm{\xkk - \xopt}_{\Ak}^2}
    \\
    &+ (2\eta_{\max} - (2-M)\eta_{\min}) \paren{\fk(\xopt) - \fkopt}.
}
Taking expectations and summing across iterations yields
\aligns{
    (2 - M) \eta_{\min} \sum_{k=1}^T \Expect{\fk(\xk) - \fk(\xopt)}
    \leq 
    & \Expect{\sum_{k=1}^T \paren{\norm{\xk - \xopt}_{\Ak}^2 - \norm{\xkk - \xopt}_{\Ak}^2}}
    \\
    &+ \paren{2 \eta_{\max} - (2-M) \eta_{\min}} T \sigma^2.
}
Using \cref{lem:telescoping-distance} to telescope the distances and using the \ref{eq:bounded-preconditioner},
\aligns{
    \sum_{k=1}^{T} \norm{\xk - \xopt}_{\Ak}^2 - \norm{\xkk-\xopt}_{\Ak}^2 
    &\leq 
    \sum_{k=1}^T  \norm{\xk - \xopt}_{\Ak - A_{k-1}}^2 \leq D^2 \, \Tr{A_T} \leq D^2 \, d \, a_{\max},
}
which guarantees that
\aligns{
    (2 - M) \eta_{\min} \sum_{k=1}^T \Expect{f(\xk) - f(\xopt)}
    \leq 
    & D^2 d a_{\max} \,+\, \paren{2 \eta_{\max} - (2-M) \eta_{\min}} T \sigma^2.
}
Dividing by $T (2-M) \eta_{\min}$ and using Jensen's inequality finishes the proof, giving the rate for the averaged iterate,
\aligns{
    \Expect{f(\bar{\x}_T) - f(\xopt)}
    \leq 
    \frac{1}{T} \frac{D^2 d a_{\max}}{(2-M) \eta_{\min}} 
    \,+\, \paren{\frac{2}{2-M} \frac{\eta_{\max}}{\eta_{\min}} - 1} \sigma^2. 
    \tag*{\qedhere}
}
\end{proof}

\clearpage

\section{\amsgrad/ with momentum}
\label{app:amsgrad-momentum-proofs}
We first show the relation between the \amsgrad/ momentum and heavy ball momentum and then present the proofs with \amsgrad/ momentum in~\ref{app:amsgrad-mom-proofs}
and heavy ball momentum in~\ref{app:amsgrad-hb-proofs}.  
\subsection{Relation between the \amsgrad/ update and preconditioned SGD with heavy-ball momentum}
\label{app:amsgrad-momentum-equivalence}
Recall that the \amsgrad/ update is given as:
\begin{align*}
\xkk & = \xk - \etak \,  \smallAkInverse \mk  \quad \text{;} \quad \mk = \beta  \mkm + (1 - \beta) \gradk{\xk}
\end{align*}
Simplifying, 
\begin{align*}
\xkk & = \xk - \etak \,  \smallAkInverse (\beta \mkm + (1 -\beta) \gradk{\xk}) \\
\xkk & = \xk - \etak (1 - \beta) \, \smallAkInverse \gradk{\xk} - \etak  \beta \, \smallAkInverse \mkm 
\intertext{From the update at iteration $k-1$,}
\xk & = \xkm - \eta_{k-1} \,  \smallAkmInverse m_{k-1} \\
\implies - m_{k-1} & = \frac{1}{\eta_{k-1}} \smallAkm \left(\xk - \xkm \right)
\intertext{From the above relations,}
\xkk & = \xk - \etak (1 - \beta) \, \smallAkInverse \gradk{\xk} + \beta \, \frac{\etak}{\eta_{k-1}} \, \smallAkInverse \smallAkm \left(\xk - \xkm \right)
\end{align*}
which is of the same form as 
\aligns{
\xkk & = \xk - \etak \,  \smallAkInverse + \gamma (\xk - \xkm),
}
the update with heavy ball momentum. The two updates are equivalent up to constants except for the key difference that for \amsgrad/, the momentum vector $\paren{\xk - \xkm}$ is further preconditioned by $\smallAkInverse \smallAkm$.
\newpage

\subsection{Proofs for \amsgrad/ with momentum}
\label{app:amsgrad-mom-proofs}
We now give the proofs for \amsgrad/ having the update. 
\begin{align*}
\xkk & = \xk - \etak \,  \smallAkInverse \mk  \quad \text{;} \quad \mk = \beta  \mkm + (1 - \beta) \gradk{\xk}
\end{align*}
We analyze it in the smooth setting using a constant step-size (\cref{thm:amsgrad-constant-mom}), conservative Armijo SPS (\cref{thm:amsgrad-sps-mom})
and conservative Armijo SLS (\cref{thm:amsgrad-sls-mom}). 
As before, we abstract the common elements to a general proposition 
and specialize it for each of the theorems.

\begin{thmbox}
\begin{proposition}
\label{prop:amsgrad-mom}
In addition to assumptions of \cref{thm:adagrad-constant}, assume
that (iv) the preconditioners are non-decreasing and have (v) bounded eigenvalues in the $[a_{\min}, a_{\max}]$ range. If the step-sizes are lower-bounded and non-increasing, 
$\eta_{\min} \leq \etak \leq \etakm$ and satisfy
\alignn{
    \etak \norm{\nabla \fk(\xk)}^2_{A_k^{-1}} \leq
    M \paren{\fk(\xk) - \fkopt},
    \quad \text{ for some $M < 2\frac{1-\beta}{1+\beta}$},
    \label{eq:amsgrad-mvgavg-assumption-step-size}
}
using uniform averaging $\bar{\x}_T = \frac{1}{T}\sum_{k=1}^T \xk$ leads to the rate
\aligns{
	\Expect{f(\bar{\x}_T) - f^*}
	\leq
	\frac{1+\beta}{1-\beta}\paren{
		2 - \frac{1+\beta}{1-\beta}M
	}^{-1}
	\brackets{
	\frac{D^2 d a_{\max}}{\eta_{\min} T} 
	+ M \sigma^2
	}.
}
\end{proposition}
\end{thmbox}

We first show how the convergence rate of each step-size method can be derived from
\cref{prop:amsgrad-mom}.
\myquote{\begin{thmbox}
\restateThmConstantAmsgradmom*
\end{thmbox}
}
\begin{proof}[Proof of \cref{thm:amsgrad-constant-mom}]
Using \ref{eq:bounded-preconditioner} and \ref{eq:individual-smoothness}, we have that 
\aligns{
    \eta \norm{\nabla \fk(\xk)}^2_{A_k^{-1}}
    \leq
    \eta \frac{1}{a_{\min}} \norm{\nabla \fk(\xk)}^2 
    \leq 
    \eta \frac{2L_{\max}}{a_{\min}} (\fk(\xk) - \fkopt).
}
Using a constant step-size 
$\eta = \frac{1-\beta}{1+\beta}\frac{a_{\min}}{2L_{\max}}$ 
satisfies the requirement of \cref{prop:amsgrad-mom}
(\cref{eq:amsgrad-mvgavg-assumption-step-size})
with constant $M = \frac{1-\beta}{1+\beta}$.
The convergence is then,
\aligns{
	\Expect{f(\bar{\x}_T) - f(\xopt)}
	&\leq
	\frac{1+\beta}{1-\beta}\paren{
		2 - \frac{1+\beta}{1-\beta}M
	}^{-1}
	\brackets{
	\frac{D^2 d a_{\max}}{\eta_{\min} T} 
	+ M \sigma^2,
	}
	\\
	&=	
	\frac{1+\beta}{1-\beta}
	\brackets{
	\frac{D^2 d a_{\max}}{\frac{1-\beta}{1+\beta}\frac{a_{\min}}{2L_{\max}}  T} 
	+ \frac{1-\beta}{1+\beta} \sigma^2,
	}
	\\
	&=
	\paren{\frac{1+\beta}{1-\beta}}^{\!\!2}
	\frac{2 L_{\max} D^2 d \kappa}{  T} 
	+ \sigma^2,
}
with $\kappa = a_{\max}/a_{\min}$.


\end{proof}
\myquote{\begin{thmbox}\restateThmSPSAmsgradmom*\end{thmbox}}
\begin{proof}[Proof of \cref{thm:amsgrad-sps-mom}]
For Armijo SPS, \cref{lem:sps-bounds} guarantees that
\aligns{
    \etak \norm{\nabla \fk(\xk)}^2_{\Ak^{-1}} \leq \frac{1}{c}(\fk(\xk) - \fk^*),
    &&\text{and}&&
	\frac{a_{\min}}{2 c \, L_{\max}}
    \leq
    \etak.
}
Setting $c = \frac{1 + \beta}{1 - \beta}$ ensures that $M = 1/c$ satisfies the requirement of~\cref{prop:amsgrad-mom} and $\eta_{\min} \geq \frac{1-\beta}{1+\beta}\frac{a_{\min}}{2 L_{\max}}$. Plugging in these values into~\cref{prop:amsgrad-mom} completes the proof.  
\end{proof}

\begin{thmbox}
\begin{restatable}{theorem}{restateThmSLSAmsgradmom}
\label{thm:amsgrad-sls-mom}
Under the assumptions of~\cref{thm:adagrad-constant} and assuming (iv) non-decreasing preconditioners (v) bounded eigenvalues in the $[a_{\min}, a_{\max}]$ interval, \amsgrad/ with momentum with parameter $\beta \in [0,\nicefrac{1}{5}$), conservative Armijo SLS with $c = \frac{2}{3} \frac{1+\beta}{1-\beta}$ and uniform averaging converges at a rate,
\aligns{
	\Expect{f(\bar{\x}_T) - f^*}
	\leq
	3 \frac{1+\beta}{1-5\beta} \frac{L_{\max}D^2 d \kappa}{T} 
	+ 3 \sigma^2
}
\end{restatable}
\end{thmbox}
\begin{proof}[Proof of \cref{thm:amsgrad-sls-mom}]
For Armijo SLS, \cref{lem:armijo-line-search} guarantees that
\aligns{
    \etak \norm{\nabla \fk(\xk)}^2_{\Ak^{-1}} \leq \frac{1}{c}(\fk(\xk) - \fk^*),
    &&\text{and}&&
    \frac{2 (1-c) \, a_{\min}}{ L_{\max} }
    \leq
    \etak.
}
The line-search parameter $c$ is restricted to $[0,1]$ and 
relates to the the requirement parameter $M$ of \cref{prop:amsgrad-mom}
(\cref{eq:amsgrad-mvgavg-assumption-step-size})
through $M = 1/c$. The combined requirements on $M$
are then that 
$1 < M < 2 \frac{1-\beta}{1+\beta}$, 
which is only feasible if $\beta < \frac{1}{3}$.
To leave room to satisfy the constraints, let 
$\beta < \frac{1}{5}$.

Setting $\frac{1}{c} = M = \frac{3}{2} \frac{1-\beta}{1+\beta}$
satisfies the constraints and requirement for \cref{prop:amsgrad-mom},
and
\aligns{
	\Expect{f(\bar{\x}_T) - f(\xopt)}
	&\leq
	\frac{1+\beta}{1-\beta}\paren{
		2 - \frac{1+\beta}{1-\beta}M
	}^{-1}
	\brackets{
	\frac{D^2 d a_{\max}}{\eta_{\min} T} 
	+ M \sigma^2
	},
	\\
	&=
	\frac{1+\beta}{1-\beta}\paren{
		2 - \frac{3}{2}
	}^{-1}
	\brackets{
	\frac{ L_{\max} }{ 2 (1-c) \, a_{\min}}\frac{D^2 d a_{\max}}{T} 
	+ \frac{3}{2} \frac{1-\beta}{1+\beta} \sigma^2
	},
	\\
	&=
	\frac{1+\beta}{1-\beta} \frac{ L_{\max} }{(1-c) }\frac{D^2 d \kappa}{T} 
	+ 3 \sigma^2
	=
	3 \frac{1+\beta}{1-5\beta} \frac{L_{\max}D^2 d \kappa}{T} 
	+ 3 \sigma^2
	.
}
where the last step substituted $1/(1-c)$, 
\aligns{
	1-c = 1 - \frac{2}{3}\frac{1+\beta}{1-\beta} 
	= \frac{3(1-\beta) - 2(1+\beta)}{3(1-\beta)} 
	= \frac{1}{3}\frac{1-5\beta}{1-\beta}.
	\tag*{\qedhere}
}



\end{proof}

Before diving into the proof of \cref{prop:amsgrad-mom},
we prove the following lemma,
\begin{thmbox}
\begin{lemma}
\label{lemma:four-point}
For any set of vectors $a,b,c,d$, if $a = b + c$, then,
\aligns{
\normsq{a - d} & = \normsq{b - d} - \normsq{a - b} + 2 \langle c, a - d \rangle
}
\end{lemma}
\end{thmbox}
\begin{proof}
\aligns{
\normsq{a - d} &= \normsq{b + c - d} = \normsq{b - d} + 2 \langle c, b - d \rangle + \normsq{c} \\
\intertext{Since $c = a - b$,}
& = \normsq{b - d} + 2 \langle a - b, b - d \rangle + \normsq{a - b} \\
& = \normsq{b - d} + 2 \langle a - b, b - a + a - d \rangle + \normsq{a - b} \\
& = \normsq{b - d} + 2 \langle a - b, b - a \rangle + 2 \langle a - b, a - d \rangle + \normsq{a - b} \\
& = \normsq{b - d} - 2 \normsq{a - b} + 2 \langle a - b, a - d \rangle + \normsq{a - b} \\
& = \normsq{b - d} - \normsq{a - b} + 2 \langle c, a - d \rangle 
}
\end{proof}

We now move to the proof of the main proposition. Our proof follows the structure of~\citet{reddi2019convergence, alacaoglu2020new}.  
\begin{proof}[Proof of \cref{prop:amsgrad-mom}]
To reduce clutter, let $\Bk = \Ak/\etak$.
Using the update, we have the expansion
\aligns{
	\xkk - \xopt 
	&= \, \paren{\xk - \Bkinv \mk} - \xopt,
	\\
	&= \, \paren{\xk - (1 - \beta) \Bkinv \gradk{\xk} - \beta \Bkinv \mkm} 
	- \xopt,
}
Measuring distances in the $\norm{\cdot}_{\Bk}\!$-norm,
such that $\norm{x}_{\Bk}^2 = \lin{x, \Bk x}$,
\aligns{
	\indnormsq{\xkk - \xopt}{\Bk} 
	= \indnormsq{\xk - \xopt}{\Bk} 
	&- 2 (1 - \beta) \, \langle \xk - \xopt, \gradk{\xk} \rangle,
	\\
	& - 2 \beta \, \langle \xk - \xopt, \mkm \rangle + \indnormsq{\mk}{\Bkinv}.
}
We separate the distance to $\xopt$ from the momentum
in the second inner product
using the update
and \cref{lemma:four-point} with 
$a = c = \Bkm^{\scriptscriptstyle 1/2} (\xk - \xopt)$, $b = \textbf{0}$, 
$d = \Bkm^{\scriptscriptstyle 1/2} (\xkm - \xopt)$.
\aligns{
	- 2 \langle \mkm, \xk - \xopt \rangle 
	&= - 2 \, \langle \Bkm (\xkm - \xk), \xk - \xopt \rangle,
	\\
	&= 
	\brackets{
		\indnormsq{\xk - \xkm}{\Bkm} 
		+ \indnormsq{\xk - \xopt}{\Bkm} 
		- \indnormsq{\xkm - \xopt}{\Bkm}
	},
	\\
	&= 
		\indnormsq{\mkm}{\Bkminv} 
		+ \indnormsq{\xk - \xopt}{\Bkm} 
		- \indnormsq{\xkm - \xopt}{\Bkm}
	,
	\\
	&\leq 
	\indnormsq{\mkm}{\Bkminv} 
	+ \indnormsq{\xk - \xopt}{\Bk} 
	- \indnormsq{\xkm - \xopt}{\Bkm},
}
where the last inequality uses 
the fact that $\etak \leq \etakm$ and $\Ak \succeq \Akm$, 
which implies $\Bk \succeq \Bkm$,
and $\indnormsq{\xk - \xopt}{\Bkm} \leq \indnormsq{\xk - \xopt}{\Bk}$.
Plugging this inequality in and grouping terms yields
\aligns{
	2 (1 - \beta) \, \langle \xk - \xopt, \gradk{\xk} \rangle 
	\leq
	&\,
	\brackets{\indnormsq{\xk - \xopt}{\Bk} - \indnormsq{\xkk - \xopt}{\Bk}}
	\\
	&+ \beta \brackets{ \indnormsq{\xk - \xopt}{\Bk} - \indnormsq{\xkm - \xopt}{\Bkm}}
	\\ 
	&+ \brackets{\beta \, \indnormsq{\mkm}{\Bkminv} + \indnormsq{\mk}{\Bkinv}}
}
By convexity, 
the inner product on the left-hand-side 
is bounded by 
$\langle \xk - \xopt, \gradk{\xk} \rangle \geq \fk(\xk) - \fk(\xopt)$.
The first two lines of the right-hand-side will telescope
if we sum all iterations, so we only need to treat the norms of the momentum terms.
We introduce a free parameter $\delta \geq 0$, 
that is only used for the analysis,
and expand
\aligns{
	\beta \, \indnormsq{\mkm}{\Bkminv} + \indnormsq{\mk}{\Bkinv}
	= 
	\beta \, \indnormsq{\mkm}{\Bkminv} + (1+\delta)\indnormsq{\mk}{\Bkinv} 
	- \delta\indnormsq{\mk}{\Bkinv}.
}
To bound $\indnormsq{\mk}{\Bkinv}$, we expand it by its update 
and use Young's inequality to get
\aligns{
	\indnormsq{\mk}{\Bkinv} 
	&=
	\indnormsq{\beta \mkm + (1-\beta) \gradk{\xk}}{\Bkinv} 
	\\
	&\leq
	(1+\epsilon)\beta^2 \indnormsq{\mkm}{\Bkinv} 
	+ (1+\nicefrac{1}{\epsilon})(1-\beta)^2\indnormsq{\gradk{\xk}}{\Bkinv},
}
where $\epsilon > 0$ is also a free parameter, 
introduced to control the tradeoff of the bound.
Plugging this bound in the momentum terms, we get
\aligns{
	\beta \, \indnormsq{\mkm}{\Bkminv} + \indnormsq{\mk}{\Bkinv}
	\leq& \,
	\beta \, \indnormsq{\mkm}{\Bkminv} 
	+ (1+\epsilon) (1+\delta) \beta^2 \indnormsq{\mkm}{\Bkinv} 
	- \delta \indnormsq{\mk}{\Bkinv},
	\\
	&+ (1+\nicefrac{1}{\epsilon}) (1+\delta) (1-\beta)^2 \indnormsq{\gradk{\xk}}{\Bkinv}.
\intertext{As $\Bkinv \preceq \Bkminv$, we have that 
$\indnormsq{\mkm}{\Bkinv} \leq \indnormsq{\mkm}{\Bkminv}$
which implies}
	\leq&\,
	\paren{\beta + (1+\epsilon) (1+\delta)\beta^2 }\indnormsq{\mkm}{\Bkminv} - \delta \indnormsq{\mk}{\Bkinv} 
	\\
	&+ (1+\nicefrac{1}{\epsilon}) (1+\delta) (1-\beta)^2 \indnormsq{\gradk{\xk}}{\Bkinv}.
}
To get a telescoping sum, we set $\delta$ to be equal to $\beta + (1+ \epsilon)(1+\delta)\beta^2$, 
which is satisfied if $\delta = \frac{\beta + (1+ \epsilon)\beta^2}{1-(1+ \epsilon)\beta^2}$,
and $\delta > 0$ is satisfied if $\beta < \nicefrac{1}{\m!\sqrt{1+\epsilon}}$.
We now plug back the inequality
\aligns{
	\beta \, \indnormsq{\mkm}{\Bkminv} + \indnormsq{\mk}{\Bkinv}
	\leq&\,
	\delta \brackets{\indnormsq{\mkm}{\Bkminv} - \indnormsq{\mk}{\Bkinv} }
	\\
	&+ (1+\nicefrac{1}{\epsilon}) (1+\delta) (1-\beta)^2 \indnormsq{\gradk{\xk}}{\Bkinv},
}
in the previous expression to get
\aligns{
	2 (1 - \beta) \, \paren{\fk(\xk) - \fk(\xopt)}
	\leq
	&\,
	\indnormsq{\xk - \xopt}{\Bk} 
	- \indnormsq{\xkk - \xopt}{\Bk} 
	\\
	&+ \beta \left[\indnormsq{\xk - \xopt}{\Bk} - \indnormsq{\xkm - \xopt}{\Bkm} \right] 
	\\ 
	&+ \delta \brackets{\indnormsq{\mkm}{\Bkminv} - \indnormsq{\mk}{\Bkinv}}
	\\
	&+ (1+ \nicefrac{1}{\epsilon}) (1+\delta) (1-\beta)^2 \indnormsq{\gradk{\xk}}{\Bkinv}.
}
All terms now telescope, except the gradient norm which we bound using the step size assumption, 
\aligns{
	\indnormsq{\gradk{\xk}}{\Bkinv} 
	& = \etak \indnormsq{\gradk{\xk}}{\Akinv} \leq M \paren{\fk(\xk) - \fkopt}, \\
	& = M \paren{\fk(\xk) - \fk(\xopt)} + M \paren{\fk(\xopt) - \fkopt}.
}
This gives the expression
\aligns{
	\alpha \, \paren{\fk(\xk) - \fk(\xopt)}
	\leq
	&\,
	\indnormsq{\xk - \xopt}{\Bk} 
	- \indnormsq{\xkk - \xopt}{\Bk} 
	\\
	&+ \beta \left[\indnormsq{\xk - \xopt}{\Bk} - \indnormsq{\xkm - \xopt}{\Bkm} \right] 
	\\ 
	&+ \delta \brackets{\indnormsq{\mkm}{\Bkminv} - \indnormsq{\mk}{\Bkinv}} \\
	&+ (1+ \nicefrac{1}{\epsilon}) (1+\delta) (1-\beta)^2 M \paren{\fk(\xopt) - \fkopt},
}
with $\alpha = 2 (1 - \beta) - (1+ \nicefrac{1}{\epsilon}) (1+\delta) (1-\beta)^2 M$.
Summing all iterations, the individual terms are bounded by 
the \ref{eq:bounded-iterates}
and \cref{lem:telescoping-distance};
\aligns{
	&\fullsum
	\indnormsq{\xk - \xopt}{\Bk} 
	- \indnormsq{\xkk - \xopt}{\Bk} 
	&&\leq D^2 \Tr{\BT} &&\leq \frac{D^2}{\eta_{\min}} \Tr{\AT}
	\\
	\beta &\fullsum
	\indnormsq{\xk - \xopt}{\Bk} - \indnormsq{\xkm - \xopt}{\Bkm}
	&& \leq  
	\beta \indnormsq{\x_T - \xopt}{\BT} &&\leq \beta \frac{D^2}{\eta_{\min}} \Tr{\AT}
	\\
	\delta &\fullsum 
	\indnormsq{\mkm}{\Bkminv} - \indnormsq{\mk}{\Bkinv}
	&&\leq
	\delta \indnormsq{m_0}{\Bstart} &&= 0.
}
Using the boundedness of the preconditioners gives $\Tr{\AT} \leq d a_{\max}$
and the total bound
\aligns{
	\alpha \fullsum \paren{\fk(\xk) - \fk(\xopt)}
	\leq
	\frac{(1+\beta)D^2 d a_{\max}}{\eta_{\min}} + (1+ \nicefrac{1}{\epsilon}) (1+\delta) (1-\beta)^2 M \fullsum \paren{\fk(\xopt) - \fkopt}.
}
Taking expectations,
\aligns{
	\alpha \fullsum \Expect{f(\xk) - f(\xopt)}
	\leq
	\frac{(1+\beta)D^2 d a_{\max}}{\eta_{\min}} + (1+ \nicefrac{1}{\epsilon}) (1+\delta) (1-\beta)^2 M \sigma^2 T.
}
It remains to expand $\alpha$ and simplify the constants. 
We had defined
\aligns{
	\alpha = 2 (1 - \beta) - (1+ \nicefrac{1}{\epsilon}) (1+\delta) (1-\beta)^2 M > 0,
	&&\text{ and }&&
	\delta = \frac{\beta+(1+\epsilon)\beta^2}{1-(1+\epsilon)\beta^2} > 0,
}
where $\epsilon > 0$ is a free parameter. This puts the requirement on $\beta$
that $\beta < {1}/{\sqrt{1+\epsilon}}$.
To simplify the bounds, we set $\beta = {1}/(1+\epsilon)$, $\epsilon = {1}/{\beta}-1$, 
which gives the substitutions
\aligns{
	1 + \epsilon = \frac{1}{\beta} 
	&& 
	1 + \frac{1}{\epsilon} 
	= \frac{1}{1-\beta}
	&&
	\delta = 2 \frac{\beta}{1-\beta}
	&&
	1+\delta = \frac{1+\beta}{1-\beta}.
}
Plugging those into the rate gives
\aligns{
	\alpha \fullsum \Expect{f(\xk) - f(\xopt)}
	\leq
	\frac{(1+\beta)D^2 d a_{\max}}{\eta_{\min}} 
	+ (1+\beta) M \sigma^2 T,
}
while plugging them into $\alpha$ gives
\aligns{
	\alpha 
	&= 2 (1 - \beta) - (1+ \nicefrac{1}{\epsilon}) (1+\delta) (1-\beta)^2 M,
	\\
	&= (1-\beta)\brackets{
		2 - \frac{1+\beta}{1-\beta}M
	},
	\text{ which is positive if } M < 2\frac{1-\beta}{1+\beta}.
}
Dividing by $\alpha T$, using Jensen's inequality and averaging finishes the proof,
with the rate
\aligns{
	\fullsum \Expect{f(\xk) - f(\xopt)}
	\leq
	\frac{1+\beta}{1-\beta}\paren{
		2 - \frac{1+\beta}{1-\beta}M
	}^{-1}
	\brackets{
	\frac{D^2 d a_{\max}}{\eta_{\min} T} 
	+ M \sigma^2
	}
	.
	\tag*{\qedhere}
}
\end{proof}

\newpage

\subsection{Proofs for \amsgrad/ with heavy ball momentum}
\label{app:amsgrad-hb-proofs}
We now give the proofs for \amsgrad/ with heavy ball momentum with the update. 
\begin{align*}
\xkk = \xk - \etak \, \smallAkInverse \gradk{\xk} + \gamma \left(\xk - \xkm \right)    
\end{align*}
We analyze it in the smooth setting using a constant step-size (\cref{thm:amsgrad-constant-HB}), 
a conservative Armijo SPS (\cref{thm:amsgrad-sps-HB})
and conservative Armijo SLS (\cref{thm:amsgrad-sls-HB}). 
As before, we abstract the common elements to a general proposition 
and specialize it for each of the theorems.

\begin{thmbox}
\begin{proposition}
\label{prop:amsgrad-hb}
In addition to assumptions of \cref{thm:adagrad-constant}, assume
that (iv) the preconditioners are non-decreasing and have (v) bounded eigenvalues in the $[a_{\min}, a_{\max}]$ range. If the step-sizes are lower-bounded and non-increasing, $\eta_{\min} \leq \etak \leq \eta_{k-1}$ and satisfy
\alignn{
    \etak \norm{\nabla \fk(\xk)}^2_{A_k^{-1}} \leq
    M \paren{\fk(\xk) - \fkopt},
    \quad \text{ for some $M < 2 - 2 \gamma$},
    \label{eq:amsgrad-hb-assumption-step-size}
}
\amsgrad/ with heavy ball momentum with parameter $\gamma < 1$
and uniform averaging $\bar{\x}_T = \frac{1}{T}\sum_{k=1}^T \xk$ leads to the rate
\aligns{
    \Expect{f(\bar{\x}_T) - f^*}
    \leq 
	\frac{1}{2-2\gamma - M}\brackets{\frac{1}{T}\paren{
		\frac{2 (1+\gamma^2) D^2 a_{\max} d}{\eta_{\min}}
		+ 2\gamma [f(\x_0) - f(\xopt)]
	} + M \sigma^2}.
} 
\end{proposition}
\end{thmbox}

We first show how the convergence rate of each step-size method can be derived from
\cref{prop:amsgrad-hb}.

\begin{thmbox}
\begin{restatable}{theorem}{restateThmConstantAmsgradHB}
\label{thm:amsgrad-constant-HB}
Under the assumptions of~\cref{thm:adagrad-constant} and assuming (iv) non-decreasing preconditioners (v) bounded eigenvalues in the $[a_{\min}, a_{\max}]$ range, \amsgrad/ with heavy ball momentum with parameter $\gamma \in [0,1)$, constant step-size $\eta = \frac{2 a_{\min} \, (1 - \gamma)}{3 L_{\max}}$ and uniform averaging converges at a rate
\aligns{
    \Expect{f(\bar{\x}_T) - f^*}
	&\leq 
	\frac{1}{T}\paren{
		\frac{9}{2}
		\frac{1+\gamma^2}{(1-\gamma)^{2}}
		L_{\max} \, D^2 \kappa d
		+ \frac{3 \gamma}{(1-\gamma)} [f(\x_0) - f(\xopt)]
	} + 2 \sigma^2.
}
\end{restatable}
\end{thmbox}
\begin{proof}[Proof of \cref{thm:amsgrad-constant-HB}]
Using \ref{eq:bounded-preconditioner} and \ref{eq:individual-smoothness}, we have that 
\aligns{
    \eta \norm{\nabla \fk(\xk)}^2_{A_k^{-1}}
    \leq
    \eta \frac{1}{a_{\min}} \norm{\nabla \fk(\xk)}^2 
    \leq 
    \eta \frac{2L_{\max}}{a_{\min}} (\fk(\xk) - \fkopt).
}
A constant step-size $\eta = \nicefrac{2 a_{\min} \, (1 - \gamma)}{3 L_{\max}}$ 
means the requirement for \cref{prop:amsgrad-hb} is satisfied with
$M = \frac{4}{3}(1 - \gamma)$.  
Plugging 
$\paren{2 - 2\gamma - M} = \frac{2}{3}(1-\gamma)$ in \cref{prop:amsgrad-hb} finishes the proof.  
\end{proof}

\begin{thmbox}
\begin{restatable}{theorem}{restateThmSPSAmsgradHB}
\label{thm:amsgrad-sps-HB}
Under the assumptions of~\cref{thm:adagrad-constant} and assuming (iv) non-decreasing preconditioners (v) bounded eigenvalues in the $[a_{\min}, a_{\max}]$ interval, \amsgrad/ with heavy ball momentum with parameter $\gamma \in [0,1)$, conservative Armijo SPS with $c = \nicefrac{3}{4(1-\gamma)}$ and uniform averaging converges at a rate,
\aligns{
	\Expect{f(\bar{\x}_T) - f^*}
	&\leq 
	\frac{1}{T}\paren{
		\frac{9}{2}\frac{1+\gamma^2}{(1-\gamma)^2} L_{\max} D^2 \kappa d
		+ \frac{3\gamma }{(1-\gamma)} [f(\x_0) - f(\xopt)]
	} + 2 \sigma^2.
}
\end{restatable}
\end{thmbox}
\begin{proof}[Proof of \cref{thm:amsgrad-sps-HB}]
For Armijo SPS, \cref{lem:sps-bounds} guarantees that
\aligns{
    \etak \norm{\nabla \fk(\xk)}^2_{\Ak^{-1}} \leq \frac{1}{c}(\fk(\xk) - \fk^*),
    &&\text{and}&&
    \frac{a_{\min}}{2 c \, L_{\max}}
    \leq
    \etak.
}
Selecting $c = \nicefrac{3}{4(1-\gamma)}$ 
gives $M = \nicefrac{4}{3}(1 - \gamma) \leq 2 (1 - \gamma)$
and the requirement of \cref{prop:amsgrad-hb} are satisfied. 
The minimum step-size is then
$\eta_{\min} = \frac{a_{\min}}{2c L_{\max}} = \frac{2 a_{\min} \, (1-\gamma)}{3 L_{\max}}$,
so $\eta_{\min}$ and $M$ are the same as in the constant step-size case 
(\cref{thm:amsgrad-constant-HB}) and the same rate applies.
\end{proof}

\begin{thmbox}
\begin{restatable}{theorem}{restateThmSLSAmsgradHB}
\label{thm:amsgrad-sls-HB}
Under the assumptions of~\cref{thm:adagrad-constant} and assuming (iv) non-decreasing preconditioners (v) bounded eigenvalues in the $[a_{\min}, a_{\max}]$ interval, \amsgrad/ with heavy ball momentum with parameter $\gamma \in [0,\nicefrac{1}{4})$, conservative Armijo SLS with $c = \nicefrac{3}{4 (1 - \gamma)}$ and uniform averaging converges at a rate,
\aligns{
	\Expect{f(\bar{\x}_T) - f^*}
	&\leq 
	\frac{1}{T}\paren{
		6
		\frac{1+\gamma^2}{1-4\gamma}
		L_{\max} D^2 \kappa d
		+ \frac{3 \gamma}{(1-\gamma)} [f(\x_0) - f(\xopt)]
	} + 2 \sigma^2.
}
\end{restatable}
\end{thmbox}
\begin{proof}[Proof of \cref{thm:amsgrad-sls-HB}]
Selecting $c = \nicefrac{3}{4 (1-\gamma)}$ is feasible if
$\gamma < 1/4$ as $c < 1$. The Armijo SLS (\cref{lem:armijo-line-search}) then guarantees that
\aligns{
    \etak \norm{\nabla \fk(\xk)}^2_{\Ak^{-1}} \leq \frac{1}{c}(\fk(\xk) - \fk^*),
    &&\text{and}&&
    \frac{2 (1-c) \, a_{\min}}{ L_{\max}}
    \leq
    \eta,
}
which satisfies the requirements of \cref{prop:amsgrad-hb}
with $M = \frac{4}{3}(1 - \gamma)$.
Plugging $M$ in the rate yields
\aligns{
    \Expect{f(\bar{\x}_T) - f(\xopt)}
	&\leq 
	\frac{1}{T}\paren{
		6 \frac{1+\gamma^2}{1-\gamma}\frac{D^2 a_{\max} d}{\eta_{\min}}
		+ \frac{3 \gamma}{(1-\gamma)} [f(\x_0) - f(\xopt)]
	} + 2 \sigma^2,
}
With $c = \frac{\nicefrac{3}{4}}{1-\gamma}$, 
$\eta_{\min} \geq \frac{2(1-c)a_{\min}}{L_{\max}} 
= \frac{2 a_{\min}}{L_{\max}}\frac{1-4\gamma}{4(1-\gamma)}$. 
Plugging it into the above bound yields
\aligns{
	\Expect{f(\bar{\x}_T) - f(\xopt)}
	&\leq 
	\frac{1}{T}\paren{
		6
		\frac{1+\gamma^2}{1-4\gamma}
		L_{\max} D^2 \kappa d
		+ \frac{3 \gamma}{(1-\gamma)} [f(\x_0) - f(\xopt)]
	} + 2 \sigma^2.
	\tag*{\qedhere}
}
\end{proof}

We now move to the proof of the main proposition. Our proof follows the structure of~\citet{ghadimi2015global, sebbouh2020convergence}. 
\begin{proof}[Proof of \cref{prop:amsgrad-hb}]
\begin{tsmath}
Recall the update for \amsgrad/ with heavy-ball momentum,
\alignn{\label{eq:mom-def}
	\xkk = \xk - \etak \Ak^{-1} \nabla \fk(\xk) + \gamma(\xk - \xkm).
}
The proof idea is to analyze the distance from $\xopt$ to $\xk$ \emph{and a momentum term},
\alignn{\label{eq:mom-lyapunov}
	\norm{\delta_k}^2 = \norm{\xk + \mk - \xopt}_{\Ak}^2,
	&& \text{where } \mk = \frac{\gamma}{1-\gamma}(\xk - \xkm),
}
by considering the momentum update (Eq. \ref{eq:mom-def}) 
as a preconditioned step on the joint iterates $(\xk + \mk)$,
\alignn{\label{eq:mom-evolution}
	\xkk + \mkk
	= \xk + \mk - \frac{\etak}{1-\gamma} \Ak^{-1} \nabla \fk(\xk).
}
Let us verify \cref{eq:mom-evolution}.
First, expressing $\xkk + \mkk$ as a weighted difference of $\xkk$ and $\xk$,
\aligns{
	\xkk + \mkk
	&= \xkk + \frac{\gamma}{1-\gamma}(\xkk - \xk)
	= \frac{1}{1-\gamma} \xkk - \frac{\gamma}{1-\gamma}\xk.
\intertext{Expanding $\xkk$ in terms of the update rule then gives}
	&= \frac{1}{1-\gamma} (\xk - \etak \Ak^{-1} \nabla \fk(\xk) + \gamma(\xk - \xkm)) - \frac{\gamma}{1-\gamma}\xk,
	\\
	&= \frac{1}{1-\gamma} (\xk - \etak \Ak^{-1} \nabla \fk(\xk) - \gamma \xkm),
	\\
	&= \frac{1}{1-\gamma} \xk - \frac{\gamma}{1-\gamma}\xkm - \frac{\etak}{1-\gamma} \Ak^{-1}\nabla \fk(\xk),
}
which can then be re-written as $\xk + \mk - \frac{\etak}{1-\gamma} \Ak^{-1} \nabla \fk(\xk)$.
\end{tsmath}
The analysis of the method then follows similar steps 
as the analysis without momentum. 
Using \cref{eq:mom-evolution}, we have the recurrence
\alignn{\begin{aligned}\label{eq:mom-sgd-recursion}
	\norm{\delta_{k+1}}_{\Ak}^2
	&=
	\norm{\xkk + \mkk - \xopt}^2_{\Ak}
	=\textstyle
	\norm{\xk + \mk - \frac{\etak}{1-\gamma} \Ak^{-1}\nabla\fk(\xk) - \xopt}^2_{\Ak},
	\\
	&= \norm{\delta_k}^2_{\Ak} - \frac{2\etak}{1-\gamma} \lin{\nabla \fk(\xk), \xk + \mk - \xopt} 
	+ \frac{\etak^2}{(1-\gamma)^2}\norm{\nabla \fk(\xk)}^2_{\Ak^{-1}}.
\end{aligned}}
To bound the inner-product, we use \ref{eq:individual-convexity} to relate it to the optimality gap,
\aligns{
	\lin{\nabla \fk(\xk), \xk + \mk - \xopt}
	&=
	\lin{\nabla \fk(\xk), \xk - \xopt} + \frac{\gamma}{1-\gamma} \lin{\nabla \fk(\xk), \xk - \xkm},
	\\
	&\geq 
	\fk(\xk) - \fk(\xopt) + \frac{\gamma}{1-\gamma}[\fk(\xk) - \fk(\xkm)],
	\\
	&=
	\frac{1}{1-\gamma}[\fk(\xk) - \fk(\xopt)] - \frac{\gamma}{1-\gamma}[\fk(\xkm) - \fk(\xopt)].
}
To bound the gradient norm, we use the step-size assumption
that 
\aligns{
	\etak \norm{\nabla \fk(\xk)}^2_{\Ak^{-1}} \leq M[\fk(\xk) - \fk^*] = M [\fk(\xk) - \fk(\xopt)] + M [\fk(\xopt) - \fk^*].
}
For simplicity of notation, let us define the shortcuts 
\aligns{
	\hk(\x) = \fk(\x) - \fk(\xopt),
	&&
	\sigma_k^2 = \fk(\xopt) - \fk^*.
}
Plugging those two inequalities in the recursion of \cref{eq:mom-sgd-recursion}
gives
\aligns{
	\norm{\delta_{k+1}}^2_{\Ak}
	& \leq
	\norm{\delta_{k}}^2_{\Ak}
	- \frac{\etak}{(1-\gamma)^2}\paren{
		2 - M}
	\hk(\xk)
	+ \frac{2\etak \gamma}{(1-\gamma)^2}\hk(\xkm)
	+ \frac{M \etak}{(1 - \gamma)^2}\sigma_k^2.
}
We can now divide by $\nicefrac{\etak}{(1-\gamma)^2}$ and reorganize the inequality as 
\aligns{
	\paren{2 - M} \hk(\xk)
	- 2\gamma \hk(\xkm)
	\leq \frac{(1-\gamma)^2}{\etak} \paren{
	\norm{\delta_{k}}^2_{\Ak}
	-\norm{\delta_{k+1}}^2_{\Ak} 
	} + M \sigma_k^2.
}
Taking the average over all iterations, 
the inequality yields
\aligns{
	\frac{1}{T} \sum_{k=1}^T
	\paren{2 - M} \hk(\xk)
	- 2\gamma \hk(\xkm)
	& \leq 
	\frac{1}{T} \sum_{k=1}^T
	\frac{(1-\gamma)^2}{\etak} \paren{
	\norm{\delta_{k}}^2_{\Ak}
	-\norm{\delta_{k+1}}^2_{\Ak}
	} + M \sigma_k^2.
}
To bound the right-hand side, 
under the assumption that the iterates are bounded by
$\norm{\xk - \xopt} \leq D$, 
we use Young's inequality to get a bound on $\norm{\delta_k}^2$;
\aligns{
	\norm{\delta_k}_{2}^2
	&= \norm{\xk + \mk - \xopt}_{2}^2
	\textstyle 
	= \norm{\frac{1}{1-\gamma}(\xk - \xopt) - \frac{\gamma}{1-\gamma}(\xkm - \xopt)}_{2}^2
	\\
	&\leq \frac{2}{(1-\gamma)^2} \paren{
		\norm{\xk - \xopt}_{2}^2 + \gamma^2 \norm{\xkm - \xopt}_{2}^2
	}
	\leq
	\frac{2(1+\gamma^2)}{(1-\gamma)^2}D^2 = \Delta^2.
}
Given the upper bound $\norm{\delta_k}_{2} \leq \Delta$,
a reorganization of the sum 
lets us apply \cref{lem:telescoping-distance} to get
\begin{tsmath}\aligns{
	\sum_{k=1}^T
	\frac{1}{\etak} \paren{
	\norm{\delta_{k}}^2_{\Ak}
	-\norm{\delta_{k+1}}^2_{\Ak}
	}
	&=
	\sum_{k=1}^T
	\norm{\delta_{k}}^2_{\frac{1}{\etak}\Ak}
	-\sum_{k=1}^T \norm{\delta_{k+1}}^2_{\frac{1}{\etak}\Ak}
	\\
	&= 
	\fullsum
	\norm{\delta_{k}}^2_{\frac{1}{\etak}\Ak}
	-
	\sum_{k=2}^{T+1} \norm{\delta_{k}}^2_{\frac{1}{\etakm}\Akm}
	\\
	&\leq
	\fullsum
	\norm{\delta_{k}}^2_{\frac{1}{\etak}\Ak}
	-
	\sum_{k=1}^{T} \norm{\delta_{k}}^2_{\frac{1}{\etakm}\Akm}
	+ \norm{\delta_1}^2_{\frac{1}{\eta_0}A_0}
	\\
	&=
	\fullsum
	\norm{\delta_{k}}^2_{\frac{1}{\etak}\Ak - \frac{1}{\etakm}\Akm}
	\leq
	\frac{\Delta^2 a_{\max} d}{\eta_{\min}},
}\end{tsmath}
where the last step uses the convention $A_0 = 0$
and \cref{lem:telescoping-distance}
on $\delta_k$ instead of $\xk-\xopt$.
Plugging this inequality in, we get the simpler bound on the right-hand-side
\aligns{
	\frac{1}{T} \sum_{k=1}^T
	\paren{2 - M} \hk(\xk)
	- 2\gamma \hk(\xkm)
	\leq 
	\frac{2 (1+\gamma^2) D^2 a_{\max} d}{T \eta_{\min}} + M \sigma_k^2.
}
Now that the step-size is bounded deterministically, we can take the expectation on both sides to get
\aligns{
	\frac{1}{T} \Expect{
	\sum_{k=1}^T
	\paren{2 - M} h(\xk)
	- 2\gamma h(\xkm)
	}
	\leq 
	\frac{2 (1+\gamma^2) D^2 a_{\max} d}{T \eta_{\min}} + M \sigma^2,
}
where $h(\x) = f(\x) - f^*$
and $\sigma^2 = \Expect{\fk(\xopt) - \fk^*}$.
To simplify the left-hand-side, 
we change the weights on the optimality gaps to get a telescoping sum,
\begin{tsmath}
\aligns{
	\sum_{k=1}^T
	\paren{2 - M} h(\xk)
	- 2\gamma h(\xkm)
	&=
	\sum_{k=1}^T
	\paren{2 - 2\gamma - M} h(\xk)
	+ 2 \gamma h(\xk)
	- 2\gamma h(\xkm)
	,
	\\
	&= \,
	\paren{2 - 2\gamma - M} \brackets{\sum_{k=1}^T h(\xk)}
	+ 2\gamma \paren{h(\x_T) - h(\x_0)}
	,
	\\
	&\geq 
	\paren{2 - 2\gamma - M} \brackets{\sum_{k=1}^T h(\xk)}
	- 2\gamma h(\x_0)
	.
}
\end{tsmath}
The last inequality uses $h(\x_T) \geq 0$.
Moving the initial optimality gap to the right-hand-side, we get
\aligns{
	\frac{1}{T} 
	\paren{2 - 2\gamma - M}
	\Expect{
		\sum_{k=1}^T h(\xk)
	}
	\leq 
	\frac{1}{T}\paren{
		\frac{2 (1+\gamma^2) D^2 a_{\max} d}{\eta_{\min}}
		+ 2\gamma h(\x_0)
	} + M \sigma^2.
}
Assuming $2 - 2\gamma - M > 0$ and dividing, we get
\aligns{
	\frac{1}{T} 
	\Expect{
		\sum_{k=1}^T h(\xk)
	}
	\leq
	\frac{1}{2-2\gamma-M} 
	\brackets{
		\frac{1}{T}\paren{
			\frac{2 (1+\gamma^2) D^2 a_{\max} d}{\eta_{\min}}
			+ 2\gamma h(\x_0)
		} + M \sigma^2
	}.
}
Using Jensen's inequality and averaging the iterates finishes the proof.
\end{proof} \qedhere

\clearpage
\section{Experimental details}
\label{app:exp-details}

Our proposed adaptive gradient methods with SLS and SPS step-sizes are presented in~\cref{alg:conservative-sls,alg:conservative-sps}. We now make a few additional remarks on the practical use of these methods.

\begin{algorithm}[H]
    \caption{Adaptive methods with SLS($f$, $\texttt{precond}$, $\beta$, $\texttt{conservative}$, $\texttt{mode}$, $\x_0$, $\eta_{\max}$, $b$, $c \in (0,1)$, $\gamma < 1$)}
    \label{alg:conservative-sls}
    \begin{algorithmic}[1]
    \For{$k = 0, \dots, T-1$}
    \State $i_k \gets$ sample mini-batch of size $b$
    \State $\Ak\gets \texttt{precond}(k)$ \Comment{Form the preconditioner.}
    \If {\texttt{mode} $==$ \texttt{Lipschitz}}
    \State $p_k\gets\sgradf{i_k}{\xk}$
    \ElsIf {\texttt{mode} $==$ \texttt{Armijo}}
    \State $p_k\gets\invAk\sgradf{i_k}{\xk}$
    \EndIf
    \If {\texttt{conservative}}
        \If {k == 0}
            \State $\etak\gets\eta_{\max}$
        \Else
            \State $\etak\gets\eta_{k-1}$
        \EndIf
    \Else
    \State $\etak\gets\eta_{\max}$
    \EndIf
    \While { $\fk(\xk-\etak\cdot p_k) > \fk(\xk) -  c \, \etak \,  \inner{\sgradf{i_k}{\xk}}{p_k}$} \Comment{Line-search loop}
    \State $\etak \gets$ $\gamma \, \etak$
    \EndWhile
    \State $\mk \gets \beta \mkm + (1 - \beta) \sgradf{i_k}{\xk}$
    \State $\xkk \gets \xk - \etak \invAk \mk$
    \EndFor
    \State \Return $\x_T$
    \end{algorithmic}
\end{algorithm}
    
\begin{algorithm}[H]
    \caption{reset$(\eta, \eta_{\max}, k, b, n, \gamma, \texttt{opt})$}
    \label{alg:ls-reset}
    \begin{algorithmic}[1]
    \If {$k=0$}
    \State \Return $\eta_{\max}$
    \ElsIf {\texttt{opt}$=0$}
    \State $\eta\gets\eta$
    \ElsIf {\texttt{opt}$=1$}
    \State $\eta\gets\eta\cdot\gamma^{b/n}$
    \ElsIf{\texttt{opt}$=2$}
    \State $\eta\gets\eta_{\max}$
    \EndIf
    \State \Return $\eta$
    \end{algorithmic}
\end{algorithm}

As suggested by \citet{vaswani2019painless}, the standard backtracking search can sometimes result in step-sizes that are too small while taking bigger steps can yield faster convergence. To this end, we adopted their strategies to reset the initial step-size at every iteration (\cref{alg:ls-reset}). In particular, using reset option $0$ corresponds to starting every backtracking line search from the step-size used in the previous iteration. Since the backtracking never increases the step-size, this option enables the ``conservative step-size`` constraint for the Lipschitz line-search to be automatically satisfied. For the Armijo line-search, we use the heuristic from~\cite{vaswani2019painless} corresponding to reset option $1$. This option begins every backtracking with a slightly larger (by a factor of $\gamma^{\nicefrac{b}{n}}$, $\gamma = 2$ throughout our experiments) step-size compared to the step-size at the previous iteration, and works well consistently across our experiments. Although we do not have theoretical guarantees for Armijo SLS with general preconditioners such as Adam, our experimental results indicate that this is in fact a promising combination that also performs well in practice. 

\begin{algorithm}[H]
    \caption{Adaptive methods with SPS($f$, $[f_i^*]_{i = 1}^{n}$, $\texttt{precond}$, $\beta$,$\texttt{conservative}$,  $\texttt{mode}$, $\x_0$, $\eta_{\max}$, $b$, $c$)}
    \label{alg:conservative-sps}
    \begin{algorithmic}[1]
    \For{$k = 0, \dots, T-1$}
    \State $i_k \gets$ sample mini-batch of size $b$
    \State $\Ak\gets \texttt{precond}(k)$ \Comment{Form the preconditioner}
    \If {\texttt{mode} $==$ \texttt{Lipschitz}}
    \State $p_k\gets\sgradf{i_k}{\xk}$
    \ElsIf {\texttt{mode} $==$ \texttt{Armijo}}
    \State $p_k\gets\invAk\sgradf{i_k}{\xk}$
    \EndIf
  
    \If {\texttt{conservative}}
        \If {k == 0}
            \State $\eta_{B} \gets \eta_{\max}$
        \Else
            \State $\eta_{B} \gets\eta_{k-1}$
        \EndIf
    \Else
    \State $\eta_{B} \gets \eta_{\max}$
    \EndIf
    \State $\etak\gets\min\left\{\frac{f_{i_k}(\xk)-f^*_{i_k})}{c \, \inner{\sgradf{i_k}{\xk}}{p_k}},\,\eta_{B}\right\}$
    \State $\mk \gets \beta \mkm + (1 - \beta) \sgradf{i_k}{\xk}$
    \State $\xkk \gets \xk - \etak \invAk \mk$
    \EndFor
    \State \Return $\x_T$
    \end{algorithmic}
\end{algorithm}

On the other hand, rather than being too conservative, the step-sizes produced by SPS between successive iterations can vary wildly such that convergence becomes unstable. \citet{loizou2020stochastic} suggested to use a smoothing procedure that limits the growth of the SPS from the previous iteration to the current. We use this strategy in our experiments with $\tau=2^{\nicefrac{b}{n}}$ and show that both SPS and Armijo SPS work well. For the convex experiments, for both SLS and SPS, we set $c = 0.5$ as is suggested by the theory. For the non-convex experiments, we observe that all values of $c \in [0.1, 0.5]$ result in reasonably good performance, but use the values suggested in~\cite{vaswani2019painless,loizou2020stochastic}, i.e. $c = 0.1$ for all adaptive methods using SLS and $c = 0.2$ for methods using SPS. 
\clearpage
\section{Additional experimental results}
\label{app:additional-exps}

\begin{figure}[ht]
    \centering
    \subfigure[AdaGrad]{
    \includegraphics[scale =0.22]{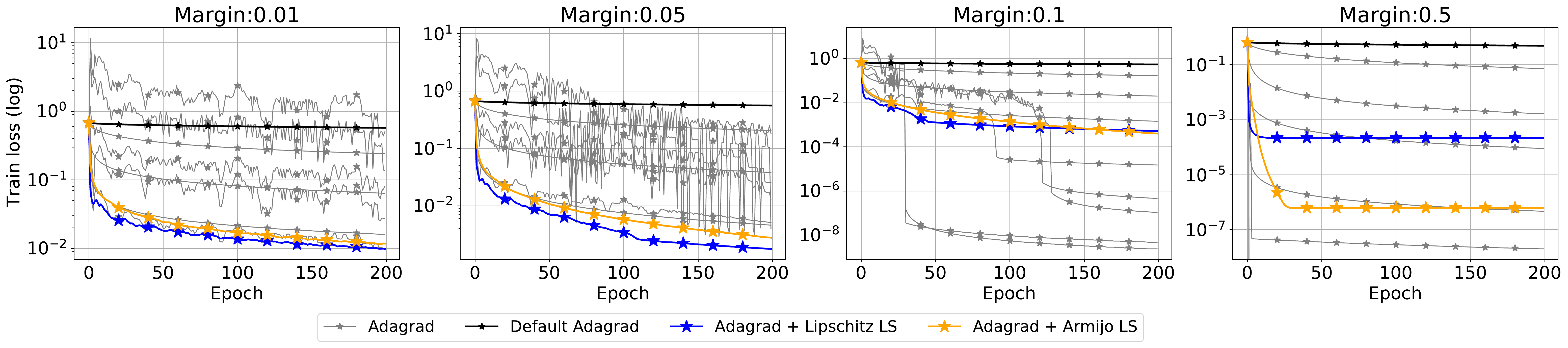}
    }
    \subfigure[AMSGrad]{
    \includegraphics[scale =0.22]{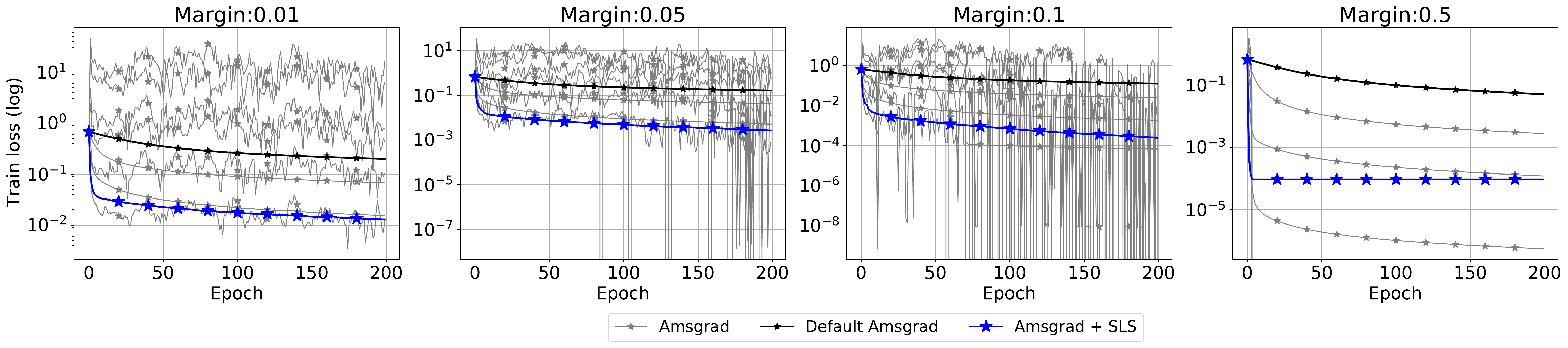}
    }
    \caption{Effect of step-size on the performance of adaptive gradient methods for binary classification on a linearly separable synthetic dataset with different margins. We observe that the large variance for the adaptive gradient methods, and the variants with SLS have consistently good performance across margins and optimizers.}
    \label{fig:app-syn}
\end{figure}

\begin{figure}[ht]
    \centering
    \subfigure[]{ \includegraphics[scale=0.3]{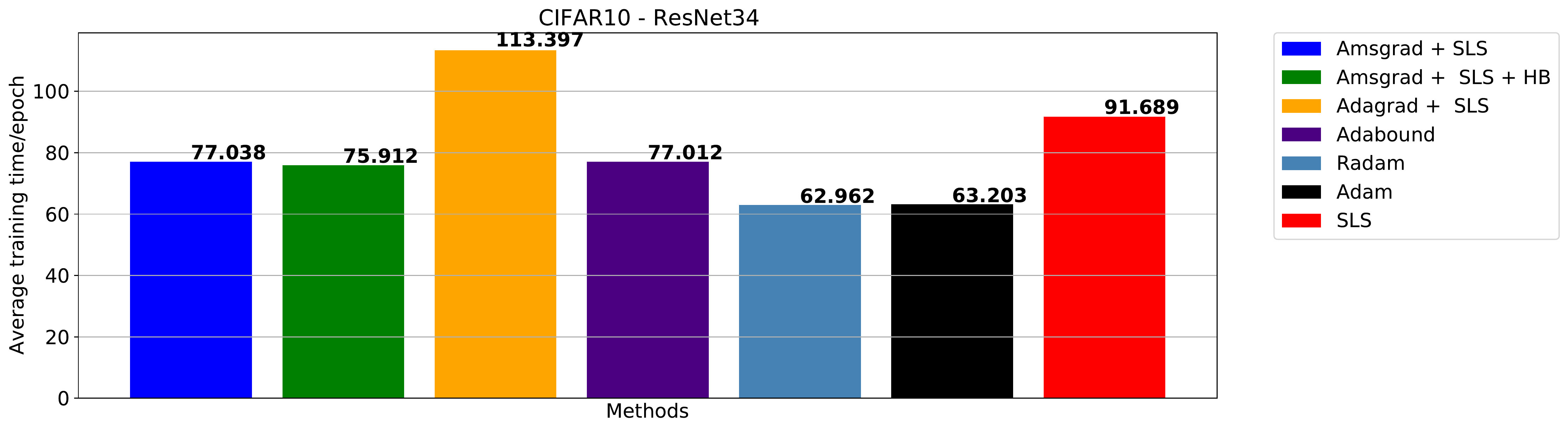}}
    \subfigure[]{ \includegraphics[scale=0.3]{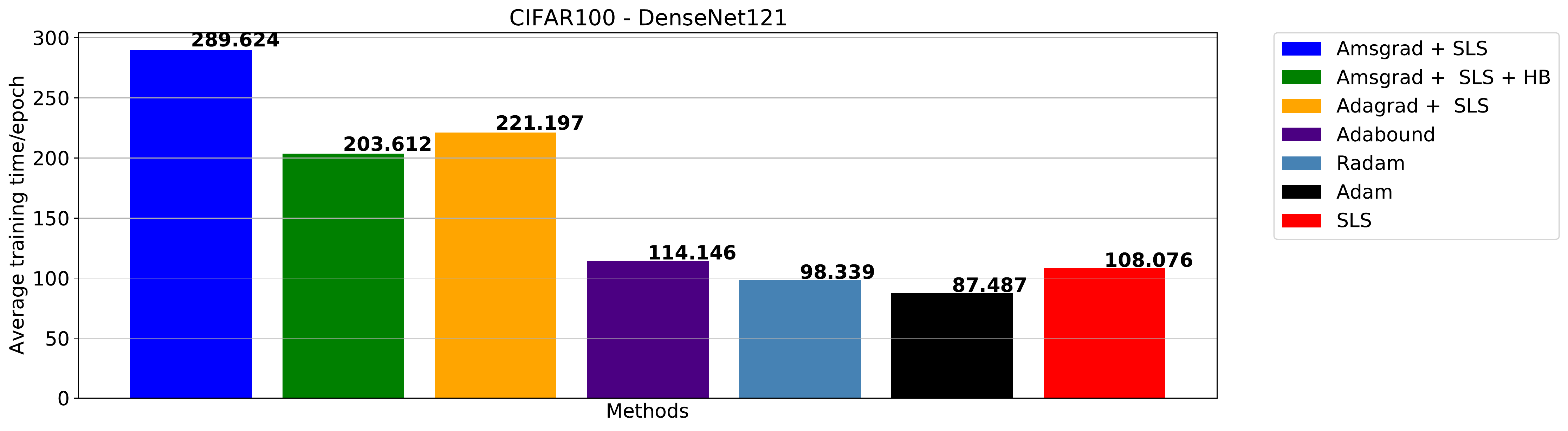}}
    \subfigure[]{ \includegraphics[scale=0.3]{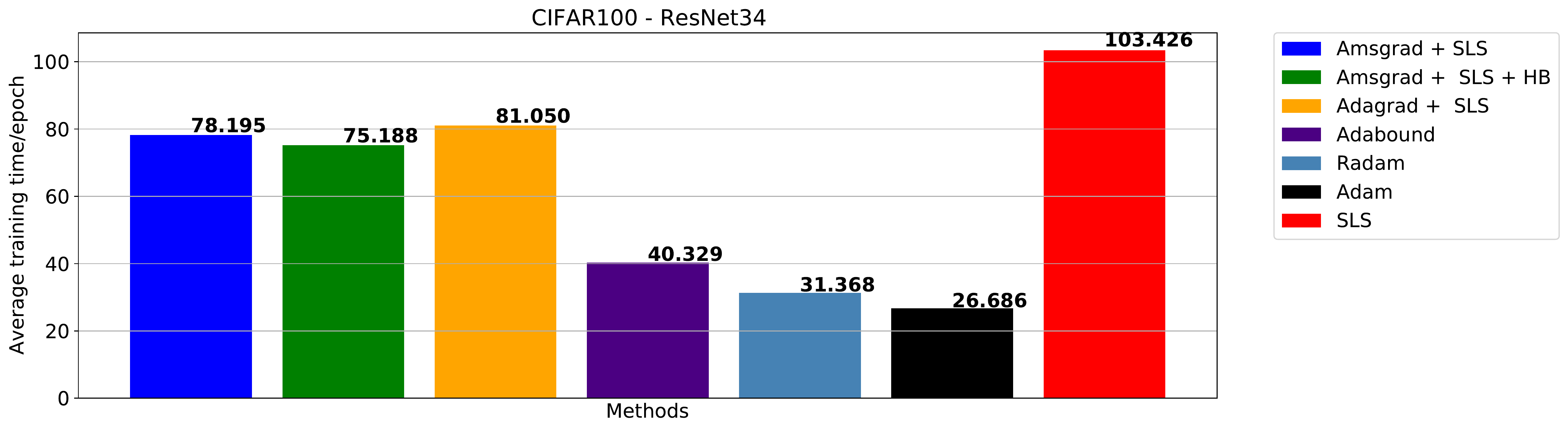}}
    \subfigure[]{ \includegraphics[scale=0.3]{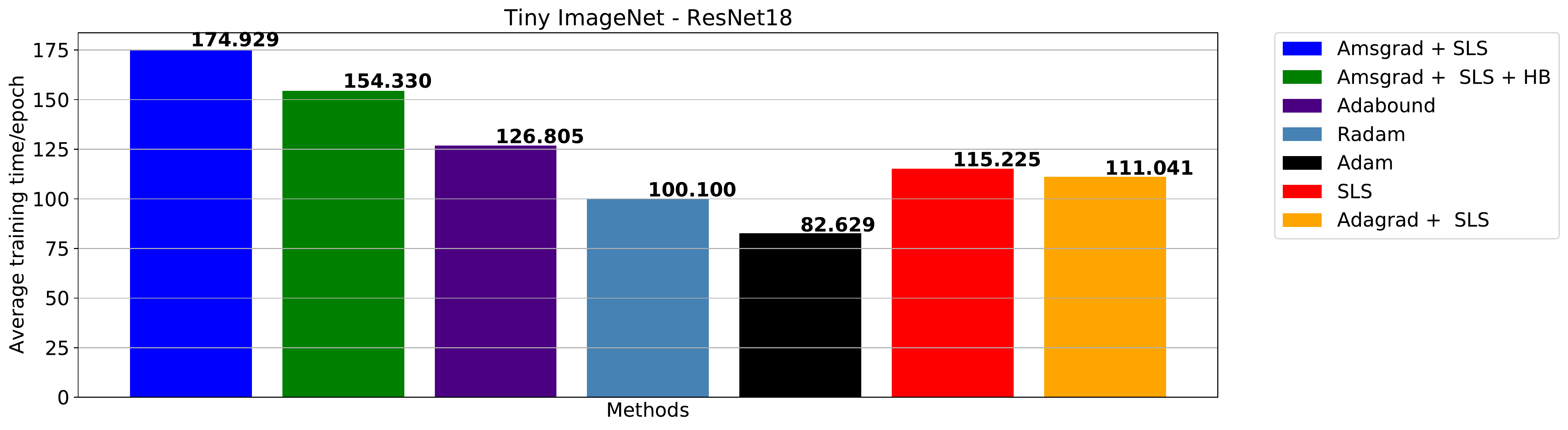}}
    \caption{Runtime (in seconds/epoch) for optimization methods for multi-class classification using the deep network models in ~\cref{fig:deep}. Although the runtime/epoch is larger for the SLS/SPS variants, they require fewer epochs to reach the maximum test accuracy (Figure~\ref{fig:deep}). This justifies the moderate increase in wall-clock time. }
    \label{fig:app-deep-runtime}
\end{figure}

\begin{figure}[!ht]
  \centering
  \subfigure[CIFAR-10 ResNet]{
    \includegraphics[scale =0.3]{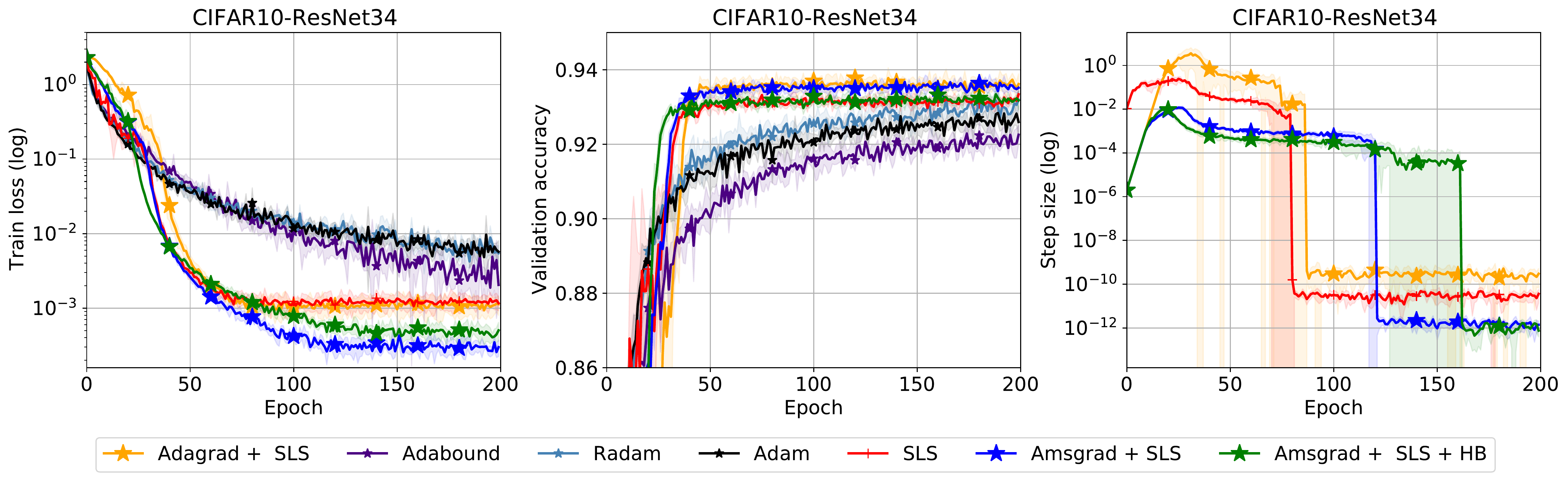}
    }
    \subfigure[CIFAR-10 DenseNet]{
    \includegraphics[scale =0.3]{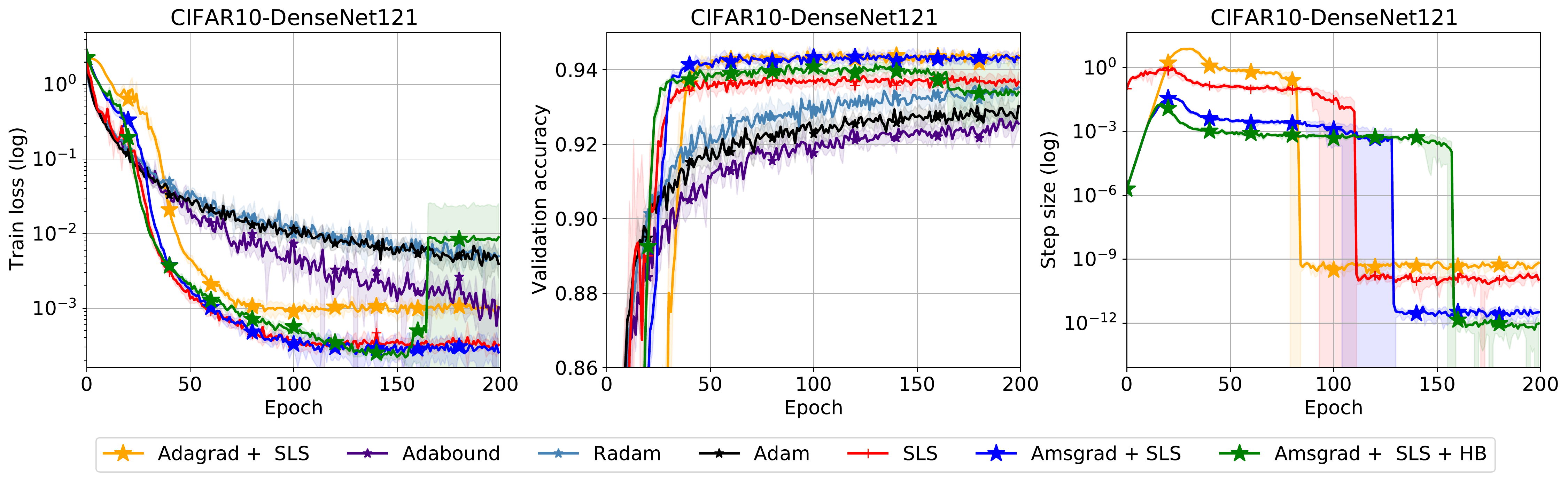}
    }
    \subfigure[CIFAR-100 ResNet]{
    \includegraphics[scale =0.3]{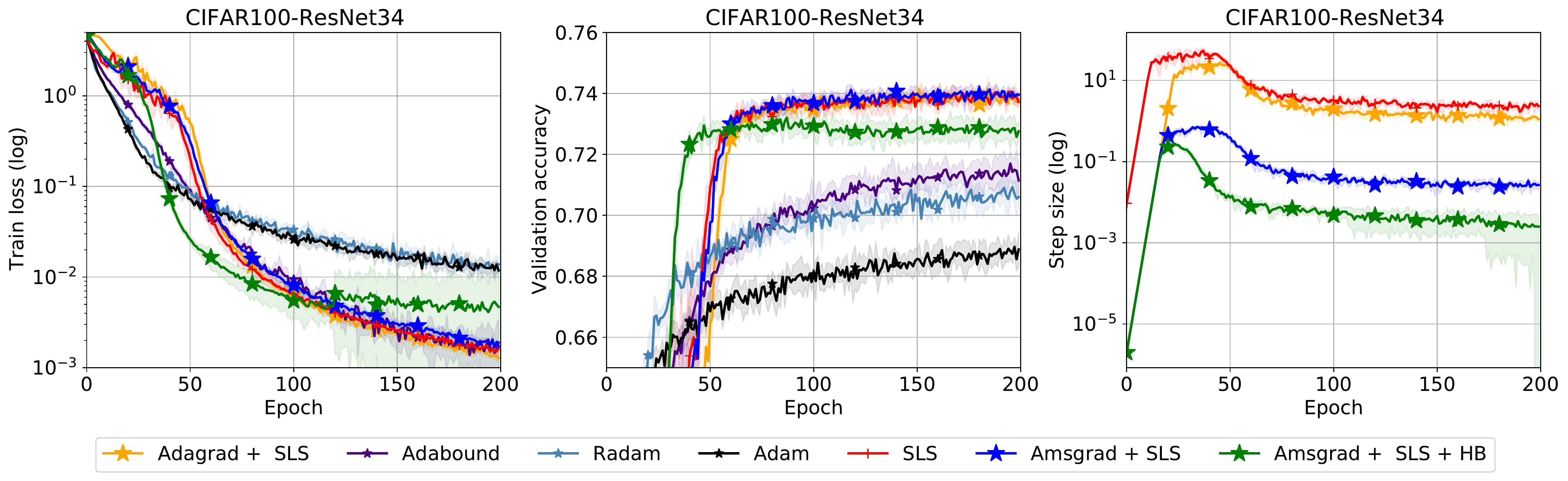}
    }
    \subfigure[CIFAR-100 DenseNet]{
    \includegraphics[scale =0.3]{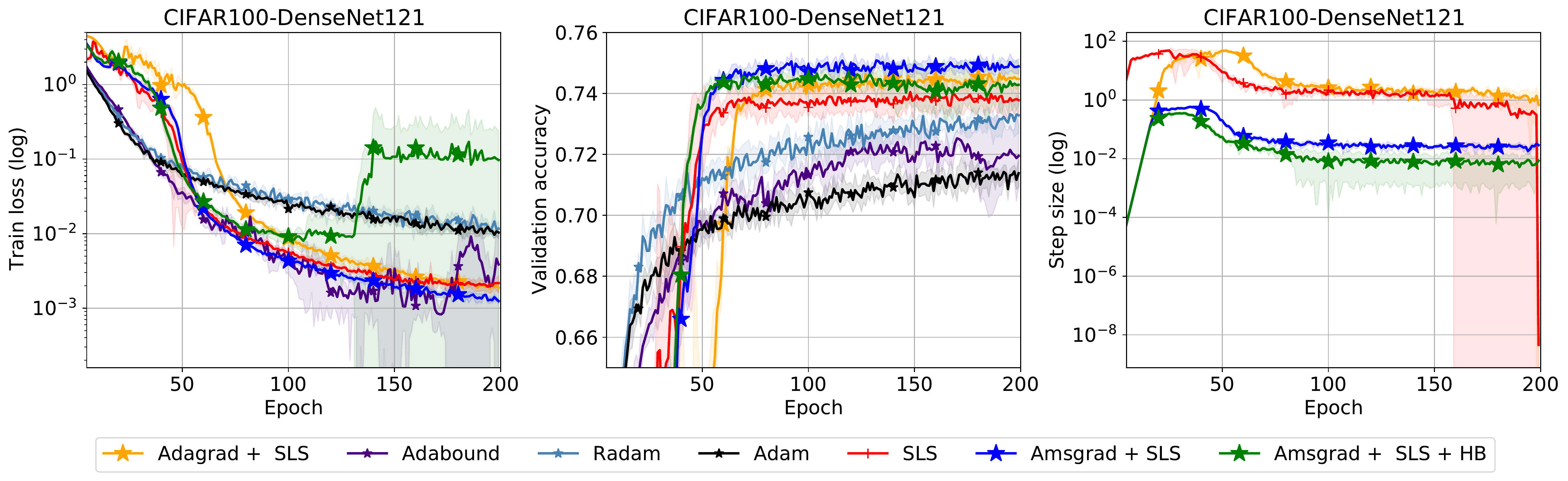}
    }
    \caption{Comparing optimization methods on image classification tasks using ResNet and DenseNet models on the CIFAR-10/100 datasets. For the SLS/SPS variants, refer to the experimental details in~\cref{app:exp-details}. For Adam, we did a grid-search and use the best step-size. We use the default hyper-parameters for the other baselines. We observe the consistently good performance of \adagrad/ and \amsgrad/ with Armijo SLS. We also show the variation in the step-size and observe a cyclic pattern~\citep{loshchilov2016sgdr} - an initial warmup in the learning rate followed by a decrease or saturation to a small step-size~\citep{goyal2017accurate}.}
   \label{fig:app-deep-cifar}    
\end{figure}

\begin{figure}
    \centering
    \subfigure[Imagewoof]{
    \includegraphics[scale =0.3]{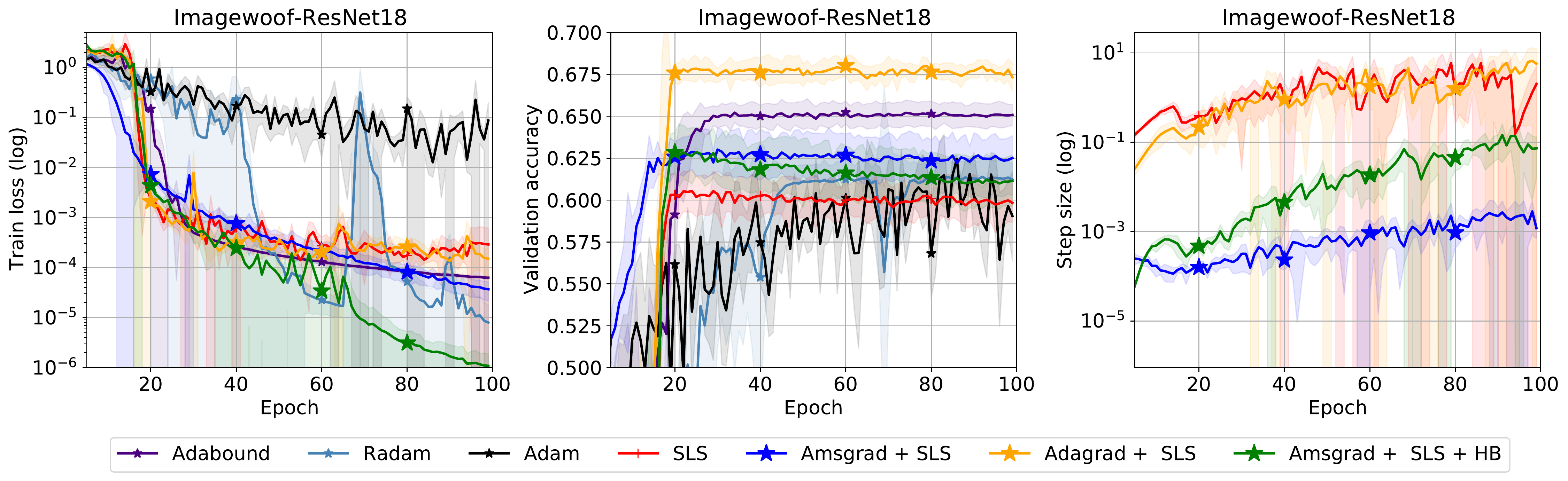}    
    }
    \subfigure[ImageNette]{
    \includegraphics[scale =0.3]{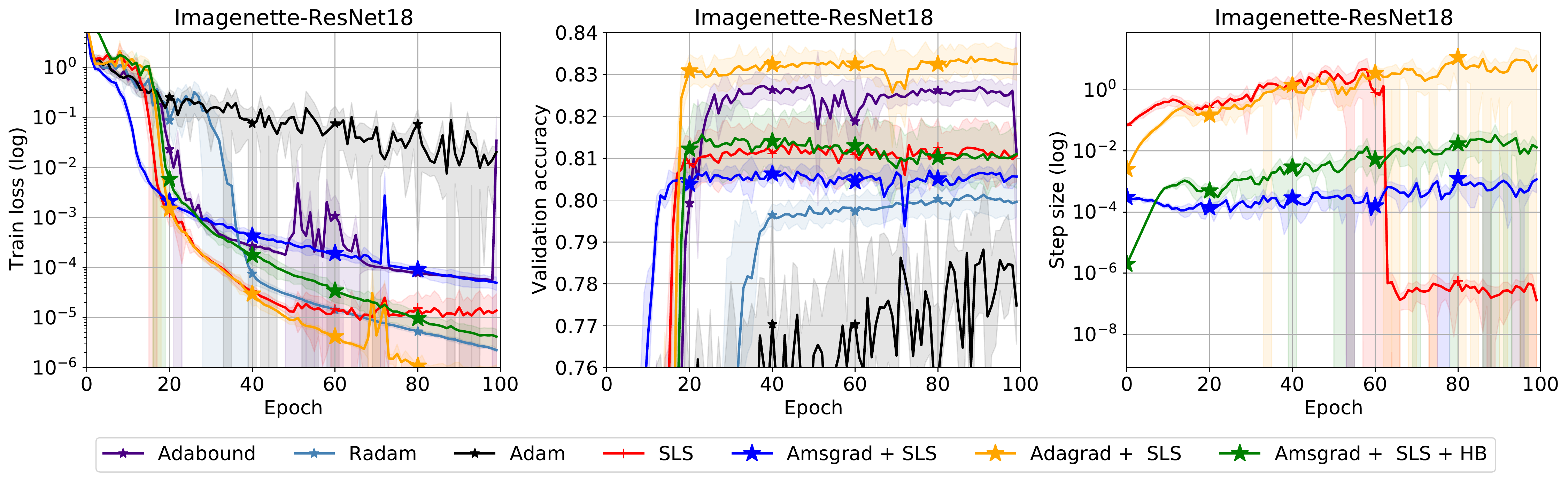}    
    }
    \subfigure[Tiny Imagenet]{
    \includegraphics[scale =0.3]{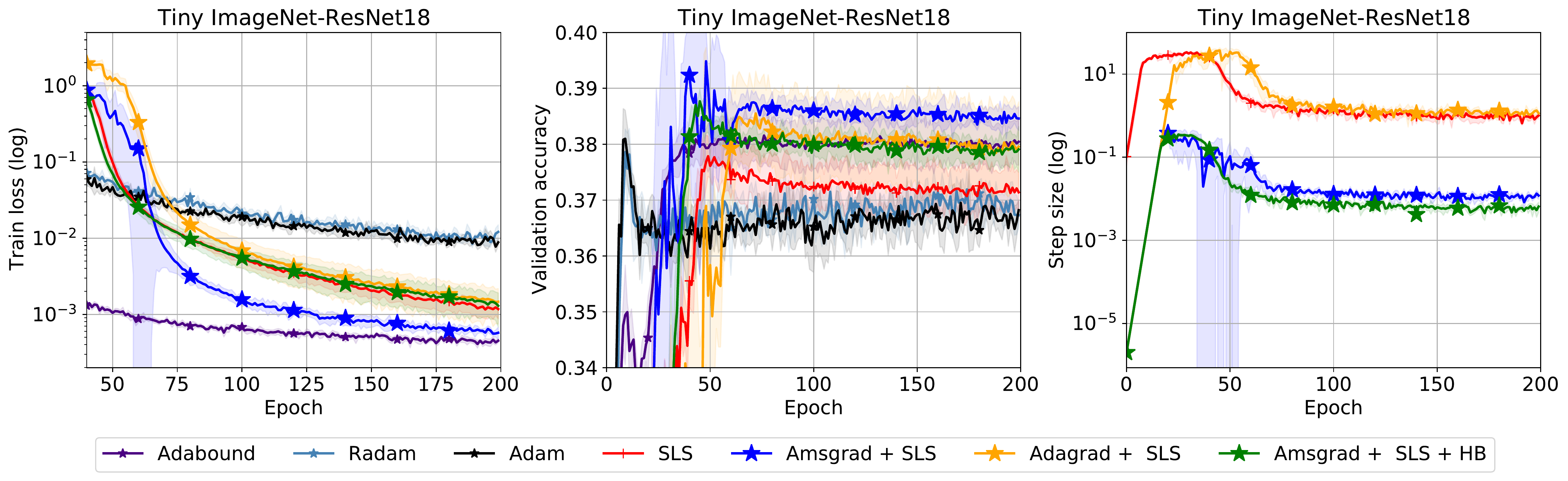}    
    }
    \caption{Comparing optimization methods on image classification tasks using variants of ImageNet. We use the same settings as the CIFAR datasets and observe that \adagrad/ and \amsgrad/ with Armijo SLS is consistently better.}
   \label{fig:app-deep-imagenet}  
\end{figure}

\begin{figure}
    \centering
    \includegraphics[scale =0.32]{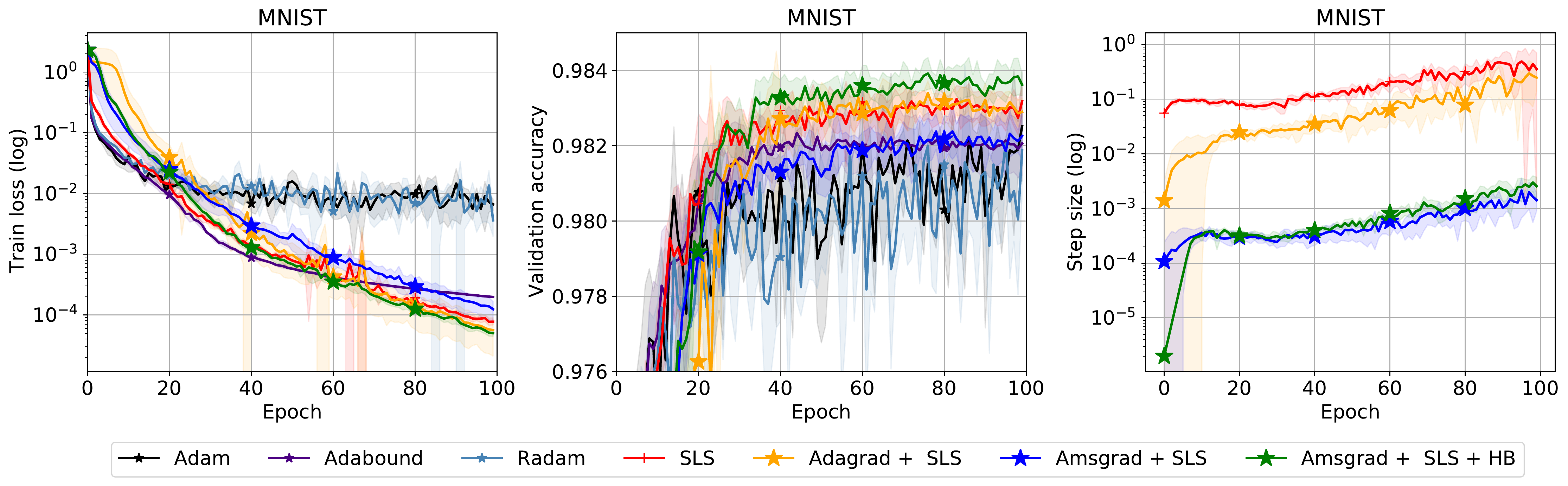}
    \caption{Comparing optimization methods on MNIST.}
    \label{fig:app-mnist}
\end{figure}

\begin{figure}[!ht]
    \centering
    \includegraphics[scale =0.31]{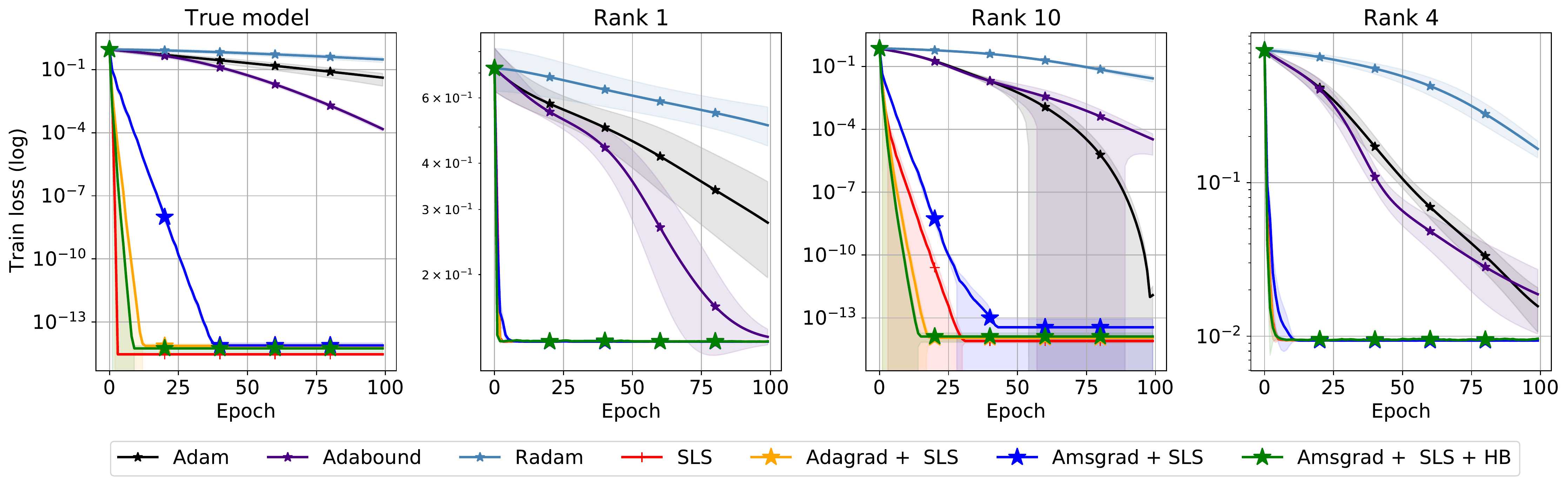}
    \caption{Comparison of optimization methods for deep matrix factorization. Methods use the same hyper-parameter settings as above and we examine the effects of over-parameterization on the problem: $\min_{W_1, W_2} \E_{x \sim N(0,I)} \norm{W_2 W_1 x - Ax}\kern-.1em{}^2$~\citep{vaswani2019painless, rolinek2018l4}. We choose $A \in \mathbb{R}^{10 \times 6}$ with condition number $\kappa(A) = 10^{10}$ and control the over-parameterization via the rank $k$ (equal to 1,4, 10)  of $W_1 \in \mathbb{R}^{k \times 6}$ and $W_2 \in \mathbb{R}^{10 \times k}$. We also compare against the true model. In each case, we use a fixed dataset of $1000$ samples. We observe that as the over-parameterization increases, the performance of all methods improves, with the methods equipped with SLS performing the best. 
    }
    \label{fig:app-mf}
\end{figure}

\begin{figure}[!ht]
    \centering
    \subfigure{
    \hskip-.3em\includegraphics[scale=0.3025]{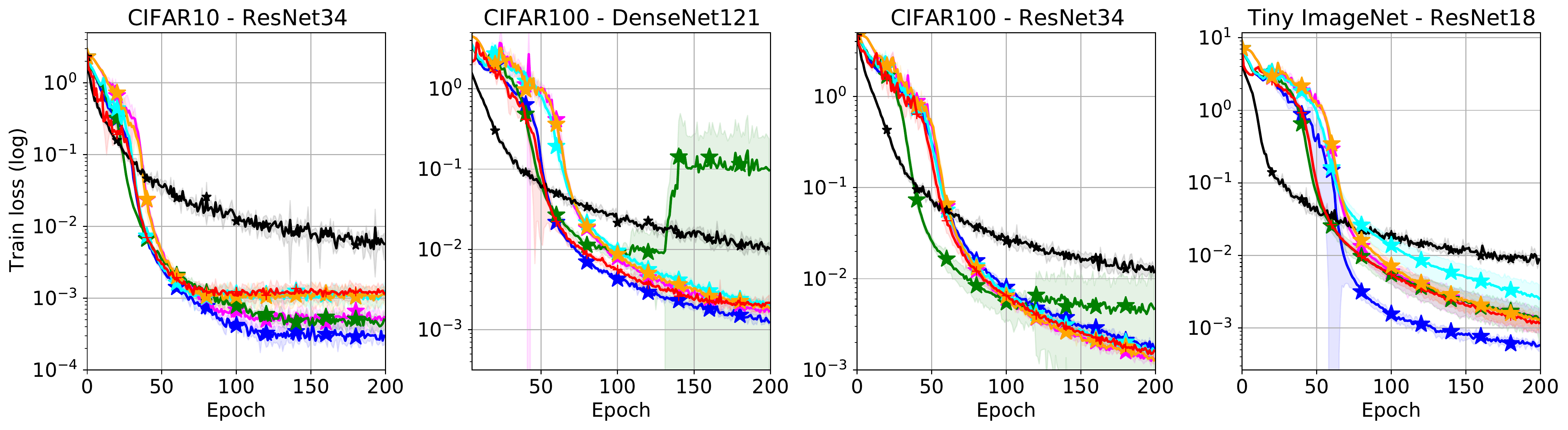}
    }\\[-.5em]%
    \subfigure{
    \includegraphics[scale=0.3]{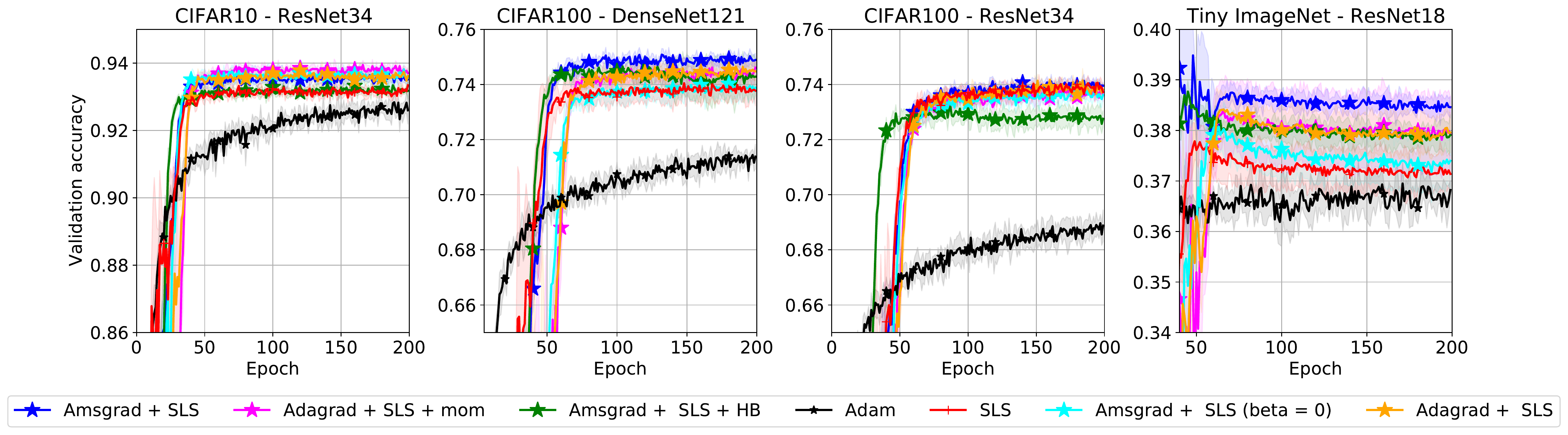}
    }\vskip-.5em
    \caption{Ablation study comparing variants of the basic optimizers for multi-class classification with deep networks. Training loss (top) and validation accuracy (bottom) for CIFAR-10, CIFAR-100 and Tiny ImageNet. We consider the \adagrad/ with \amsgrad/-like momentum and do not find improvements in performance. We also benchmark the performance of \amsgrad/ without momentum, and observe that incorporating \amsgrad/ momentum does improve the performance, whereas heavy-ball momentum has a minor, sometimes detrimental effect. We use SLS and Adam as benchmarks to study the effects of incorporating preconditioning vs step-size adaptation.}
\label{fig:deep-extra}    
\vspace{-3ex}
\end{figure}

This section presents additional experimental results showing the effect of the step-size for adaptive gradient methods using a synthetic dataset (\cref{fig:app-syn}). 
We show the wall-clock times for the optimization methods (\cref{fig:app-deep-runtime}). 
We show the variation in the step-size for the SLS methods when training deep networks for both the CIFAR in~\cref{fig:app-deep-cifar} and ImageNet (\cref{fig:app-deep-imagenet}) datasets. 
We evaluate these methods on easy non-convex objectives - classification on MNIST (\cref{fig:app-mnist}) and deep matrix factorization 
to examine the effect of over-parameterization on the performance of the optimization methods
(\cref{fig:app-mf}). 
Finally in~\cref{fig:deep-extra}, we quantify the gains of incorporating momentum in \amsgrad/ by comparing against the  performance \amsgrad/ \emph{without momentum}.  

\end{document}